\documentclass{article}
\usepackage{times}

\usepackage{amsmath,amsfonts,bm}

\def\eqref#1{equation~\ref{#1}}

\def\1{\bm{1}}

\DeclareMathAlphabet{\mathsfit}{\encodingdefault}{\sfdefault}{m}{sl}
\SetMathAlphabet{\mathsfit}{bold}{\encodingdefault}{\sfdefault}{bx}{n}

\DeclareMathOperator*{\argmax}{arg\,max}

\DeclareMathOperator{\sign}{sign}

\usepackage[margin=1in]{geometry}
\usepackage[T1]{fontenc}
\usepackage{amsmath,amssymb,amsfonts,bm}
\usepackage{graphicx}
\usepackage{float}
\usepackage[section]{placeins}
\usepackage{tikz}
\usepackage{booktabs}
\usepackage{longtable}

\usepackage{adjustbox}
\usepackage{algorithm}
\usepackage{algpseudocode}
\usepackage[expansion=false]{microtype}
\usepackage{caption}
\usepackage{enumitem}
\usepackage[colorlinks=true,linkcolor=black,citecolor=black,urlcolor=blue,unicode=false]{hyperref}
\usepackage{bookmark}
\usepackage{natbib}
\usepackage{pdflscape}
\DeclareMathOperator{\median}{median}
\DeclareMathOperator{\diag}{diag}
\providecommand{\std}[1]{\ensuremath{_{\scriptscriptstyle \pm #1}}}


\title{Extremely Fast and Compact Binary Graph Representations via Randomized Operator Sketching}

\author{Srajan Agarwal$^{1}$ \quad Megha P$^{1}$ \quad Bikas C Das$^{2}$ \quad Zakaria Laskar$^{1}$ \quad Saptarshi Bej$^{1}$ \\
	$^{1}$School of Data Science, IISER Thiruvananthapuram, India \\
	$^{2}$Department of Physical Sciences, IISER Berhampur, India}

\date{}

\usepackage{lineno}
\begin{document}
	\raggedbottom
	\maketitle
	
	\begin{abstract}
		Graph neural networks typically rely on dense, floating-point node vectors, which impose severe memory and computational bottlenecks. Binary graph hashing provides an alternative by compressing node information into compact bit strings. However, prior techniques either sacrifice global topological context for speed or suffer from prohibitive generation costs. To bridge this gap, we introduce an ultra-fast, entirely algebraic hashing method that constructs highly discriminative binary codes directly from graph structure, bypassing the need for node features or gradient-based training. Our approach approximates the high-order structural transition matrix using a randomized column-sampling technique inspired by the Nystr\"om method, and subsequently merges it with an efficient, label-safe semantic propagation mechanism. By discretizing the resulting continuous blend via column-wise thresholding, we produce geometry-preserving discrete representations. Comprehensive evaluations across ten node classification datasets demonstrate that our method consistently exceeds the classification accuracy of existing feature-free binary baselines, achieving this with sub-second generation latencies, often orders of magnitude faster than iterative or sketch-based alternatives. Moreover, because our discrete vectors map natively to event-driven computations, they pair exceptionally well with neuromorphic spiking neural networks and gradient-free learning rules. Ultimately, we show that simple, theoretically grounded algebraic approximations can decisively outperform complex learned pipelines for discrete representation learning.
	\end{abstract}

	\section{Introduction}
	Modern graph machine learning produces high-quality node representations, but the
	representations themselves are usually dense floating-point vectors. At scale this
	is a bottleneck. A single-precision embedding of width $256$ occupies one
	kilobyte per node, so a graph with a hundred million nodes needs a hundred
	gigabytes just to store the embedding table, before any inference. In resource-constrained
	environments, both the storage and the
	cost of floating-point similarity are limiting.
	
	Binary node codes are a direct response. A $K$-bit code is $32\times$ smaller than
	a $K$-dimensional \texttt{float32} vector, significantly accelerating downstream
	classification tasks while maintaining a highly compact memory footprint. The open question
	is not whether binary codes are desirable but whether we can build codes that are
	simultaneously (i) fast to construct, (ii) discriminative enough for downstream
	tasks, and (iii) portable across a range of classifiers, including non-standard
	ones. Existing work tends to satisfy at most two of the three.
	
	Two broad families dominate. \emph{Learned} binarizations, exemplified by Bi-GCN
	\citep{wang2021bigcn} and \texttt{node2binary} \citep{talukder2025node2binary},
	optimize an objective and typically preserve accuracy, but their construction is
	slow, sometimes minutes to hours per graph. \emph{Sketching and projection}
	methods, such as NodeSketch \citep{yang2019nodesketch} and NodeSig
	\citep{celikkanat2022nodesig}, avoid training, but recursive multi-hop sketching
	scales poorly and shallow projections lose accuracy on graphs with intricate
	structure. In practice one must decide whether to pay in time or in quality.
	
	We introduce a feature-free, algebraic framework for constructing compact binary graph representations directly from graph topology. Our central contribution is a unified, modular algebraic pipeline that couples low-rank structural sketching of higher-order transition operators with a label-leakage-safe semantic diffusion channel, polynomial spectral smoothing, and adaptive quantile thresholding. While the structural channel can accommodate various low-rank approximations, we find that an asymmetric, randomized Nystr\"om-inspired landmark extension provides a highly effective closed-form alternative: it completely avoids the repetitive matrix-vector products and polynomial recursions of iterative approaches, delivering a favorable accuracy-efficiency trade-off compared to full spectral decomposition. Layering the label-leakage-safe semantic diffusion channel on top of these structural coordinates yields binary codes that achieve state-of-the-art downstream performance while substantially reducing construction cost and storage relative to learned binary representation methods.
	
	\paragraph{Contributions.}
	\begin{enumerate}[leftmargin=*,itemsep=2pt]
		\item We introduce NyS-Binary, a closed-form binary graph hashing framework
		that couples a randomized Nystr\"om-inspired sketch of a higher-order transition
		operator with a label-leakage-safe semantic diffusion channel and adaptive median
		quantization (Section~\ref{sec:method}). The construction is closed-form
		and near-linear in the graph size $(N + |E|)$ for sparse real-world networks.
		\item We systematically evaluate on ten benchmarks and nine downstream classifiers, reporting
		competitive or superior performance across diverse probe architectures. We evaluate both a semi-supervised blend and a strictly unsupervised variant ($\text{struct\_ratio}=1.0$); our unsupervised variant establishes a new state-of-the-art among label-free, feature-free binary hashing baselines at sub-second generation cost, while our semi-supervised blend provides further discriminative refinement (Section~\ref{sec:experiments}).
		\item We evaluate on neuromorphic spiking architectures, including SADP (a classifier trained via gradient-free agreement-driven synaptic learning without backpropagation \citep{s2026building}) and a surrogate-gradient spiking network. We find that our discrete binary codes natively match event-driven spike dynamics without lossy continuous-to-spike rate coding, achieving strong compatibility with neuromorphic substrates and leading prior binary hashing baselines (e.g., $+26.81\%$ over NodeSig) as well as the learned Bi-GCN baseline ($+9.45\%$) on SADP (Section~\ref{sec:experiments}).
		\item We analyze the geometry of the codes through class-separation ratios (Section~\ref{sec:analysis}), and document the method's behavior on strongly heterophilous graphs under extended diffusion in the appendix (Appendix~\ref{sec:appendix_deep_diffusion_heterophily}).
	\end{enumerate}
	
	\section{Related Work}
	\label{sec:related}
	\paragraph{Node embeddings and hashing.}
	Random-walk embeddings such as DeepWalk \citep{perozzi2014deepwalk} and node2vec
	\citep{grover2016node2vec} popularized skip-gram style objectives for graphs, and
	matrix-factorization views unified several of them. For scalability, hashing and
	sketching approaches map nodes into Hamming space. NetHash
	\citep{zhang2018nethash} hashes rooted trees of node attributes; NodeSketch
	\citep{yang2019nodesketch} recursively sketches the self-loop augmented adjacency
	matrix to preserve high-order proximity; NodeSig \citep{celikkanat2022nodesig}
	applies sign-stable random projections to random-walk diffusion probabilities.
	AROPE \citep{zhang2018arope} preserves arbitrary-order proximity via
	eigendecomposition, and SimHash \citep{charikar2002simhash} is a classical
	sign-random-projection technique. \texttt{node2binary}
	\citep{talukder2025node2binary} learns binary vectors through an explicit
	optimization. Our method is closest in spirit to the sketching family but differs
	by anchoring structure in a Nystr\"om-inspired low-rank frame and by adding an explicit,
	label-leakage-safe semantic channel.
	
	\paragraph{Binarized graph neural networks.}
	Bi-GCN \citep{wang2021bigcn} binarizes weights and activations of a graph
	convolutional network, reporting large memory and speed savings at near
	full-precision accuracy. Bi-GCN represents an end-to-end learned, high-capacity
	binary baseline that optimizes an objective using continuous node features and iterative backpropagation. In contrast, NyS-Binary is a closed-form, training-free algebraic framework that operates purely on graph topology, generating binary codes orders of magnitude faster while maintaining competitive or superior downstream accuracy.
	
	\paragraph{Low-rank approximation.}
	The Nystr\"om method \citep{williams2001nystrom} approximates a matrix from
	sampled columns and has been studied in depth for kernel machines
	\citep{drineas2005nystrom}; randomized SVD \citep{halko2011randomized} offers
	related probabilistic guarantees. We apply these ideas to a graph transition
	operator rather than a kernel matrix.
	
	\paragraph{Downstream and spiking classifiers.}
	We evaluate with GCN \citep{kipf2017gcn}, GraphSAGE \citep{hamilton2017graphsage},
	H2GCN \citep{zhu2020h2gcn}, and LINKX \citep{lim2021linkx}, alongside linear
	probes. To evaluate deployment feasibility on brain-inspired, ultra-low-power neuromorphic hardware, we evaluate spiking neural networks trained by gradient-free agreement-driven learning (SADP) and surrogate-gradient dynamics \citep{neftci2019surrogate,maass1997snn,s2026building}. Because spiking architectures compute via discrete all-or-nothing events, our binary codes naturally match event-driven spike dynamics without requiring continuous rate coding, demonstrating the strong compatibility of our representations on neuromorphic computing substrates.
	
	\section{Method}
	\label{sec:method}
	\subsection{Notation and problem setting}
	Let $\mathcal{G}=(V,E)$ be an undirected graph with $N=|V|$ nodes and adjacency
	$\mathbf{A}\in\{0,1\}^{N\times N}$. We are given training labels
	$\{y_i\}_{i\in\mathcal{M}_{\mathrm{tr}}}$ over $C$ classes on a mask
	$\mathcal{M}_{\mathrm{tr}}\subset\{1,\dots,N\}$, and we output a binary code
	matrix $\mathbf{B}\in\{0,1\}^{N\times K}$ for a target length $K$. In our canonical sparsely supervised setting, we divide the
	budget as $K=k_{\mathrm{struct}}+k_{\mathrm{label}}$ with $k_{\mathrm{struct}}=\lfloor K/2\rfloor$
	structural bits and $k_{\mathrm{label}}=K-k_{\mathrm{struct}}$ semantic bits (a balanced 50/50 split).
	Importantly, our framework also supports a strictly unsupervised operational mode
	by allocating the entire bit budget to structural coordinates ($\text{struct\_ratio} = 1.0$,
	i.e., $k_{\mathrm{struct}} = K, k_{\mathrm{label}} = 0$). In this unsupervised regime, the semantic
	codebook and label diffusion are bypassed entirely, generating $K$-bit representations
	purely from the graph topology without access to any node attributes or training labels.
	
	\subsection{Transition operator}
	We first augment self-loops and row-normalize to obtain a random-walk operator,
	\begin{equation}
		\tilde{\mathbf{A}}=\mathbf{A}+\mathbf{I}_N,\qquad
		\tilde{\mathbf{D}}=\diag\!\Big(\textstyle\sum_{j}\tilde{\mathbf{A}}_{ij}\Big),\qquad
		\mathbf{P}=\tilde{\mathbf{D}}^{-1}\tilde{\mathbf{A}}.
	\end{equation}
	The self-loop keeps a node's own signal during propagation, and row
	normalization makes $\mathbf{P}$ row-stochastic, so $\mathbf{P}^t$ encodes the
	distribution of $t$-step walks.
	
	\subsection{Structural channel via randomized Nystr\"om-inspired sketching}
	\label{sec:method_structural}
	Rather than decomposing $\mathbf{P}$ in full, which costs $O(N^3)$, we sketch it
	from a landmark subset. While the classical Nystr\"om method is formulated for symmetric positive semi-definite kernel matrices \citep{williams2001nystrom}, low-rank sketching of asymmetric operators via sampled submatrices is mathematically grounded in randomized numerical linear algebra through interpolative and asymmetric CUR decompositions \citep{drineas2005nystrom,drineas2006cur}. For our row-stochastic transition operator $\mathbf{P}$, we adopt an asymmetric column-projection sketch: we sample
	$m=\min\!\big(N,\max(k_{\mathrm{struct}},125)\big)$ landmark indices $\mathcal{L}$
	uniformly without replacement and extract the landmark-landmark core block
	$\mathbf{P}_{\ell\ell}=\mathbf{P}[\mathcal{L},\mathcal{L}]\in\mathbb{R}^{m\times m}$
	and the cross-block
	$\mathbf{P}_{n\ell}=\mathbf{P}[:,\mathcal{L}]\in\mathbb{R}^{N\times m}$. We compute the
	truncated SVD of the core block,
	$\mathbf{P}_{\ell\ell}=\mathbf{U}_\ell\bm{\Sigma}_\ell\mathbf{V}_\ell^\top$.
	Because $\mathbf{P}_{n\ell}$ consists of out-of-sample columns, we project them onto the left singular vectors $\mathbf{U}_\ell$ (which span the column space of the landmark core block) and scale by the inverse singular values to yield coordinates for every node:
	\begin{equation}
		\mathbf{R}_{\mathrm{str}}
		=\mathbf{P}_{n\ell}\,\mathbf{U}_\ell\,(\bm{\Sigma}_\ell+\epsilon\mathbf{I})^{-1}
		\in\mathbb{R}^{N\times k_{\mathrm{struct}}},
		\label{eq:nystrom}
	\end{equation}
	where $\epsilon$ is a small stabilizer that guards against tiny singular values.
	The dominant costs are the $O(m^3)$ decomposition of the landmark block and the
	$O(Nmk_{\mathrm{struct}})$ extension, both linear in $N$ for fixed landmark budget $m$, which is
	where the speed comes from. We emphasize that our framework is structurally modular: while we use a Nystr\"om-inspired construction for its strong accuracy-efficiency trade-off, this channel $\mathbf{R}_{\mathrm{str}}$ can be seamlessly swapped with other low-rank approximations (e.g., Randomized SVD, Chebyshev filtering) without altering the subsequent semantic diffusion or quantization stages (see Appendix~\ref{sec:appendix_backbone} for a comprehensive empirical comparison).
	
	\subsection{Semantic channel via label-leakage-safe label diffusion}
	Structure alone is class-agnostic, so we add label semantics without leaking test
	information. We first build an orthogonal class codebook by drawing a Gaussian
	matrix $\mathbf{X}\sim\mathcal{N}(0,1)\in\mathbb{R}^{k_{\mathrm{label}}\times C}$, transposing its
	thin QR factor $\mathbf{Q}$, and binarizing to hypercube vertices
	$\bm{h}_c=\sign(\mathbf{Q}^\top[c,:])\in\{-1,+1\}^{k_{\mathrm{label}}}$. Well-separated codewords
	keep classes apart in the semantic subspace.
	
	\emph{Ground-truth stream.} We assign $\mathbf{S}_{\mathrm{gt}}[i,:]=\bm{h}_{y_i}$
	for $i\in\mathcal{M}_{\mathrm{tr}}$ and $\mathbf{0}$ otherwise.
	
	\emph{Pseudo-label stream.} We propagate one-hot training seeds $\mathbf{Y}_0$ by
	a three-step walk, $\mathbf{Y}_t=\mathbf{P}^3\mathbf{Y}_0$, normalize to a
	posterior $\hat p_{i,c}=\mathbf{Y}_t[i,c]/(\sum_{c'}\mathbf{Y}_t[i,c']+\epsilon)$,
	and for each \emph{unlabeled} node with $\max_c\hat p_{i,c}\ge\tau$ we set
	$\mathbf{S}_{\mathrm{pl}}[i,:]=\bm{h}_{c_i^\star}$ where
	$c_i^\star=\argmax_c\hat p_{i,c}$; otherwise $\mathbf{0}$. Labeled nodes keep
	their ground-truth codeword. Under the standard transductive graph representation setting, the random-walk transition matrix $\mathbf{P}$ utilizes the full topological graph structure (including edges incident to test nodes), but the supervisory signal is strictly insulated: only training labels ever seed the walk ($\mathbf{S}_{\mathrm{gt}}[i,:]=\mathbf{0}$ for $i \notin \mathcal{M}_{\mathrm{tr}}$), and pseudo-labels are formed purely via structural random-walk confidence without accessing test ground-truth. We explicitly assert that no test label index $i \notin \mathcal{M}_{\mathrm{tr}}$ seeds the propagation, ensuring that the semantic channel remains strictly \emph{label-leakage-safe}.
	
	\subsection{Fusion, diffusion, and adaptive quantization}
	We concatenate the shared structural coordinates with each semantic stream,
	\begin{equation}
		\mathbf{W}_{\mathrm{gt}}=[\mathbf{R}_{\mathrm{str}}\,\|\,\mathbf{S}_{\mathrm{gt}}],\qquad
		\mathbf{W}_{\mathrm{pl}}=[\mathbf{R}_{\mathrm{str}}\,\|\,\mathbf{S}_{\mathrm{pl}}]
		\in\mathbb{R}^{N\times K},
	\end{equation}
	smooth each with a short polynomial diffusion that mixes local and higher-order
	neighborhoods,
	\begin{equation}
		\mathcal{D}_H(\mathbf{W})=\frac{1}{H+1}\sum_{h=0}^H \mathbf{P}^h \mathbf{W},
	\end{equation}
	and blend the two streams with a convex combination weight $\alpha \in [0, 1]$ before thresholding each
	column using an adaptive median rule scaled by quantile factor $\mu > 0$ (default $\mu = 0.50$):
	\begin{equation}
		\mathbf{W}=(1-\alpha)\,\mathcal{D}_H(\mathbf{W}_{\mathrm{gt}})+\alpha\,\mathcal{D}_H(\mathbf{W}_{\mathrm{pl}}),\qquad
		\mathbf{B}[i,d]=\mathbf{1}\!\big[\mathbf{W}[i,d]>\mu\cdot\median_j\mathbf{W}[j,d]\big].
		\label{eq:quant}
	\end{equation}
	In the unsupervised variant (\textbf{NyS-Binary Unsupervised}, where $k_{\mathrm{label}} = 0$), the semantic channel is omitted and the representation simplifies directly to $\mathbf{W} = \mathcal{D}_H(\mathbf{R}_{\mathrm{str}}) \in \mathbb{R}^{N \times K}$, which is quantized via column-wise thresholding $\mathbf{B}[i,d]=\mathbf{1}[\mathbf{W}[i,d]>\mu\cdot\median_j\mathbf{W}[j,d]]$ with zero label input.
	The per-column median makes each bit roughly balanced, which maximizes the
	information entropy of the discrete code. Throughout we use $K=250$, $H=3$, blend weight $\alpha=0.50$, quantile threshold multiplier $\mu=0.50$, confidence gate $\tau=0.50$, landmark budget $m=125$, and a balanced 50/50 bit split ($k_{\mathrm{struct}}=125, k_{\mathrm{label}}=125$). As demonstrated in the 4-classifier factorial interaction heatmaps across all 10 benchmarks (Figure~\ref{fig:alpha_tau_grid} in Appendix~\ref{sec:appendix_alpha_tau}), the combination $(\alpha=0.50, \tau=0.50)$ provides a highly stable, architecture-agnostic default across GCN, GraphSAGE, MLP, and LINKX. Establishing these defaults on an internal hyperparameter tuning split partitioned strictly from the training set, and keeping them strictly frozen for the 70/30 experiments in Table~\ref{tab:overall}, demonstrates cross-regime parameter stability.
	Figure~\ref{fig:arch} and Algorithm~\ref{alg:nys} summarize the pipeline.
	
	%
	%
	
	\usetikzlibrary{calc,positioning,shapes.geometric,arrows.meta,fit,backgrounds}
	
	\providecommand{\NySdefcolor}[3]{\definecolor{#1}{#2}{#3}}
	\NySdefcolor{bandgray}{RGB}{248, 249, 250}   
	\NySdefcolor{bandblue}{RGB}{241, 243, 244}   
	\NySdefcolor{bandyellow}{RGB}{248, 249, 250} 
	\NySdefcolor{bandsky}{RGB}{241, 243, 244}    
	\NySdefcolor{banddiff}{RGB}{248, 249, 250}   
	\NySdefcolor{bandblend}{RGB}{241, 243, 244}  
	\NySdefcolor{boxgreenbg}{RGB}{230, 244, 234}     
	\NySdefcolor{boxgreenborder}{RGB}{168, 218, 181} 
	\NySdefcolor{boxorangebg}{RGB}{254, 247, 224}    
	\NySdefcolor{boxorangeborder}{RGB}{253, 214, 99} 
	\NySdefcolor{boxpurplebg}{RGB}{252, 232, 230}    
	\NySdefcolor{boxpurpleborder}{RGB}{252, 175, 169}
	\NySdefcolor{boxbluebg}{RGB}{232, 240, 254}      
	\NySdefcolor{boxblueborder}{RGB}{174, 203, 250}  
	\NySdefcolor{boxskybg}{RGB}{232, 240, 254}       
	\NySdefcolor{boxskyborder}{RGB}{174, 203, 250}   
	\NySdefcolor{c1bg}{RGB}{230, 244, 234}           
	\NySdefcolor{c1border}{RGB}{129, 201, 149}       
	\NySdefcolor{c1box}{RGB}{92, 186, 113}           
	\NySdefcolor{c2bg}{RGB}{254, 247, 224}           
	\NySdefcolor{c2border}{RGB}{252, 214, 99}        
	\NySdefcolor{c2box}{RGB}{242, 153, 0}            
	\NySdefcolor{c2codebook}{RGB}{238, 103, 92}      
	\NySdefcolor{maroonbox}{RGB}{102, 157, 246}      
	
	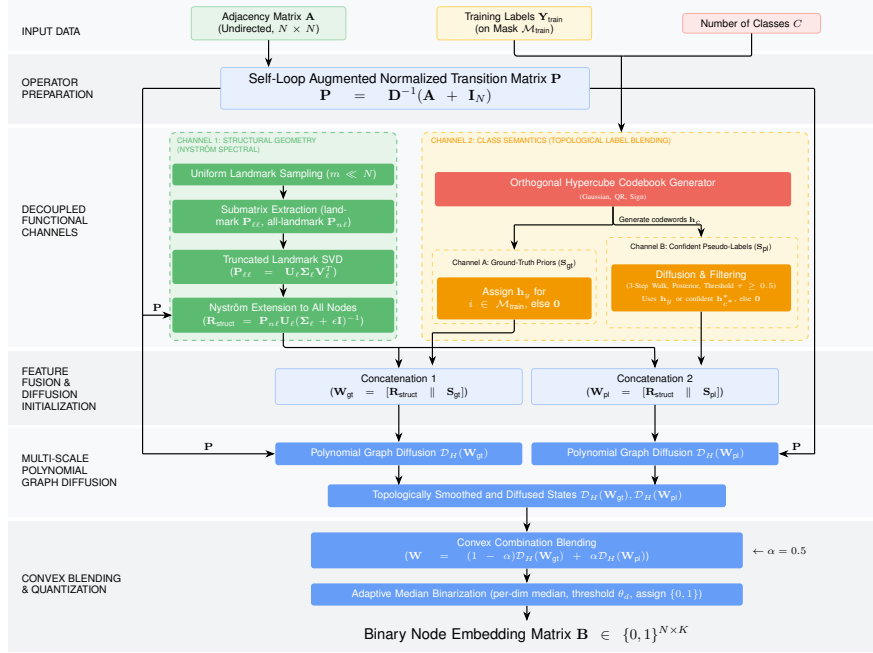
\begin{figure}[htbp!]
		\centering
		\resizebox{0.70\textwidth}{!}{%
			\begin{tikzpicture}[
				>=Stealth,
				x=1cm,
				y=1cm,
				font=\sffamily,
				action node/.style={
					align=center,
					rounded corners=2pt,
					font=\sffamily\scriptsize,
					inner sep=4pt
				},
				c1 node/.style={
					action node,
					fill=c1box,
					text=white,
					text width=5.0cm
				},
				maroon node/.style={
					action node,
					fill=maroonbox,
					text=white
				}
				]
				
				\begin{scope}[on background layer]
					\fill[bandgray]   (-11.2,0.2) rectangle (9.6,-1.1);
					\fill[bandblue]   (-11.2,-1.2) rectangle (9.6,-2.85);
					\fill[bandyellow] (-11.2,-2.95) rectangle (9.6,-8.15);
					\fill[bandsky]    (-11.2,-8.25) rectangle (9.6,-10.0);
					\fill[banddiff]   (-11.2,-10.1) rectangle (9.6,-12.2);
					\fill[bandblend]  (-11.2,-12.3) rectangle (9.6,-15.4);
				\end{scope}
				
				\node[anchor=west, font=\sffamily\scriptsize, align=left]
				at (-11.0,-0.65) {INPUT DATA};
				\node[anchor=west, font=\sffamily\scriptsize, align=left]
				at (-11.0,-2.0)  {OPERATOR\\PREPARATION};
				\node[anchor=west, font=\sffamily\scriptsize, align=left]
				at (-11.0,-5.15) {DECOUPLED\\FUNCTIONAL\\CHANNELS};
				\node[anchor=west, font=\sffamily\scriptsize, align=left]
				at (-11.0,-9.15) {FEATURE\\FUSION \&\\DIFFUSION \\INITIALIZATION};
				\node[anchor=west, font=\sffamily\scriptsize, align=left]
				at (-11.0,-11.1) {MULTI-SCALE\\POLYNOMIAL\\GRAPH DIFFUSION};
				\node[anchor=west, font=\sffamily\scriptsize, align=left]
				at (-11.0,-13.8) {CONVEX BLENDING\\\& QUANTIZATION};
				
				\node[action node, fill=boxgreenbg, draw=boxgreenborder, text=black,
				text width=3.4cm] (adj) at (-5.0,-0.45)
				{Adjacency Matrix $\mathbf{A}$\\
					\scriptsize (Undirected, $N \times N$)};
				\node[action node, fill=boxorangebg, draw=boxorangeborder, text=black,
				text width=3.4cm] (ytrain) at (0.85,-0.45)
				{Training Labels $\mathbf{Y}_{\text{train}}$\\
					\scriptsize (on Mask $\mathcal{M}_{\text{train}}$)};
				\node[action node, fill=boxpurplebg, draw=boxpurpleborder, text=black,
				text width=3.0cm] (nclasses) at (6.5,-0.45)
				{Number of Classes $C$};
				
				\node[action node, fill=boxbluebg, draw=boxblueborder, text=black,
				text width=8.5cm] (transP) at (-1.75,-2.0)
				{\small Self-Loop Augmented Normalized Transition Matrix
					$\mathbf{P}$\\
					\normalsize $\mathbf{P} = \mathbf{D}^{-1}(\mathbf{A}+\mathbf{I}_N)$};
				
				\draw[thick, dashed, draw=c1border, fill=c1bg, rounded corners=4pt]
				(-7.35,-3.05) rectangle (-1.95,-8);
				\node[anchor=north west, align=left, font=\sffamily\tiny, text=c1box]
				at (-7.3,-3.05)
				{CHANNEL 1: STRUCTURAL GEOMETRY \\ (NYSTR\"OM SPECTRAL)};
				\node[c1 node] (c1_1) at (-4.65,-4.05)
				{Uniform Landmark Sampling ($m \ll N$)};
				\node[c1 node] (c1_2) at (-4.65,-5.05)
				{Submatrix Extraction
					(landmark $\mathbf{P}_{\ell\ell}$, all-landmark $\mathbf{P}_{n\ell}$)};
				\node[c1 node] (c1_3) at (-4.65,-6.25)
				{Truncated Landmark SVD
					($\mathbf{P}_{\ell\ell} = \mathbf{U}_\ell\bm{\Sigma}_\ell\mathbf{V}_\ell^T$)};
				\node[c1 node] (c1_4) at (-4.65,-7.4)
				{Nystr\"om Extension to All Nodes\\[1pt]
					($\mathbf{R}_{\text{struct}} =
					\mathbf{P}_{n\ell}\mathbf{U}_\ell(\bm{\Sigma}_\ell+\epsilon\mathbf{I})^{-1}$)};
				
				\draw[thick, dashed, draw=c2border, fill=c2bg, rounded corners=4pt]
				(-1.35,-3.05) rectangle (7.85,-8);\node[anchor=north west, font=\sffamily\tiny, text=c2box]
				at (-1.25,-3.05)
				{CHANNEL 2: CLASS SEMANTICS (TOPOLOGICAL LABEL BLENDING)};
				
				\node[action node, fill=c2codebook, text=white, text width=8.2cm]
				(c2_cb) at (3.2,-4.4)
				{Orthogonal Hypercube Codebook Generator\\
					\vspace{1pt}\normalfont\tiny (Gaussian, QR, Sign)};
				
				\draw[dashed, thin, draw=c2border, rounded corners=2pt]
				(-1.1,-5.85) rectangle (2.8,-7.6);
				\node[font=\sffamily\tiny, text=black] at (0.85,-6.2)
				{Channel A: Ground-Truth Priors ($\mathbf{S}_{\text{gt}}$)};
				\node[action node, fill=c2box, text=white, text width=3.4cm,
				minimum height=1cm] (c2_a) at (0.85,-7.0)
				{Assign $\mathbf{h}_y$ for\\
					$i \in \mathcal{M}_{\text{train}}$, else $\mathbf{0}$};
				
				\draw[dashed, thin, draw=c2border, rounded corners=2pt]
				(3.05,-5.55) rectangle (7.6,-7.7);
				\node[font=\sffamily\tiny, text=black] at (5.3,-5.8)
				{Channel B: Confident Pseudo-Labels ($\mathbf{S}_{\text{pl}}$)};
				\node[action node, fill=c2box, text=white, text width=4.0cm,
				minimum height=1cm] (c2_b) at (5.3,-6.75)
				{Diffusion \& Filtering\\
					\vspace{1pt}\normalfont\tiny
					(3-Step Walk, Posterior, Threshold $\tau \ge 0.5$)\\
					\vspace{1pt} Uses $\mathbf{h}_y$ or confident
					$\mathbf{h}_{c^*}^*$, else $\mathbf{0}$};
				
				\node[action node, fill=boxskybg, draw=boxskyborder, text=black,
				text width=5.6cm] (fuse1) at (-1.9,-9.1)
				{Concatenation 1\\[1pt]
					($\mathbf{W}_{\text{gt}}
					= [\mathbf{R}_{\text{struct}}\parallel\mathbf{S}_{\text{gt}}]$)};
				\node[action node, fill=boxskybg, draw=boxskyborder, text=black,
				text width=5.6cm] (fuse2) at (4.2,-9.1)
				{Concatenation 2\\[1pt]
					($\mathbf{W}_{\text{pl}}
					= [\mathbf{R}_{\text{struct}}\parallel\mathbf{S}_{\text{pl}}]$)};
				
				\node[maroon node, text width=5.6cm] (diff1) at (-1.9,-10.7)
				{Polynomial Graph Diffusion
					$\mathcal{D}_H(\mathbf{W}_{\text{gt}})$};
				\node[maroon node, text width=5.6cm] (diff2) at (4.2,-10.7)
				{Polynomial Graph Diffusion
					$\mathcal{D}_H(\mathbf{W}_{\text{pl}})$};
				\node[maroon node, text width=9.2cm] (diff_states) at (1.15,-11.7)
				{Topologically Smoothed and Diffused States
					$\mathcal{D}_H(\mathbf{W}_{\text{gt}}), \mathcal{D}_H(\mathbf{W}_{\text{pl}})$};
				
				\node[maroon node, text width=10cm] (blend) at (1.15,-13.0)
				{Convex Combination Blending\\[1pt]
					($\mathbf{W}
					= (1-\alpha)\mathcal{D}_H(\mathbf{W}_{\text{gt}})
					+ \alpha\mathcal{D}_H(\mathbf{W}_{\text{pl}})$)};
				\node[right=0.1cm of blend, font=\sffamily\scriptsize,
				text=black] {$\leftarrow \alpha=0.5$};
				\node[maroon node, text width=10cm] (quant) at (1.15,-14.05)
				{Adaptive Median Binarization
					(per-dim median, threshold $\theta_d$, assign $\{0,1\}$)};
				\node[text=black, text width=10.5cm, align=center,
				font=\sffamily\normalsize] (out) at (1.15,-15.0)
				{Binary Node Embedding Matrix
					$\mathbf{B}\in\{0,1\}^{N\times K}$};
				
				\draw[thick,->] (adj.south) -- (adj.south |- transP.north);
				\draw[thick,->] (3.4, -1.23) -- (3.4,-3.05);
				\draw[thick,rounded corners=2pt]
				(ytrain.south) -- (0.85,-1.23) -- (3.45,-1.23);
				\draw[thick,rounded corners=1pt] (nclasses.south) -- (6.5,-1.23) -- (3.45,-1.23) ;
				\draw[thick,->] (c1_1.south) -- (c1_2.north);
				\draw[thick,->] (c1_2.south) -- (c1_3.north);
				\draw[thick,->] (c1_3.south) -- (c1_4.north);
				\draw[thick,rounded corners=1pt]
				(c2_cb.south) -- (3.2,-5.25) -- (0.85,-5.25);
				\draw[thick,->] (0.85,-5.25) -- (0.85,-5.85);
				\draw[thick,rounded corners=2pt]
				(c2_cb.south) -- (3.2,-5.25) -- (5.3,-5.25);
				\draw[thick,->] ((5.3,-5.25) -- (5.3,-5.55);
				\node[font=\sffamily\tiny, align=left, anchor=west] at (3.22,-5.14)
				{Generate codewords $\mathbf{h}_c$};
				\draw[thick,rounded corners=2pt] (c2_a.south) -- (0.85,-7.82) -- (-1.1,-7.82);
				\draw[thick,->] (-1.1,-7.82) -- (-1.1,-8.6) ;
				\draw[thick,->] (c2_b.south) -- (5.3,-8.6) ;
				\draw[thick,->] (-1.9,-8.17) -- (fuse1.north);
				\draw[thick,rounded corners=2pt]
				(c1_4.south) -- (-4.65,-8.17) -- (4.19,-8.17);
				\draw[thick,->] (4.19,-8.17) -- (fuse2.north);
				\draw[thick,->] (fuse1.south) -- (diff1.north);
				\draw[thick,->] (fuse2.south) -- (diff2.north);
				\draw[thick,->] (diff1.south) -- (-1.9,-11.45);
				\draw[thick,->] (diff2.south) -- (4.2,-11.45);
				\draw[thick,->] (diff_states.south) -- (blend.north);
				\draw[thick,->] (blend.south) -- (quant.north);
				\draw[thick,->] (quant.south) -- (out.north);
				
				\draw[thick,rounded corners=1pt]
				(transP.west) -- (-8,-2.0) -- (-8,-10.7);
				\draw[thick,->] (-8,-7.4) -- (c1_4.west)
				node[midway, above=-1pt, font=\sffamily\scriptsize]
				{$\mathbf{P}$};
				\draw[thick,->] (-8,-10.7) -- (diff1.west)
				node[midway, above=-1pt, font=\sffamily\scriptsize]
				{$\mathbf{P}$};
				\draw[thick,rounded corners=1pt]
				(transP.east) -- (8.0,-2.0) -- (8.0,-10.7);
				\draw[thick,->] (8.0,-10.7) -- (diff2.east)
				node[midway, above=-1pt, font=\sffamily\scriptsize]
				{$\mathbf{P}$};
				
			\end{tikzpicture}%
		}
		\caption{The NyS-Binary pipeline. A row-stochastic transition operator feeds
			a Nystr\"om structural channel and a leakage-safe semantic channel; the two
			are diffused, convexly blended, and quantized by an adaptive per-column
			median rule.}
		\label{fig:arch}
	\end{figure}
	
	\begin{algorithm}[htbp!]
		\caption{NyS-Binary: closed-form binary graph hashing}
		\label{alg:nys}
		\begin{algorithmic}[1]
			\Require Adjacency $\mathbf{A} \in \{0,1\}^{N \times N}$; training labels on $\mathcal{M}_{\mathrm{tr}}$ with labels $y_i$; code length $K$; hyper-parameters $(H, \alpha, \mu, \tau, m, \epsilon)$ with defaults $(H=3, \alpha=0.50, \mu=0.50, \tau=0.50, m=125, \epsilon=10^{-10})$.
			\Ensure Binary node codes $\mathbf{B}\in\{0,1\}^{N\times K}$.
			
			\Statex \textbf{\textit{Phase 1: Graph Normalization \& Bit Allocation}}
			\State $\mathbf{P} \gets \tilde{\mathbf{D}}^{-1}(\mathbf{A}+\mathbf{I}_N)$ \Comment{Normalized random-walk matrix}
			\State $k_{\mathrm{struct}} \gets \lfloor K/2\rfloor$,\ $k_{\mathrm{label}} \gets K-k_{\mathrm{struct}}$ \Comment{Budget split (Structure vs Label)}
			
			\Statex \textbf{\textit{Phase 2: Structural Channel (Nystr\"om-Inspired Approximation)}}
			\State $m \gets \min(N, \max(k_{\mathrm{struct}}, 125))$ \Comment{Determine landmark budget ($m=125$)}
			\State Sample landmark nodes $\mathcal{L} \subset \mathcal{V}$ uniformly at random, where $|\mathcal{L}|=m$
			\State $\mathbf{U}_\ell, \bm{\Sigma}_\ell \gets \text{trunc-SVD}_{k_{\mathrm{struct}}}(\mathbf{P}[\mathcal{L},\mathcal{L}])$ \Comment{Eigendecomposition of landmark core}
			\State $\mathbf{R}_{\mathrm{str}} \gets \mathbf{P}[:,\mathcal{L}]\,\mathbf{U}_\ell(\bm{\Sigma}_\ell+\epsilon\mathbf{I})^{-1}$ \Comment{Project nodes via Nystr\"om-inspired extension Eq.~\ref{eq:nystrom}}
			
			\Statex \textbf{\textit{Phase 3: Semantic Channel (Label Codebook \& Diffusion)}}
			\State Generate orthogonal codebook $\{\bm{h}_c\}_{c=1}^C \in \{-1,1\}^{k_{\mathrm{label}}}$ via Gaussian QR-decomposition
			\State Init ground-truth $\mathbf{S}_{\mathrm{gt}} \in \mathbb{R}^{N \times k_{\mathrm{label}}}$: $\mathbf{S}_{\mathrm{gt}}[i,:] \gets \bm{h}_{y_i}$ if $i \in \mathcal{M}_{\mathrm{tr}}$, else $\mathbf{0}$
			\State $\mathbf{Y}_t \gets \mathbf{P}^3\mathbf{Y}_0$ \Comment{Diffuse raw label one-hot distribution}
			\State Init pseudo-label $\mathbf{S}_{\mathrm{pl}} \in \mathbb{R}^{N \times k_{\mathrm{label}}}$: $\mathbf{S}_{\mathrm{pl}}[i,:] \gets \bm{h}_{\hat{y}_i}$ if confidence $\hat{p} \ge \tau$, else $\mathbf{0}$
			
			\Statex \textbf{\textit{Phase 4: Multi-Channel Blending \& Quantization}}
			\State $\mathbf{E}_{\mathrm{gt}} \gets \mathcal{D}_H([\mathbf{R}_{\mathrm{str}} \| \mathbf{S}_{\mathrm{gt}}])$ \Comment{Diffuse concatenated ground-truth representation}
			\State $\mathbf{E}_{\mathrm{pl}} \gets \mathcal{D}_H([\mathbf{R}_{\mathrm{str}} \| \mathbf{S}_{\mathrm{pl}}])$ \Comment{Diffuse concatenated pseudo-label representation}
			\State $\mathbf{W} \gets (1{-}\alpha)\mathbf{E}_{\mathrm{gt}} + \alpha\,\mathbf{E}_{\mathrm{pl}}$ \Comment{Convex combination of certainty and coverage}
			\State \Return $\mathbf{B}[i,d]=\mathbf{1}[\mathbf{W}[i,d] > \mu\cdot\median_j\mathbf{W}[j,d]]$ \Comment{Adaptive quantile thresholding Eq.~\ref{eq:quant}}
		\end{algorithmic}
	\end{algorithm}
	
	\subsection{Complexity and memory}
	For fixed landmark count $m$ and code length $K$, construction is $O(m^3 + NmK + |E|KH)$: the landmark core decomposition, out-of-sample projection, and sparse polynomial diffusion. Under standard real-world graph sparsity ($|E| = O(N)$), construction scales near-linearly as $O(N)$ (vs.\ exact spectral methods). Storage is $NK$ bits, a $32\times$ reduction over $N\times K$ \texttt{float32} vectors.
	
	\subsection{Spiking Classifier Architectures and Graph Adaptations}
	\label{sec:appendix_spiking_classifiers}
	
	In this section, we describe the neuromorphic spiking neural network classifiers evaluated throughout our experiments, specifically detailing the exact architectural adaptations made to extend the original Spiking Agreement-Driven Plasticity (SADP) model to graph topologies, the mechanics of the two-way variant, and the surrogate-gradient baseline.
	
	\subsubsection{Graph Adaptation of Spiking Agreement-Driven Plasticity (SADP)}
	The original SADP formulation \citep{s2026building} was developed for non-relational, Euclidean data (e.g., vision and tabular benchmarks) using a standard multilayer perceptron (MLP) topology. To evaluate our binary representations in a biologically plausible, gradient-free neuromorphic setting on relational data, we extended the original model to graph topologies through two fundamental adaptations:
	\begin{enumerate}[leftmargin=*,itemsep=2.5pt]
		\item \textbf{Spiking Message Passing Forward Dynamics:} In the original MLP formulation, incoming Poisson spike trains $\mathbf{S}_{\mathrm{in}}[t]$ are projected directly via dense synaptic weights $\mathbf{I}_h[t] = \mathbf{S}_{\mathrm{in}}[t] \mathbf{W}_1$. In our graph extension, we substitute this with spiking neighborhood message passing:
		\begin{equation}
			\mathbf{I}_h[t] = \big(\hat{\mathbf{A}} \, \mathbf{S}_{\mathrm{in}}[t]\big) \, \mathbf{W}_1, \quad \text{where } \hat{\mathbf{A}} = \tilde{\mathbf{D}}^{-1/2}\tilde{\mathbf{A}}\tilde{\mathbf{D}}^{-1/2}.
		\end{equation}
		Because the message aggregation is linear, neighborhood spikes are accumulated across graph edges prior to the synaptic transformation. A two-layer message-passing core is employed, while the final class readout remains a purely local, per-node linear transformation without graph aggregation ($\mathbf{I}_o[t] = \mathbf{S}_{h2}[t] \mathbf{W}_2$). Membrane potentials obey discrete-time leaky integrate-and-fire (LIF) dynamics with reset-by-subtraction, and class predictions are decoded from accumulated output spike counts.
		\item \textbf{Presynaptic Aggregation in Agreement Plasticity:} The core SADP synaptic update is driven by Cohen's kappa agreement $\kappa$ between post-synaptic spike trains and supervisory teacher spike trains generated from training labels, avoiding global backpropagation. In our graph extension, the presynaptic factor in the outer-product update is adapted to match the aggregated input actually experienced by the synapses:
		\begin{equation}
			\Delta \mathbf{W}_1 \propto \sum_{d=-k_{\mathrm{shift}}}^{k_{\mathrm{shift}}} \big(\hat{\mathbf{A}} \mathbf{X}\big)^\top \bm{\kappa}_d.
		\end{equation}
		This substitution accumulates shared-weight contributions across edges for each node. Crucially, the temporal shift parameter $d \in [-k_{\mathrm{shift}}, k_{\mathrm{shift}}]$ in the Cohen's kappa calculation naturally models multi-hop propagation delays: spikes traversing multi-hop neighborhood paths arrive with temporal latency, and the shifted agreement window aligns synaptic updates with these delayed causal arrivals.
	\end{enumerate}
	
	\subsubsection{Two-Way SADP Classification (SADP Two-Way)}
	\label{sec:appendix_two_way}
	To evaluate the intrinsic linear separability of binary node codes in a neuromorphic substrate without the confounding effects of multi-class competition, we introduced the two-way variant:
	\begin{itemize}[leftmargin=*,itemsep=2pt]
		\item \textbf{Subpopulation and Class Filtering:} Rather than training an output neuron for every class in a $C$-class dataset, the model isolates the \textbf{top-two most frequent classes} in the ground truth: $\mathcal{C}_{\mathrm{top2}} = \{c_{(1)}, c_{(2)}\}$. All nodes belonging to any other class are masked out from both the training and test splits. The remaining node labels are remapped to binary targets $\{0, 1\}$.
		\item \textbf{Network Configuration:} The output layer is instantiated with exactly two LIF output neurons, maintaining the identical two-layer spiking message-passing core and gradient-free Cohen's kappa learning rule.
		\item \textbf{Analytical Rationale for Performance Gap:} In multi-class spiking networks with gradient-free learning, lateral inhibition and uncoordinated spike collisions across many competing output neurons often induce class confusion (particularly on benchmarks with 7 to 40 classes). Restricting evaluation to the two dominant classes eliminates multi-class interference, directly testing whether the binary Hamming codes form cleanly separable clusters under biologically plausible spike dynamics. This explains why SADP Two-Way achieves substantially higher accuracy across all methods (reaching $83.57\%$ on NyS-Binary, compared to $62.10\%$ for full multi-class SADP).
	\end{itemize}
	
	\subsubsection{Surrogate-Gradient Spiking Network Baseline (SADP Surrogate)}
	To directly benchmark gradient-free agreement-driven learning against standard gradient-based optimization in spiking networks, we implemented a surrogate-gradient baseline \citep{neftci2019surrogate}:
	\begin{itemize}[leftmargin=*,itemsep=2pt]
		\item \textbf{Matched Forward Pass:} Employs the identical spiking message-passing architecture, Poisson feature encoding, and LIF membrane dynamics as the graph SADP model, ensuring that any performance gap is strictly attributable to the learning rule rather than the network capacity.
		\item \textbf{Surrogate Derivative Substitution:} During the backward pass, the non-differentiable Dirac delta derivative of the Heaviside spike function is replaced by a smooth sigmoid pseudo-derivative $\sigma'(u) = \beta s (1 - s)$, where $s = 1/(1 + \exp(-\beta u))$ with sharpness $\beta = 10.0$.
		\item \textbf{BPTT Optimization:} The model accumulates output logits via a leaky-integrator readout and optimizes masked cross-entropy via Backpropagation Through Time (BPTT) using the Adam optimizer.
	\end{itemize}
	
	\section{Experiments}
	\label{sec:experiments}
	\subsection{Datasets, classifiers, and protocol}
	We evaluate on ten benchmarks: four homophilous graphs (Cora, CiteSeer, PubMed, DBLP \citep{sen2008planetoid,yang2012defining}); WikiCS \citep{mernyei2020wikics}; two co-purchase graphs (Photo, Computers \citep{shchur2018pitfalls}); and three heterophilous graphs (Chameleon, Squirrel, Amazon-Ratings \citep{pei2020geomgcn,platonov2023critical}). For DBLP, we adopt the standard 2-class benchmark \citep{yang2019nodesketch} from SNAP (where multi-class and Two-Way SADP coincide). Dataset statistics appear in Table~\ref{tab:dataset_statistics} (Section~\ref{sec:appendix_datasets}).
	
	Every embedding is evaluated across nine classifiers spanning linear probes (Logistic Regression), deep networks (MLP), standard GNNs (GCN \citep{kipf2017gcn}, GraphSAGE \citep{hamilton2017graphsage}), heterophily-tailored architectures (H2GCN \citep{zhu2020h2gcn}, LINKX \citep{lim2021linkx}), and neuromorphic spiking networks. For neuromorphic deployment, we benchmark three spiking (SNN) probes: gradient-free agreement-driven synaptic learning (SADP) \citep{s2026building}, a two-class variant (SADP Two-Way, Section~\ref{sec:appendix_two_way}), and a surrogate-gradient network (SADP-Surrogate) \citep{neftci2019surrogate,maass1997snn}. Discrete bitcodes natively match event-driven spike accumulations without lossy continuous rate coding. We use a 70/30 stratified split over seeds $\{42, 123, 77\}$. Hyperparameters $(\alpha=0.50, \tau=0.50)$ were selected via grid search on an internal tuning split partitioned from the training pool (Figure~\ref{fig:alpha_tau_grid}) and frozen across all benchmarks. Codes are 250 bits. All runs use a laptop CPU (i7-11800H) and RTX~3050~Ti GPU.
	
	We evaluate two configurations: (i)~\textbf{NyS-Binary (Ours)} with a 50/50 structural-semantic blend ($k_{\mathrm{struct}}=125, k_{\mathrm{label}}=125$), and (ii)~\textbf{NyS-Binary (Unsupervised)} with $\text{struct\_ratio}=1.0$ ($k_{\mathrm{struct}}=250, k_{\mathrm{label}}=0$), generating codes strictly from topology. We benchmark against three feature-free hashing baselines (NodeSig \citep{celikkanat2022nodesig}, NodeSketch \citep{yang2019nodesketch}, node2binary \citep{talukder2025node2binary}) and three attribute-aware references (Bi-GCN \citep{wang2021bigcn}, continuous raw features, binarized raw features).
	
	\subsection{Dataset Summary Statistics}
	\label{sec:appendix_datasets}
	
	Table~\ref{tab:dataset_statistics} summarizes the key topological characteristics of the primary benchmark graph datasets used across our experiments. These include seven homophilous and three heterophilous networks spanning diverse domains such as citation networks, co-authorship graphs, co-purchase graphs, and web-page hyperlink networks. The table details the number of nodes, edges, target classes, and the edge homophily ratio for each graph, which directly influence the behavior of the structural and semantic diffusion mechanisms in NyS-Binary.
	
	\vspace{0.8em}
	\noindent
	\begin{center}
		\begin{minipage}{\linewidth}
			\centering
			\small
			\captionof{table}{Summary statistics of the primary benchmark graph datasets evaluated across our experiments, detailing the number of nodes ($N$), edges ($E$), target classes ($C$), edge homophily ratio ($h = \frac{|\{(u,v) \in \mathcal{E} : y_u = y_v\}|}{|\mathcal{E}|}$), and network domain, spanning seven homophilous and three heterophilous topologies. For \textit{Chameleon} and \textit{Squirrel}, we utilize the standard Geom-GCN version and splits \citep{pei2020geomgcn} (with 31,371 and 198,353 directed edges respectively), maintaining these specific versions for direct comparability with prior hashing literature despite their known duplicate-node overlaps \citep{platonov2023critical}.}
			\label{tab:dataset_statistics}
			\resizebox{0.92\textwidth}{!}{%
				\begin{tabular}{llrrcrl}
					\toprule
					\textbf{Dataset} & \textbf{Domain} & \textbf{Nodes ($N$)} & \textbf{Edges ($E$)} & \textbf{Classes ($C$)} & \textbf{Homophily ($h$)} & \textbf{Topology} \\
					\midrule
					Cora            & Citation network     & 2,708  & 5,278   & 7  & 0.81 & Homophilous \\
					CiteSeer        & Citation network     & 3,327  & 4,552   & 6  & 0.74 & Homophilous \\
					PubMed          & Citation network     & 19,717 & 44,324  & 3  & 0.80 & Homophilous \\
					WikiCS          & Hyperlink network    & 11,701 & 215,863 & 10 & 0.65 & Homophilous \\
					DBLP            & Co-authorship graph  & 13,326 & 34,281  & 2  & 0.83 & Homophilous \\
					Photo           & Amazon co-purchase   & 7,650  & 119,081 & 8  & 0.83 & Homophilous \\
					Computers       & Amazon co-purchase   & 13,752 & 245,861 & 10 & 0.78 & Homophilous \\
					Chameleon       & Wikipedia page-page  & 2,277  & 31,371  & 5  & 0.23 & Heterophilous \\
					Squirrel        & Wikipedia page-page  & 5,201  & 198,353 & 5  & 0.22 & Heterophilous \\
					Amazon-Ratings  & Product review graph & 24,492 & 93,050  & 5  & 0.38 & Heterophilous \\
					\bottomrule
				\end{tabular}%
			}
		\end{minipage}
	\end{center}
	\vspace{0.8em}

	\subsection{Overall accuracy}
	Table~\ref{tab:overall} aggregates accuracy across all ten benchmark datasets and three seeds. NyS-Binary (Ours) achieves the highest performance among all binary methods across every single one of the nine classifiers, establishing substantial margins on non-relational probes ($+9.41\%$ on MLP and $+9.97\%$ on Logistic Regression over NodeSig) and neuromorphic spiking classifiers ($+26.81\%$ on SADP and $+29.29\%$ on SADP-Surrogate over NodeSig; $+9.45\%$ over learned Bi-GCN on SADP).
	
	In the strictly label-free setting, \textbf{NyS-Binary (Unsupervised)} ($k_{\mathrm{struct}}=250, k_{\mathrm{label}}=0$) provides an exact like-for-like comparison against unsupervised binary hashing baselines (NodeSig, NodeSketch, node2binary). Purely from graph topology with zero label inputs, it consistently secures the second-highest accuracy across all nine probes ($72.89\%$ on GCN, $75.19\%$ on H2GCN, $73.18\%$ on MLP), leading NodeSig by $+7.20\%$ on MLP and $+9.99\%$ on SADP, and NodeSketch by $+20.71\%$ on MLP. Layering the label channel in \textbf{NyS-Binary (Ours)} integrates closed-form label diffusion into a static binary code; unlike transductive algorithms like C\&S \citep{huang2020combining} that output soft posteriors, NyS-Binary binarizes diffused signals into hardware-friendly bitcodes that serve as general-purpose features without backpropagation. Our low-supervision ablation (Section~\ref{sec:appendix_low_supervision}) confirms that NyS-Binary dominates even under sparse labels ($68.45\%$ on GCN, $67.95\%$ on LogReg vs.\ $64.17\%$ and $57.24\%$ for NodeSig; $0.37$s vs.\ $4.50$s for Bi-GCN). On datasets where top methods fall within one standard deviation (Section~\ref{sec:appendix_full_results}), differences lie within noise margins; the strength of NyS-Binary is systemic consistency across all 10 benchmarks, 9 diverse probe families, and sub-second generation efficiency.
	\begin{table}[htbp]
		\centering
		\caption{Overall mean accuracy (\%) across all ten
			datasets and three seeds. Binary hashing methods are grouped above the rule;
			attribute-aware references (Bi-GCN, raw features) utilize continuous node attributes and gradient training (or raw feature baselines) below the inner rule.
			Best binary result per column in \textbf{bold}; second-best \underline{underlined}. SADP Two-Way evaluates the two
			most frequent classes.}
		\label{tab:overall}
		\resizebox{\textwidth}{!}{%
			\begin{tabular}{lrrrrrrrrr}
				\toprule
				Method & GCN & SAGE & H2GCN & LINKX & LogReg & MLP & SADP & SADP-Surr & SADP-2W \\
				\midrule
				\textbf{NyS-Binary (Ours)} & \textbf{74.65} & \textbf{75.89} & \textbf{76.53} & \textbf{78.26} & \textbf{74.02} & \textbf{75.39} & \textbf{62.10} & \textbf{69.60} & \textbf{83.57} \\
				NyS-Binary (Unsupervised) & \underline{72.89} & \underline{73.99} & \underline{75.19} & \underline{75.31} & \underline{70.56} & \underline{73.18} & \underline{45.28} & \underline{59.53} & \underline{73.73} \\
				NodeSig & 70.68 & 71.54 & 71.56 & 70.72 & 64.05 & 65.98 & 35.29 & 40.31 & 64.34 \\
				NodeSketch & 66.73 & 68.12 & 68.00 & 72.58 & 51.08 & 52.47 & 30.47 & 29.57 & 61.67 \\
				node2binary & 57.41 & 59.88 & 57.52 & 72.64 & 44.68 & 46.69 & 20.66 & 44.51 & 57.51 \\
				\midrule
				Bi-GCN \scriptsize(attr.)                  & 74.03 & 75.28 & 74.60 & 74.88 & 71.70 & 72.08 & 52.65 & 55.83 & 77.34 \\
				Raw Features \scriptsize(attr.)           & 76.01 & 76.04 & 74.02 & 72.22 & 68.09 & 69.45 & 29.13 & 50.76 & 58.70 \\
				Raw Features Bin \scriptsize(attr.)       & 71.01 & 69.40 & 67.14 & 68.14 & 53.90 & 51.48 & 26.35 & 51.47 & 58.50 \\
				\bottomrule
			\end{tabular}%
		}
	\end{table}

	\subsection{Generation time}
\begin{figure}[htbp]
		\centering
		\begin{minipage}{0.49\textwidth}
			\includegraphics[width=\textwidth]{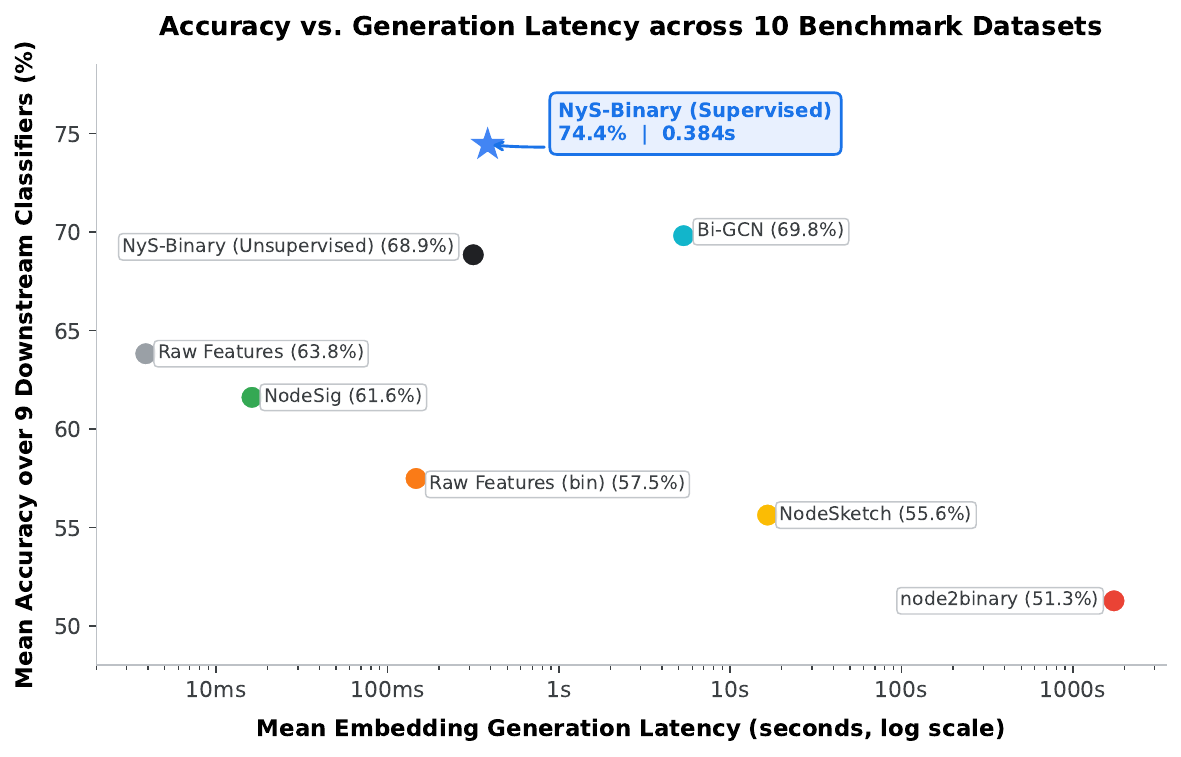}
			\caption{Accuracy vs.\ generation latency (log scale).}
			\label{fig:pareto}
		\end{minipage}\hfill
		\begin{minipage}{0.49\textwidth}
			\includegraphics[width=\textwidth]{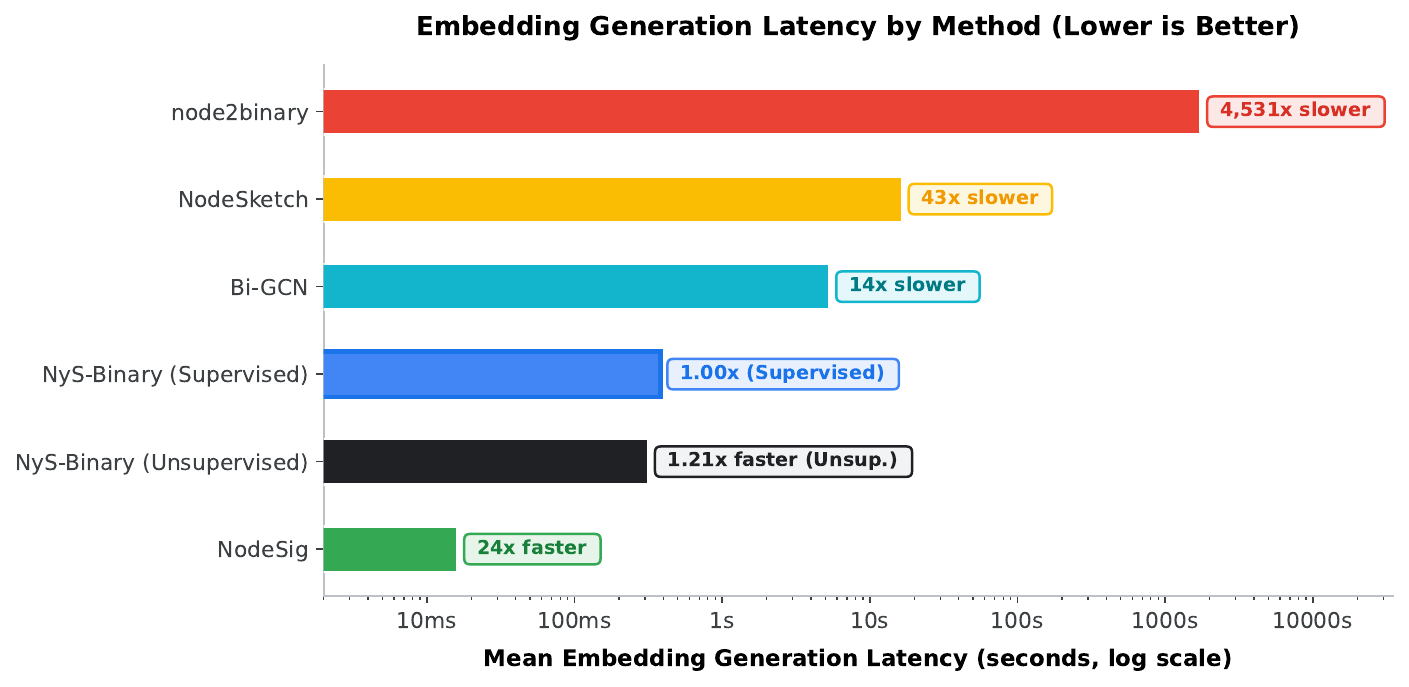}
			\caption{Generation time by method (log scale).}
			\label{fig:runtime}
		\end{minipage}
	\end{figure}

	Table~\ref{tab:runtime} reports latency. NyS-Binary generates $250$-bit codes in
	$0.38$~s on average. NodeSig is faster in absolute terms, but Table~\ref{tab:overall}
	shows it trails badly in accuracy; learned graph neural baselines are over an order of magnitude slower ($13.9\times$ for Bi-GCN), while hierarchical tree-based clustering methods are up to three orders of magnitude slower (e.g., $1739$s for node2binary on average). Figure~\ref{fig:pareto} plots the two axes
	together: NyS-Binary sits in the desirable upper-left region, and
	Figure~\ref{fig:runtime} shows the same latencies as bars on a log scale.
	
	\begin{table}[htbp]
		\centering
		\small
		\caption{Embedding generation latency across ten datasets (30 runs per method).
			NyS-Binary generates full $250$-bit codes in about half a second; NodeSig is
			faster but much less accurate, while learned and recursive methods are far
			slower.}
		\label{tab:runtime}
		\begin{tabular}{lrrrrc}
			\toprule
			Method & Mean $\pm$ Std (s) & Median (s) & Min (s) & Max (s) & vs.\ NyS-Binary \\
			\midrule
			NodeSig                & $0.016 \pm 0.010$ & $0.014$ & $0.006$ & $0.041$ & $0.04\times$ (fastest) \\
			NyS-Binary (Unsupervised) & $0.317 \pm 0.203$ & $0.342$ & $0.071$ & $0.702$ & $0.82\times$ (faster) \\
			\textbf{NyS-Binary (Ours)} & $\mathbf{0.384 \pm 0.363}$ & $\mathbf{0.335}$ & $\mathbf{0.033}$ & $\mathbf{1.245}$ & $\mathbf{1.00\times}$ (baseline) \\
			Bi-GCN                 & $5.351 \pm 3.492$ & $4.677$ & $1.290$ & $11.139$ & $13.9\times$ slower \\
			NodeSketch             & $16.543 \pm 9.026$ & $15.166$ & $8.164$ & $33.848$ & $43.1\times$ slower \\
			node2binary            & $1739.9 \pm 281.2$ & $1821.9$ & $972.3$ & $1946.5$ & $4530.9\times$ slower \\
			\bottomrule
		\end{tabular}
	\end{table}
	
	\begin{figure}[htbp]
		\centering
		\includegraphics[width=0.85\textwidth]{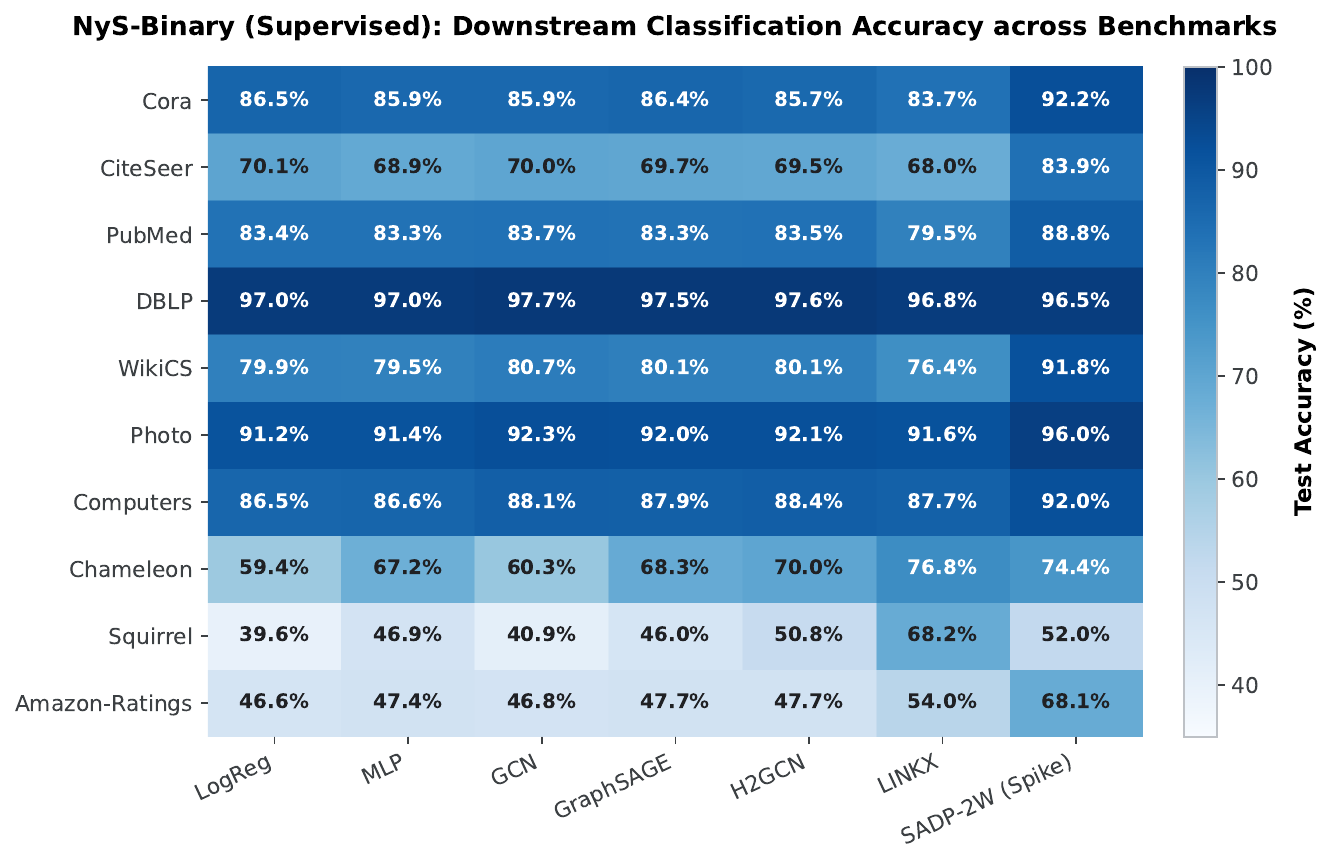}
		\caption{NyS-Binary accuracy per dataset and classifier. Homophilous graphs
			(top rows) are handled well across the board; heterophilous graphs (bottom rows)
			are harder, especially for the spiking classifier.}
		\label{fig:heat}
	\end{figure}
	
	\subsection{Per-dataset results}
	\label{sec:appendix_full_results}
	Table~\ref{tab:results_acc_std_all} 
	give a comprehensive per-dataset view across all ten benchmarks. The pattern is consistent: NyS-Binary dominates on homophilous graphs across nearly all classifiers, stays close to the best on PubMed, and is competitive but no longer dominant on heterophilous graphs like Squirrel, where structural signal is weaker. The heatmap in Figure~\ref{fig:heat} shows the full
	per-dataset, per-classifier accuracy of NyS-Binary at a glance.

	\vspace{0.8em}
	\noindent
	\begin{center}
		\begin{minipage}{\linewidth}
			\centering
			\captionof{table}{Comprehensive per-dataset classification accuracy $\pm$ std (\%) across benchmark datasets (Part 1: Amazon-Ratings, Chameleon, CiteSeer, Computers, and Cora) evaluated across nine downstream probe classifiers. Feature-free binary hashing methods are grouped above the inner rule; attribute-aware baselines are shown below. Best binary result per column in \textbf{bold}; second-best \underline{underlined}.}
			\label{tab:results_acc_std_all}
			\label{tab:results_acc_std_cora}
			{\small\renewcommand{\arraystretch}{1.12}\setlength{\tabcolsep}{2.4pt}
				\makebox[\textwidth][c]{%
					\resizebox{\textwidth}{!}{%
						\begin{tabular}{lrrrrrrrrr}
							\toprule
							Method & GCN & SAGE & H2GCN & LINKX & LogReg & MLP & SADP & SADP-Surr & SADP-2W \\
							\midrule
							\multicolumn{10}{l}{\textbf{Amazon-Ratings} (Heterophilous)} \\
							\midrule
							\textbf{NyS-Binary (Ours)} & \textbf{46.80}\std{0.37} & \textbf{47.67}\std{0.81} & \textbf{47.68}\std{0.23} & \textbf{53.98}\std{0.85} & \textbf{46.58}\std{0.40} & \textbf{47.42}\std{0.63} & \textbf{44.38}\std{0.75} & \textbf{39.77}\std{8.10} & \textbf{68.14}\std{0.83} \\
							NyS-Binary (Unsupervised) & \underline{45.34}\std{0.93} & \underline{45.15}\std{0.21} & \underline{45.77}\std{0.64} & \underline{53.83}\std{0.28} & \underline{42.22}\std{0.66} & \underline{45.22}\std{0.96} & 32.44\std{1.60} & \underline{36.90}\std{0.10} & \underline{57.38}\std{0.82} \\
							NodeSig & 40.65\std{0.77} & 39.83\std{1.13} & 41.22\std{1.13} & 52.49\std{0.41} & 37.96\std{0.47} & 40.14\std{0.83} & 32.25\std{0.28} & 36.76\std{0.04} & 56.76\std{0.51} \\
							NodeSketch & 39.43\std{1.47} & 37.31\std{0.73} & 38.39\std{1.24} & 52.62\std{0.38} & 36.28\std{0.47} & 36.22\std{1.25} & \underline{32.83}\std{0.20} & 36.55\std{0.34} & 57.07\std{0.49} \\
							node2binary & 42.20\std{0.50} & 42.28\std{0.52} & 41.57\std{0.76} & 52.91\std{0.53} & 37.38\std{0.63} & 39.57\std{0.98} & 25.94\std{0.40} & 33.70\std{1.05} & 57.17\std{0.12} \\
							\cmidrule(lr){1-10}
							Bi-GCN                  & 42.30\std{0.59} & 42.83\std{0.89} & 42.85\std{0.44} & 51.26\std{1.58} & 40.57\std{0.52} & 40.43\std{0.39} & 36.52\std{1.84} & 36.52\std{0.21} & 58.54\std{0.34} \\
							Raw Features            & 42.22\std{0.57} & 44.47\std{0.55} & 43.94\std{0.36} & 53.60\std{0.09} & 38.60\std{0.11} & 40.90\std{0.20} & 35.72\std{0.18} & 29.85\std{6.35} & 57.84\std{0.03} \\
							Raw Features Bin        & 38.36\std{0.31} & 36.98\std{0.58} & 36.95\std{0.35} & 52.09\std{0.53} & 34.04\std{0.09} & 30.48\std{1.63} & 33.62\std{0.64} & 31.05\std{4.87} & 57.13\std{0.41} \\
							\midrule
							\multicolumn{10}{l}{\textbf{Chameleon} (Heterophilous)} \\
							\midrule
							\textbf{NyS-Binary (Ours)} & 60.33\std{0.51} & \underline{68.27}\std{1.80} & \underline{69.98}\std{1.91} & \underline{76.80}\std{1.69} & 59.41\std{0.34} & \underline{67.15}\std{1.41} & \textbf{31.63}\std{4.20} & \textbf{45.32}\std{7.31} & 74.35\std{5.45} \\
							NyS-Binary (Unsupervised) & 59.94\std{1.83} & \textbf{68.52}\std{1.31} & \textbf{73.00}\std{0.81} & 74.90\std{5.68} & \textbf{63.55}\std{0.80} & \textbf{70.76}\std{1.34} & 26.07\std{0.99} & \underline{42.84}\std{8.28} & \textbf{75.48}\std{4.07} \\
							NodeSig & \underline{61.60}\std{1.55} & 67.50\std{0.88} & 69.44\std{1.25} & 76.22\std{2.70} & \underline{61.99}\std{0.39} & 66.71\std{1.31} & \underline{27.49}\std{0.89} & 31.28\std{5.85} & \underline{75.03}\std{2.21} \\
							NodeSketch & \textbf{61.94}\std{1.47} & 67.45\std{1.36} & 66.18\std{3.29} & 74.02\std{1.97} & 59.50\std{2.01} & 62.72\std{3.04} & 26.07\std{5.19} & 24.17\std{3.57} & 72.09\std{1.60} \\
							node2binary & 53.17\std{2.31} & 63.84\std{4.12} & 53.95\std{4.04} & \textbf{76.85}\std{1.55} & 41.66\std{4.25} & 42.84\std{3.12} & 24.85\std{2.94} & 41.18\std{2.57} & 58.98\std{4.25} \\
							\cmidrule(lr){1-10}
							Bi-GCN                  & 58.14\std{1.45} & 67.64\std{1.39} & 63.74\std{1.34} & 76.66\std{0.72} & 58.67\std{2.33} & 60.82\std{1.91} & 36.45\std{2.06} & 35.43\std{8.55} & 75.60\std{4.70} \\
							Raw Features            & 66.67\std{1.86} & 60.72\std{1.88} & 56.04\std{1.86} & 49.66\std{2.44} & 47.95\std{0.44} & 46.74\std{1.50} & 20.47\std{0.25} & 66.13\std{3.14} & 53.22\std{0.00} \\
							Raw Features Bin        & 66.33\std{2.49} & 57.11\std{1.86} & 51.80\std{1.50} & 56.73\std{3.25} & 43.08\std{1.50} & 41.37\std{1.34} & 20.13\std{0.17} & 68.76\std{1.10} & 53.22\std{0.00} \\
							\midrule
							\multicolumn{10}{l}{\textbf{Citeseer} (Homophilous)} \\
							\midrule
							\textbf{NyS-Binary (Ours)} & \textbf{70.00}\std{1.79} & \textbf{69.70}\std{1.76} & \textbf{69.47}\std{0.92} & \textbf{67.97}\std{1.25} & \textbf{70.10}\std{1.91} & \textbf{68.90}\std{1.76} & \textbf{66.17}\std{4.41} & \textbf{69.34}\std{1.99} & \textbf{83.86}\std{0.37} \\
							NyS-Binary (Unsupervised) & 61.59\std{1.66} & 59.63\std{1.16} & 59.63\std{2.72} & \underline{56.69}\std{3.49} & \underline{57.63}\std{1.63} & \underline{57.73}\std{2.29} & \underline{37.21}\std{0.95} & \underline{48.41}\std{4.39} & \underline{77.13}\std{1.90} \\
							NodeSig & \underline{61.83}\std{1.71} & \underline{61.49}\std{1.50} & \underline{61.46}\std{1.78} & 39.97\std{16.26} & 46.22\std{1.07} & 50.42\std{0.57} & 19.29\std{0.70} & 21.49\std{1.72} & 58.07\std{3.18} \\
							NodeSketch & 55.22\std{1.93} & 52.99\std{2.97} & 55.69\std{2.64} & 50.52\std{1.29} & 30.66\std{1.63} & 33.23\std{1.15} & 19.79\std{0.64} & 20.19\std{2.40} & 56.20\std{2.99} \\
							node2binary & 51.95\std{3.39} & 50.02\std{7.82} & 50.21\std{5.28} & 36.63\std{4.96} & 44.71\std{4.37} & 46.75\std{3.74} & 23.39\std{1.40} & 22.86\std{5.18} & 57.83\std{4.70} \\
							\cmidrule(lr){1-10}
							Bi-GCN                  & 74.04\std{1.69} & 72.84\std{2.22} & 72.14\std{1.85} & 68.64\std{2.44} & 72.77\std{1.65} & 71.34\std{2.06} & 61.53\std{4.91} & 71.97\std{1.56} & 83.21\std{2.80} \\
							Raw Features            & 72.30\std{1.72} & 75.04\std{1.35} & 75.04\std{1.37} & 64.73\std{2.63} & 71.87\std{0.92} & 70.60\std{1.00} & 10.41\std{0.17} & 69.77\std{1.06} & 47.20\std{2.23} \\
							Raw Features Bin        & 72.00\std{1.62} & 74.24\std{0.87} & 72.70\std{1.03} & 52.69\std{6.73} & 64.97\std{0.63} & 63.36\std{1.15} & 8.28\std{0.40} & 69.20\std{1.23} & 49.55\std{0.85} \\
							\midrule
							\multicolumn{10}{l}{\textbf{Computers} (Homophilous)} \\
							\midrule
							\textbf{NyS-Binary (Ours)} & \underline{88.14}\std{0.21} & 87.86\std{0.12} & \underline{88.36}\std{0.37} & \textbf{87.71}\std{0.44} & \textbf{86.47}\std{0.74} & \textbf{86.58}\std{0.13} & \textbf{66.66}\std{2.57} & \textbf{86.63}\std{1.73} & \textbf{92.03}\std{2.17} \\
							NyS-Binary (Unsupervised) & \textbf{88.29}\std{0.93} & \underline{88.20}\std{0.76} & \textbf{88.53}\std{1.07} & \underline{86.84}\std{0.55} & \underline{84.86}\std{1.16} & \underline{85.43}\std{1.07} & \underline{53.97}\std{2.73} & \underline{77.06}\std{2.37} & \underline{79.96}\std{6.20} \\
							NodeSig & 88.05\std{0.48} & \textbf{88.67}\std{0.54} & 88.17\std{0.27} & 85.08\std{0.94} & 84.18\std{0.51} & 84.04\std{0.94} & 46.64\std{0.32} & 42.46\std{1.48} & 73.18\std{0.26} \\
							NodeSketch & 86.27\std{0.42} & 86.22\std{1.03} & 83.46\std{3.37} & 85.52\std{0.45} & 62.45\std{1.52} & 61.85\std{1.63} & 36.19\std{0.25} & 34.42\std{3.32} & 70.54\std{0.41} \\
							node2binary & 44.05\std{0.36} & 44.62\std{0.85} & 42.96\std{0.50} & 86.19\std{0.18} & 39.02\std{0.35} & 39.10\std{0.41} & 3.33\std{0.10} & 40.26\std{8.93} & 70.55\std{0.14} \\
							\cmidrule(lr){1-10}
							Bi-GCN                  & 84.45\std{1.48} & 84.61\std{2.55} & 85.55\std{1.49} & 68.74\std{11.43} & 80.72\std{2.63} & 81.85\std{2.31} & 41.96\std{5.69} & 32.41\std{13.90} & 72.36\std{1.87} \\
							Raw Features            & 90.05\std{0.30} & 90.31\std{0.41} & 89.96\std{0.77} & 87.64\std{1.58} & 85.89\std{0.40} & 83.01\std{3.38} & 46.03\std{0.10} & 37.46\std{0.06} & 71.59\std{0.10} \\
							Raw Features Bin        & 90.01\std{0.25} & 88.31\std{0.61} & 87.78\std{0.17} & 85.89\std{0.86} & 77.61\std{0.43} & 74.29\std{0.14} & 39.34\std{0.56} & 37.08\std{0.35} & 70.77\std{0.39} \\
							\midrule
							\multicolumn{10}{l}{\textbf{Cora} (Homophilous)} \\
							\midrule
							\textbf{NyS-Binary (Ours)} & \textbf{85.94}\std{0.72} & \textbf{86.39}\std{1.71} & \textbf{85.69}\std{1.52} & \textbf{83.68}\std{1.07} & \textbf{86.55}\std{0.80} & \textbf{85.90}\std{0.60} & \textbf{72.69}\std{5.02} & \textbf{85.20}\std{0.75} & \textbf{92.16}\std{1.55} \\
							NyS-Binary (Unsupervised) & 82.53\std{1.74} & 81.55\std{1.86} & 81.30\std{2.10} & \underline{76.18}\std{2.29} & 78.88\std{2.46} & 78.84\std{2.57} & \underline{44.98}\std{4.21} & \underline{66.30}\std{3.09} & \underline{72.37}\std{5.90} \\
							NodeSig & 82.45\std{0.87} & 80.65\std{0.47} & 79.38\std{0.40} & 63.22\std{9.60} & 70.56\std{2.05} & 68.10\std{1.03} & 30.18\std{0.26} & 29.31\std{0.79} & 66.75\std{0.62} \\
							NodeSketch & 79.29\std{2.03} & 77.04\std{1.88} & 76.47\std{2.52} & 67.57\std{2.48} & 46.00\std{1.98} & 45.06\std{1.05} & 23.12\std{0.44} & 24.35\std{10.03} & 64.44\std{1.42} \\
							node2binary & \underline{84.66}\std{1.15} & \underline{83.85}\std{0.63} & \underline{83.52}\std{1.11} & 73.76\std{2.95} & \underline{79.95}\std{0.86} & \underline{80.69}\std{0.85} & 26.16\std{2.06} & 35.10\std{4.15} & 66.40\std{2.08} \\
							\cmidrule(lr){1-10}
							Bi-GCN                  & 87.33\std{0.69} & 87.25\std{0.31} & 86.47\std{0.25} & 78.80\std{5.42} & 85.20\std{0.87} & 83.60\std{0.38} & 66.58\std{2.92} & 85.24\std{1.17} & 85.74\std{2.86} \\
							Raw Features            & 87.12\std{0.19} & 87.66\std{0.31} & 86.59\std{1.01} & 74.09\std{2.97} & 75.16\std{1.51} & 75.15\std{1.91} & 20.62\std{0.93} & 84.71\std{0.72} & 65.51\std{0.54} \\
							Raw Features Bin        & 86.27\std{0.40} & 85.98\std{0.86} & 83.93\std{1.00} & 70.36\std{2.28} & 63.96\std{1.17} & 61.26\std{1.94} & 15.99\std{0.57} & 84.67\std{0.14} & 65.87\std{0.15} \\
							\bottomrule
						\end{tabular}%
					}%
				}
			}
		\end{minipage}
	\end{center}
	\vspace{0.8em}
	
	\vspace{0.8em}
	\noindent\begin{center}
		\addtocounter{table}{-1}
		\captionof{table}{Comprehensive per-dataset classification accuracy $\pm$ std (\%) across benchmark datasets (Part 2: DBLP, Photo, PubMed, Squirrel, and WikiCS; continued from Table~\ref{tab:results_acc_std_all}). Best binary result per column in \textbf{bold}; second-best \underline{underlined}.}
		\label{tab:results_acc_std_squirrel}
		{\small\renewcommand{\arraystretch}{1.12}\setlength{\tabcolsep}{2.4pt}
			\makebox[\textwidth][c]{%
				\resizebox{\textwidth}{!}{%
					\begin{tabular}{lrrrrrrrrr}
						\toprule
						Method & GCN & SAGE & H2GCN & LINKX & LogReg & MLP & SADP & SADP-Surr & SADP-2W \\
						\midrule
						\multicolumn{10}{l}{\textbf{Dblp} (Homophilous)} \\
						\midrule
						\textbf{NyS-Binary (Ours)} & \textbf{97.66}\std{0.38} & \textbf{97.55}\std{0.43} & \textbf{97.61}\std{0.38} & \textbf{96.77}\std{0.37} & \textbf{97.00}\std{0.30} & \textbf{96.96}\std{0.29} & \textbf{96.51}\std{0.64} & \textbf{95.21}\std{3.91} & \textbf{96.51}\std{0.64} \\
						NyS-Binary (Unsupervised) & \underline{94.60}\std{0.58} & 94.24\std{0.21} & \underline{94.86}\std{1.28} & \underline{94.36}\std{0.48} & \underline{89.54}\std{0.06} & \underline{92.63}\std{0.64} & \underline{74.32}\std{3.58} & 87.35\std{1.33} & \underline{74.32}\std{3.58} \\
						NodeSig & 82.13\std{1.73} & 82.54\std{0.78} & 83.07\std{0.95} & 86.29\std{8.70} & 71.55\std{0.32} & 74.61\std{2.60} & 57.31\std{0.38} & 56.70\std{0.00} & 57.31\std{0.38} \\
						NodeSketch & 72.16\std{0.41} & 76.18\std{1.26} & 76.65\std{1.75} & 92.47\std{1.32} & 60.78\std{1.58} & 63.04\std{0.52} & 55.59\std{0.54} & 52.29\std{7.75} & 55.59\std{0.54} \\
						node2binary & 93.60\std{1.42} & \underline{95.08}\std{0.88} & 93.81\std{0.77} & 90.78\std{2.42} & 75.11\std{3.06} & 88.36\std{1.93} & 54.50\std{2.59} & \underline{91.20}\std{2.03} & 54.50\std{2.59} \\
						\cmidrule(lr){1-10}
						Bi-GCN                  & 90.41\std{0.91} & 91.64\std{0.72} & 89.00\std{4.91} & 93.45\std{2.26} & 85.63\std{0.42} & 89.16\std{0.86} & 75.93\std{0.77} & 62.17\std{22.12} & 75.93\std{0.77} \\
						Raw Features            & 88.87\std{0.43} & 91.27\std{0.85} & 89.79\std{0.28} & 92.64\std{0.12} & 76.85\std{0.83} & 86.13\std{0.51} & 56.69\std{0.01} & 52.25\std{7.75} & 56.69\std{0.01} \\
						Raw Features Bin        & 62.97\std{0.23} & 65.10\std{1.02} & 62.32\std{2.65} & 89.85\std{3.88} & 56.05\std{0.09} & 53.18\std{0.65} & 56.44\std{0.05} & 47.77\std{7.74} & 56.44\std{0.05} \\
						\midrule
						\multicolumn{10}{l}{\textbf{Photo} (Homophilous)} \\
						\midrule
						\textbf{NyS-Binary (Ours)} & \underline{92.30}\std{0.95} & \underline{92.01}\std{0.91} & \textbf{92.11}\std{0.77} & \textbf{91.59}\std{1.05} & \textbf{91.19}\std{0.77} & \textbf{91.36}\std{0.47} & \textbf{80.70}\std{1.75} & \textbf{90.36}\std{0.74} & \textbf{96.02}\std{0.83} \\
						NyS-Binary (Unsupervised) & 92.13\std{0.72} & 91.49\std{0.31} & \underline{91.65}\std{1.13} & \underline{90.56}\std{0.11} & \underline{89.99}\std{1.10} & \underline{90.31}\std{1.28} & \underline{55.80}\std{10.93} & \underline{86.30}\std{1.43} & \underline{87.59}\std{0.42} \\
						NodeSig & \textbf{92.77}\std{0.79} & \textbf{92.10}\std{0.33} & 91.20\std{0.38} & 89.79\std{0.40} & 89.87\std{0.38} & 87.90\std{0.63} & 48.69\std{0.70} & 81.99\std{8.18} & 83.39\std{1.09} \\
						NodeSketch & 91.39\std{1.32} & 90.19\std{0.49} & 89.40\std{0.51} & 88.41\std{0.81} & 70.69\std{0.29} & 69.53\std{1.35} & 31.27\std{2.22} & 26.49\std{14.32} & 75.80\std{3.63} \\
						node2binary & 69.72\std{7.15} & 71.60\std{8.06} & 67.57\std{6.46} & 89.63\std{0.58} & 34.84\std{0.40} & 35.03\std{0.57} & 4.82\std{0.03} & 64.59\std{2.71} & 53.16\std{0.75} \\
						\cmidrule(lr){1-10}
						Bi-GCN                  & 93.39\std{0.63} & 92.95\std{0.54} & 92.90\std{0.19} & 86.22\std{8.02} & 91.24\std{0.28} & 90.50\std{0.09} & 55.31\std{10.48} & 67.15\std{23.68} & 93.48\std{2.56} \\
						Raw Features            & 93.90\std{0.68} & 95.32\std{0.51} & 88.61\std{9.51} & 93.36\std{0.18} & 92.77\std{0.11} & 91.71\std{0.11} & 30.73\std{0.14} & 31.44\std{11.56} & 76.75\std{1.45} \\
						Raw Features Bin        & 93.97\std{0.53} & 93.97\std{0.88} & 92.78\std{0.72} & 86.29\std{11.35} & 84.53\std{0.61} & 82.76\std{0.90} & 24.84\std{0.26} & 43.44\std{15.30} & 71.23\std{3.92} \\
						\midrule
						\multicolumn{10}{l}{\textbf{Pubmed} (Homophilous)} \\
						\midrule
						\textbf{NyS-Binary (Ours)} & \textbf{83.74}\std{0.30} & \textbf{83.35}\std{0.32} & \textbf{83.50}\std{0.40} & \textbf{79.51}\std{0.94} & \textbf{83.38}\std{0.20} & \textbf{83.29}\std{0.38} & \textbf{81.53}\std{1.79} & \textbf{83.16}\std{0.43} & \textbf{88.81}\std{0.25} \\
						NyS-Binary (Unsupervised) & \underline{82.40}\std{0.77} & \underline{81.64}\std{0.27} & \underline{82.49}\std{0.41} & \underline{77.92}\std{1.24} & \underline{78.39}\std{0.38} & \underline{80.32}\std{0.63} & \underline{65.22}\std{2.58} & \underline{72.14}\std{0.90} & \underline{78.54}\std{2.52} \\
						NodeSig & 74.04\std{0.81} & 76.61\std{0.51} & 74.64\std{1.22} & 71.92\std{6.13} & 62.35\std{0.16} & 64.24\std{0.57} & 39.26\std{0.25} & 39.19\std{0.10} & 55.60\std{0.24} \\
						NodeSketch & 62.83\std{0.96} & 67.98\std{1.12} & 66.05\std{1.96} & 72.98\std{0.48} & 46.62\std{0.67} & 45.98\std{0.73} & 38.20\std{1.08} & 39.83\std{0.04} & 51.75\std{0.32} \\
						node2binary & 57.63\std{2.30} & 61.08\std{1.36} & 59.60\std{2.56} & 77.08\std{0.60} & 44.10\std{0.35} & 45.30\std{0.42} & 20.98\std{0.41} & 31.34\std{0.85} & 50.87\std{0.32} \\
						\cmidrule(lr){1-10}
						Bi-GCN                  & 85.29\std{0.27} & 86.09\std{0.46} & 86.04\std{0.37} & 76.64\std{7.43} & 83.89\std{0.34} & 83.98\std{0.47} & 75.80\std{4.12} & 68.89\std{25.07} & 83.02\std{3.93} \\
						Raw Features            & 87.69\std{0.43} & 89.20\std{0.73} & 89.61\std{0.55} & 85.56\std{0.21} & 81.64\std{0.60} & 87.46\std{0.36} & 20.83\std{0.02} & 87.27\std{0.18} & 50.44\std{0.00} \\
						Raw Features Bin        & 78.60\std{0.14} & 80.05\std{0.39} & 79.66\std{0.06} & 73.76\std{2.28} & 54.96\std{0.92} & 53.33\std{0.46} & 20.82\std{0.04} & 78.07\std{0.66} & 50.44\std{0.00} \\
						\midrule
						\multicolumn{10}{l}{\textbf{Squirrel} (Heterophilous)} \\
						\midrule
						\textbf{NyS-Binary (Ours)} & 40.85\std{1.69} & 46.00\std{3.01} & 50.82\std{1.98} & \textbf{68.23}\std{1.06} & 39.61\std{1.32} & 46.85\std{3.24} & \textbf{24.51}\std{1.65} & \underline{24.81}\std{3.07} & \underline{52.00}\std{2.09} \\
						NyS-Binary (Unsupervised) & \underline{42.32}\std{0.58} & \textbf{49.95}\std{1.21} & \textbf{55.37}\std{0.31} & 66.17\std{0.33} & \textbf{43.58}\std{0.87} & \textbf{53.19}\std{2.78} & \underline{22.29}\std{1.33} & 21.74\std{0.39} & 48.48\std{2.91} \\
						NodeSig & \textbf{43.24}\std{0.45} & 45.97\std{2.87} & 48.90\std{1.41} & 67.14\std{0.36} & \underline{40.15}\std{0.71} & \underline{49.35}\std{1.83} & 22.10\std{0.12} & 21.33\std{1.40} & \textbf{52.05}\std{1.85} \\
						NodeSketch & 40.49\std{1.68} & \underline{47.83}\std{2.35} & \underline{51.95}\std{0.90} & 67.44\std{1.09} & 39.68\std{1.65} & 49.26\std{1.17} & 21.83\std{2.85} & 21.42\std{1.14} & 51.57\std{1.76} \\
						node2binary & 30.09\std{0.96} & 36.56\std{2.53} & 35.92\std{0.68} & \underline{67.80}\std{1.89} & 22.72\std{0.13} & 21.95\std{0.77} & 20.12\std{0.17} & \textbf{32.80}\std{2.15} & 50.08\std{0.32} \\
						\cmidrule(lr){1-10}
						Bi-GCN                  & 42.04\std{1.83} & 44.01\std{2.56} & 45.31\std{2.30} & 67.37\std{0.61} & 36.96\std{1.94} & 38.78\std{2.05} & 25.50\std{2.48} & 24.73\std{1.99} & 54.83\std{1.96} \\
						Raw Features            & 48.99\std{1.70} & 41.79\std{0.85} & 36.52\std{1.89} & 39.25\std{2.00} & 33.97\std{1.39} & 32.61\std{1.57} & 20.07\std{0.04} & 36.22\std{3.85} & 50.08\std{0.00} \\
						Raw Features Bin        & 49.60\std{0.93} & 39.89\std{0.55} & 35.98\std{0.43} & 40.79\std{1.02} & 32.29\std{0.62} & 30.92\std{0.20} & 20.05\std{0.00} & 40.89\std{1.43} & 50.08\std{0.00} \\
						\midrule
						\multicolumn{10}{l}{\textbf{Wikics} (Homophilous)} \\
						\midrule
						\textbf{NyS-Binary (Ours)} & \textbf{80.71}\std{0.70} & \textbf{80.10}\std{0.92} & \textbf{80.09}\std{0.61} & \textbf{76.35}\std{2.20} & \textbf{79.91}\std{0.57} & \textbf{79.52}\std{0.10} & \textbf{56.27}\std{2.85} & \textbf{76.24}\std{0.36} & \textbf{91.82}\std{1.32} \\
						NyS-Binary (Unsupervised) & 79.77\std{0.42} & 79.54\std{1.60} & \underline{79.29}\std{1.05} & \underline{75.63}\std{1.14} & \underline{76.94}\std{0.83} & \underline{77.36}\std{0.87} & \underline{40.55}\std{4.10} & \underline{56.27}\std{14.47} & \underline{86.05}\std{2.48} \\
						NodeSig & \underline{80.03}\std{0.38} & \underline{80.01}\std{0.83} & 78.12\std{0.68} & 75.11\std{1.18} & 75.71\std{0.36} & 74.31\std{0.13} & 29.73\std{0.74} & 42.53\std{10.94} & 65.29\std{0.46} \\
						NodeSketch & 78.28\std{0.38} & 77.97\std{0.32} & 75.75\std{1.23} & 74.22\std{0.81} & 58.11\std{1.27} & 57.77\std{2.26} & 19.78\std{0.45} & 15.99\std{3.93} & 61.61\std{2.37} \\
						node2binary & 47.02\std{1.98} & 49.88\std{2.95} & 46.12\std{1.55} & 74.81\std{0.86} & 27.26\std{0.73} & 27.29\std{0.64} & 2.47\std{0.09} & 52.08\std{1.23} & 55.59\std{0.43} \\
						\cmidrule(lr){1-10}
						Bi-GCN                  & 82.89\std{0.37} & 82.89\std{0.45} & 81.95\std{0.34} & 81.03\std{1.13} & 81.35\std{0.45} & 80.38\std{0.48} & 50.89\std{0.52} & 73.79\std{3.57} & 90.67\std{0.69} \\
						Raw Features            & 82.33\std{0.34} & 84.64\std{0.92} & 84.12\std{0.15} & 81.70\std{1.33} & 76.18\std{0.43} & 80.23\std{0.78} & 29.68\std{0.40} & 12.51\std{8.48} & 57.73\std{0.88} \\
						Raw Features Bin        & 72.02\std{0.81} & 72.34\std{0.78} & 67.52\std{0.19} & 72.94\std{0.71} & 27.53\std{0.74} & 23.83\std{0.68} & 24.02\std{0.14} & 13.77\std{8.51} & 60.30\std{1.88} \\
						
						\bottomrule
					\end{tabular}%
				}%
			}
		}
	\end{center}
	\vspace{0.8em}

	\vspace{0.8em}
	\noindent
	\begin{center}
		\small
		\captionof{table}{Dataset-by-dataset embedding generation time (seconds) averaged across seeds, ordered by graph size $N$. NyS-Binary completes embedding generation in under one second on every benchmark dataset.}
		\label{tab:runtime_per_dataset}
		\resizebox{\textwidth}{!}{%
			\begin{tabular}{lrrrrrrrrrr}
				\toprule
				\textbf{Method} & \textbf{Chameleon} & \textbf{Cora} & \textbf{CiteSeer} & \textbf{Squirrel} & \textbf{Photo} & \textbf{WikiCS} & \textbf{DBLP} & \textbf{Computers} & \textbf{PubMed} & \textbf{Amazon-R.} \\
				\textit{Nodes ($N$)} & 2,277 & 2,708 & 3,327 & 5,201 & 7,650 & 11,701 & 13,326 & 13,752 & 19,717 & 24,492 \\
				\midrule
				NodeSig                & 0.008 & 0.006 & 0.010 & 0.011 & 0.012 & 0.016 & 0.019 & 0.019 & 0.019 & 0.041 \\
				NyS-Binary (Unsupervised) & 0.116 & 0.120 & 0.071 & 0.447 & 0.345 & 0.511 & 0.158 & 0.702 & 0.338 & 0.359 \\
				\textbf{NyS-Binary (Ours)} & \textbf{0.090} & \textbf{0.112} & \textbf{0.033} & \textbf{0.466} & \textbf{0.347} & \textbf{1.245} & \textbf{0.173} & \textbf{0.704} & \textbf{0.347} & \textbf{0.324} \\
				Bi-GCN                 & 2.312 & 1.290 & 2.192 & 8.967 & 5.844 & 9.697 & 2.714 & 11.139 & 4.192 & 5.162 \\
				NodeSketch             & 8.751 & 10.054 & 8.164 & 15.678 & 16.030 & 31.252 & 10.236 & 33.848 & 16.762 & 14.654 \\
				node2binary            & 1632.1 & 972.3 & 1810.1 & 1818.3 & 1825.5 & 1827.7 & 1818.0 & 1876.1 & 1871.9 & 1946.5 \\
				\bottomrule
			\end{tabular}%
		}%
	\end{center}
	\vspace{0.8em}
	\vspace{1.5em}
	
	\subsection{Downstream Benchmark under Low-Supervision Regime across 10 Datasets}
	\label{sec:appendix_low_supervision}
	
	To evaluate performance under sparse label regimes and address questions regarding dependence on label abundance, Table~\ref{tab:low_label_overall} benchmarks downstream classification accuracy across all ten datasets and three random seeds under the 10\% low-supervision regime, spanning six canonical probe architectures (GCN, GraphSAGE, H2GCN, LINKX, Logistic Regression, MLP). Table~\ref{tab:low_label_runtime} reports the corresponding embedding generation latency.
	
	Even under low supervision, \textbf{NyS-Binary (Ours)} exhibits robust discriminative performance:
	\begin{itemize}
		\item \textbf{Dominance over feature-free binary baselines:} NyS-Binary consistently outperforms NodeSig across all message-passing architectures ($+4.28\%$ on GCN, $+5.52\%$ on GraphSAGE, $+7.38\%$ on H2GCN, $+10.10\%$ on LINKX) as well as non-relational probes ($+10.71\%$ on Logistic Regression, $+11.10\%$ on MLP). It exceeds NodeSketch by $+8.11\%$ to $+25.29\%$ and node2binary by $+8.67\%$ to $+26.68\%$.
		\item \textbf{Outperforming learned GNNs on non-relational probes:} On simple linear and feed-forward probes, NyS-Binary surpasses the end-to-end trained, attribute-aware Bi-GCN ($67.95\%$ vs. $67.56\%$ on Logistic Regression; $67.58\%$ vs. $67.12\%$ on MLP), and on heterophily-tailored LINKX ($65.52\%$ vs. $60.23\%$).
		\item \textbf{Sub-second generation latency:} While node2binary requires over $1739$ seconds ($\approx 29$ minutes) and Bi-GCN takes $4.50$ seconds, NyS-Binary constructs its entire 250-bit representations in just $\mathbf{0.3720}$ seconds (median $0.2919$s), delivering a $12.1\times$ speedup over Bi-GCN and $4,677\times$ speedup over node2binary.
	\end{itemize}
	
	\vspace{0.8em}
	\noindent
	\begin{center}
		\captionof{table}{Overall average accuracy (mean $\pm$ standard deviation \% across all 10 datasets and 3 seeds) under 10\% low-supervision regime across six downstream probe classifiers. Feature-free binary methods are grouped above the rule; attribute-aware reference baselines below. Best binary result per column in \textbf{bold}; second-best \underline{underlined}.}
		\label{tab:low_label_overall}
		\resizebox{\textwidth}{!}{%
			\begin{tabular}{lcccccc}
				\toprule
				\textbf{Method} & \textbf{GCN} & \textbf{GraphSAGE} & \textbf{H2GCN} & \textbf{LINKX} & \textbf{Logistic Reg.} & \textbf{MLP} \\
				\midrule
				\textbf{NyS-Binary (Ours)} & \textbf{68.45 $\pm$ 20.31} & \textbf{68.68 $\pm$ 19.27} & \textbf{68.64 $\pm$ 18.82} & \textbf{65.52 $\pm$ 17.22} & \textbf{67.95 $\pm$ 20.10} & \textbf{67.58 $\pm$ 18.87} \\
				NodeSig & \underline{64.17 $\pm$ 18.89} & \underline{63.16 $\pm$ 18.29} & \underline{61.26 $\pm$ 17.75} & 55.42 $\pm$ 15.32 & \underline{57.24 $\pm$ 18.12} & \underline{56.48 $\pm$ 17.31} \\
				NodeSketch & 60.34 $\pm$ 18.54 & 57.14 $\pm$ 17.13 & 54.37 $\pm$ 15.50 & 53.54 $\pm$ 14.67 & 43.37 $\pm$ 11.46 & 42.29 $\pm$ 10.75 \\
				node2binary & 53.14 $\pm$ 16.71 & 53.16 $\pm$ 16.59 & 52.55 $\pm$ 15.97 & \underline{56.85 $\pm$ 16.38} & 41.27 $\pm$ 16.02 & 41.62 $\pm$ 17.30 \\
				\midrule
				Bi-GCN (attr) & 70.29 $\pm$ 19.73 & 70.12 $\pm$ 19.72 & 69.80 $\pm$ 19.20 & 60.23 $\pm$ 18.32 & 67.56 $\pm$ 20.16 & 67.12 $\pm$ 19.66 \\
				Raw Features (attr) & 72.43 $\pm$ 19.54 & 70.45 $\pm$ 22.35 & 69.11 $\pm$ 22.68 & 58.05 $\pm$ 17.86 & 62.38 $\pm$ 20.49 & 63.41 $\pm$ 21.70 \\
				Raw Features Bin (attr) & 66.22 $\pm$ 18.62 & 59.54 $\pm$ 20.23 & 55.98 $\pm$ 20.90 & 50.87 $\pm$ 14.09 & 45.96 $\pm$ 17.79 & 44.32 $\pm$ 17.00 \\
				\bottomrule
			\end{tabular}%
		}
	\end{center}
	\vspace{0.8em}
	
	\vspace{0.8em}
	\noindent
	\begin{center}
		\begin{minipage}{\linewidth}
			\centering
			\captionof{table}{Overall embedding generation latency (mean $\pm$ standard deviation and median seconds across all runs) under the 10\% low-supervision benchmark.}
			\label{tab:low_label_runtime}
			\small
			\setlength{\tabcolsep}{14pt}
			\begin{tabular}{lcc}
				\toprule
				\textbf{Method} & \textbf{Generation Time (Mean $\pm$ Std)} & \textbf{Median (s)} \\
				\midrule
				\textbf{NyS-Binary (Ours)} & \textbf{0.3720s $\pm$ 0.3076s} & \textbf{0.2919s} \\
				Bi-GCN & 4.4997s $\pm$ 2.5960s & 4.1312s \\
				NodeSig & 0.0162s $\pm$ 0.0118s & 0.0135s \\
				NodeSketch & 17.0010s $\pm$ 10.7617s & 13.4900s \\
				Raw Features & 0.0036s $\pm$ 0.0025s & 0.0030s \\
				Raw Features Bin & 0.1724s $\pm$ 0.1064s & 0.1464s \\
				node2binary & 1739.8500s $\pm$ 274.8455s & 1822.4500s \\
				\bottomrule
			\end{tabular}
		\end{minipage}
	\end{center}
	\vspace{0.8em}
	
	\section{Analysis and Discussion}
	\label{sec:analysis}
	\subsection{Embedding geometry}
\begin{figure}[htbp]
		\centering
		\includegraphics[width=0.85\textwidth]{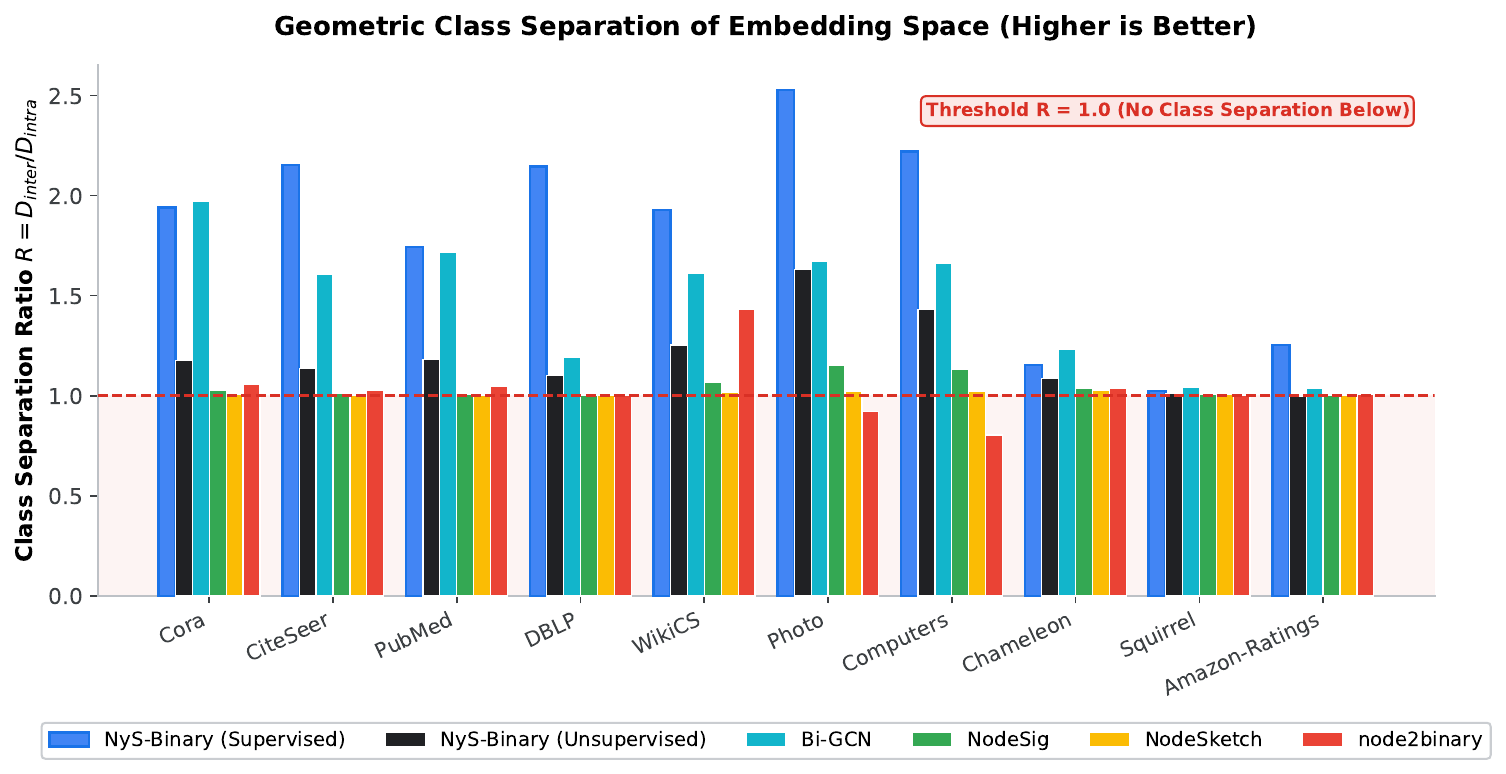}
		\caption{Class-separation ratio $R=D_{\mathrm{inter}}/D_{\mathrm{intra}}$ per
			dataset. Higher is better; the dashed line marks $R=1$.}
		\label{fig:sep}
	\end{figure}

	We quantify separability with the class-separation ratio
	$R=D_{\mathrm{inter}}/D_{\mathrm{intra}}$, the ratio of mean squared
	between-class to within-class distances, where $R=1$ indicates no separation.
	Figure~\ref{fig:sep} shows NyS-Binary produces the best-separated codes on
	homophilous graphs, with $R$ well above $2$ on several, and that $R$ falls toward
	$1$ on the strongly heterophilous graphs, tracking accuracy trends and
	explaining where the method is and is not strong.
	
	\subsection{Component Contributions and Dataset-Dependent Behavior}
	Two design choices drive effectiveness: the Nystr\"om-inspired landmark projection projects the graph transition columns through the landmark block, while the label-leakage-safe semantic channel injects task-specific class signals. Comparing NyS-Binary (Ours) against NyS-Binary (Unsupervised) in Table~\ref{tab:overall} isolates the exact quantitative contribution: adding sparse label diffusion yields a consistent $+1.76\%$ to $+1.90\%$ gain across message-passing GNNs (GCN, GraphSAGE) and $+2.21\%$ to $+3.46\%$ on non-relational probes (MLP, LogReg), while extending gradient-free neuromorphic learning (SADP) by $+16.82\%$.
	
	Importantly, our ablation studies (Appendix~\ref{sec:appendix_ablation}) show that the unsupervised backbone alone ($\text{struct\_ratio}=1.0$) provides a highly potent representation surpassing all prior binary hashing baselines. Different information sources contribute differently depending on network topology: on strongly homophilous graphs (DBLP, PubMed), semantic diffusion provides significant refinement, whereas on heterophilic graphs (Appendix~\ref{sec:appendix_deep_diffusion_heterophily}), the unsupervised structural backbone excels by avoiding discordant label contamination. Crucially, binary quantization preserves geometric separability without accuracy collapse.
	
	\subsection{Limitations}
	\label{sec:limitations}
	We state the boundaries of the method plainly. On graphs with strong heterophily, while NyS-Binary remains competitive with prior binary hashing baselines, performance margins are more moderate than on homophilous networks, and continuous attribute-aware models can have an advantage when node attributes carry the primary discriminative signal. Under non-homophilous edges, our unsupervised variant ($\text{struct\_ratio}=1.0$) is often preferable as it relies purely on topological landmark geometry without diffusing neighbor labels (further analyzed in Appendix~\ref{sec:appendix_deep_diffusion_heterophily}). Investigating adaptive or signed polynomial filters for heterophilous topologies, evaluating degree-stratified landmark sampling, and scaling to web-scale networks (10M+ nodes) remain fruitful directions for future work.
	
	\section{Conclusion}
	NyS-Binary is an algebraic pipeline that constructs compact, discriminative binary node codes by sketching a graph's transition operator with a randomized Nystr\"om-inspired approximation and fusing in a label-leakage-safe label channel. Across ten benchmarks and nine classifiers it delivers the best average accuracy among binary methods at sub-second construction cost and $32\times$ memory reduction, transferring seamlessly even to neuromorphic spiking probes. Furthermore, our fully unsupervised variant demonstrates that Nystr\"om-inspired landmark sketching and polynomial diffusion alone establish a new state of the art for label-free, feature-free graph hashing. Principled algebraic sketching of graph operators provides a strong, practical default when speed and memory are the binding constraints.
	
	\subsubsection*{LLM Usage Disclosure}
	Large language models (LLMs) were consulted during algorithmic refinement, used to structure portions of the benchmarking code, and used to polish manuscript phrasing. The scientific formulations, algorithmic decisions, experiments, and conclusions were developed, verified, and validated by the authors, who take full responsibility for the contents of this paper.
	
	\subsubsection*{Data and Code Reproducibility}
	To facilitate future research and ensure complete transparency, the official source code and benchmark datasets are publicly accessible at \url{https://github.com/Srajansstar/Fast_Binary_Graph_Representation}. The repository provides a comprehensive, ready-to-run environment for generating the structural and semantic binary node embeddings and evaluating them across our 10-dataset benchmark suite. 
	
	Key repository contents include:
	\begin{itemize}[leftmargin=*,itemsep=2pt]
		\item \textbf{Modular Pipeline:} The core algebraic embedding logic, highly-optimized sparse matrix operators, and external baseline implementations are provided as modular Python scripts for seamless integration into external workflows.
		\item \textbf{Interactive Notebooks:} We provide heavily documented Jupyter Notebooks (e.g., \texttt{Main.ipynb} and the ablation suite) that offer step-by-step walkthroughs to independently reproduce the main classification results, hyperparameter grid-searches, and Nystr\"om scaling studies.
		\item \textbf{Pre-processed Datasets:} All homophilous and strongly heterophilic graphs evaluated in this work are bundled with automated loaders for instant out-of-the-box execution.
	\end{itemize}
	
	\bibliographystyle{plainnat}
	\bibliography{references}
	
	\clearpage
	\appendix
	\section*{Appendix}
	
	\section{External Baselines, Adaptations, and Embedding Generators}
	\label{sec:appendix_baselines}
	
	To place NyS-Binary in context, we benchmarked against leading binary graph hashing algorithms, continuous spectral references, and structural backbone variants. Below, we summarize each baseline and explicitly document where our evaluation protocol or implementation adapted the original author-published procedures:
	
	\subsubsection{External Binary Graph Hashing Baselines}
	\begin{enumerate}[leftmargin=*,itemsep=2.5pt]
		\item \textbf{Bi-GCN \citep{wang2021bigcn} (CVPR 2021 / TPAMI 2024):}
		A learned binary graph convolutional network that constrains both network weights and node representations to $\{-1, +1\}$ using $\operatorname{sign}(x)$, trained via the straight-through estimator (STE) on the raw continuous node features $\mathbf{X}$.
		\begin{itemize}[leftmargin=12pt,itemsep=1.5pt]
			\item \textit{Adaptation from Original Paper:} In the original work, Bi-GCN was designed strictly as an end-to-end task-specific classifier that outputs class prediction logits directly. For our benchmark, we trained the two-layer Bi-GCN model on the training mask using the graph's actual continuous node attributes, and \textbf{extracted the intermediate binarized hidden activations} ($\mathbf{H}_2 \in \{-1, +1\}^{N \times K}$, mapped to $\{0, 1\}^{N \times K}$) as a static binary node embedding. This representation was then evaluated across all nine downstream classifiers for a consistent, like-for-like comparison against the feature-free binary hashing baselines.
		\end{itemize}
		\item \textbf{node2binary \citep{talukder2025node2binary} (WWW 2025):}
		An unsupervised binary embedding framework that combines Leiden community detection with random-walk co-occurrence modeling to build a hierarchical tree of binary codes where shared prefixes reflect community membership.
		\begin{itemize}[leftmargin=12pt,itemsep=1.5pt]
			\item \textit{Evaluation Setup:} We executed the author's official, unmodified implementation with default hyper-parameters (Leiden trees=1, depth=3) to generate bitcodes purely from graph topology, exactly as described in the original paper. Because the author's script bundles internal downstream classification with embedding generation, we extracted the generated unsupervised binary vectors directly and evaluated them across our standardized 70/30 stratified train/test splits across all nine downstream classifiers, ensuring a fair and uniform benchmark.
		\end{itemize}
		\item \textbf{NodeSig \citep{celikkanat2022nodesig} (IEEE/ACM ASONAM 2022):}
		A shallow, non-parametric binary hashing algorithm that computes multi-hop random-walk transition probability vectors and projects them onto Hamming space via random hyperplane sign hashing.
		\begin{itemize}[leftmargin=12pt,itemsep=1.5pt]
			\item \textit{Adaptation from Original Paper:} We executed the author's official compiled C++ executable. Because the C++ implementation packs bits into 8-bit bytes, it strictly requires the embedding dimension to be a multiple of 8. To maintain an identical $K=250$ bit budget across all methods, we generated $\lceil K/8 \rceil \times 8 = 256$ bits and truncated the resulting representation to exactly $K=250$ bits.
		\end{itemize}
		\item \textbf{NodeSketch \citep{yang2019nodesketch} (ACM SIGKDD 2019):}
		A discrete locality-sensitive hashing (ALSH) method that sketches multi-hop decay-weighted neighborhood distributions via recursive Min-Hash.
		\begin{itemize}[leftmargin=12pt,itemsep=1.5pt]
			\item \textit{Adaptation from Original Paper:} The author's official MATLAB/C MEX implementation outputs integer-valued min-hash sketch codes ($\mathbf{H} \in \mathbb{N}^{N \times K}$). To convert these into compact binary bitcodes suitable for Hamming-distance computation and binary downstream classifiers without violating the locality-sensitive hashing property, we adopt the established \textbf{b-bit minwise hashing} framework \citep{li2010bbit}. Specifically, we extract the least significant bit (LSB) of each min-hash signature: $\mathbf{B} = (\mathbf{H} \bmod 2 == 1) \in \{0, 1\}^{N \times K}$. As proven by \citet{li2010bbit}, retaining only the lowest-order bit preserves the underlying Jaccard similarity in Hamming space with provable collision probabilities ($P(b_i = b_j) = \frac{1 + J}{2}$), providing a mathematically grounded binary representation for recursive min-hash signatures.
		\end{itemize}
	\end{enumerate}
	
	\subsubsection{Continuous and Attribute-Aware Reference Baselines}
	\begin{enumerate}[leftmargin=*,itemsep=2pt]
		\item \textbf{Raw Features \& Raw Features Binarized:} Attribute-aware references that directly feed raw continuous node feature vectors (or median-binarized bitcodes) into downstream classifiers, establishing the performance ceiling when node attributes are available.
	\end{enumerate}
	
	\section{Downstream Classifier Architectures and Hyperparameter Configurations}
	\label{sec:appendix_classifier_architectures}
	
	To rigorously evaluate representation quality across diverse operational modalities, we evaluate nine downstream probe classifiers across all benchmarks. These encompass standard message-passing graph neural networks, heterophily-tailored architectures, non-relational linear and feedforward probes, and biologically plausible neuromorphic spiking networks. Below we specify their architectural mechanics and configuration hyperparameters:
	
	\subsubsection{Message-Passing Graph Neural Networks}
	\begin{itemize}[leftmargin=*,itemsep=3pt]
		\item \textbf{Graph Convolutional Network (GCN) \citep{kipf2017gcn}:}
		The model applies two successive graph convolutional layers with symmetrically normalized adjacency aggregation:
		\begin{equation}
			\mathbf{Z}^{(1)} = \text{ReLU}\big(\tilde{\mathbf{D}}^{-1/2}\tilde{\mathbf{A}}\tilde{\mathbf{D}}^{-1/2} \, \mathbf{X} \, \mathbf{W}^{(0)}\big), \quad
			\hat{\mathbf{Y}} = \tilde{\mathbf{D}}^{-1/2}\tilde{\mathbf{A}}\tilde{\mathbf{D}}^{-1/2} \, \mathbf{Z}^{(1)} \, \mathbf{W}^{(1)},
		\end{equation}
		where $\tilde{\mathbf{A}} = \mathbf{A} + \mathbf{I}_N$. Layer dimensions are $\text{dim}_{\mathrm{in}} \to 64 \to C$. Optimization uses Adam with learning rate $\eta = 0.01$, weight decay $\lambda = 5 \times 10^{-4}$, and cross-entropy loss trained for 200 epochs.
		
		\item \textbf{GraphSAGE \citep{hamilton2017graphsage}:}
		The architecture uses two inductive mean-aggregation convolutional layers:
		\begin{equation}
			\begin{aligned}
				\mathbf{h}_v^{(1)} &= \text{ReLU}\Big(\mathbf{W}_1 \cdot \text{CONCAT}\Big(\mathbf{x}_v, \, \frac{1}{|\mathcal{N}(v)|}\sum_{u \in \mathcal{N}(v)} \mathbf{x}_u\Big)\Big), \\
				\hat{\mathbf{y}}_v &= \mathbf{W}_2 \cdot \text{CONCAT}\Big(\mathbf{h}_v^{(1)}, \, \frac{1}{|\mathcal{N}(v)|}\sum_{u \in \mathcal{N}(v)} \mathbf{h}_u^{(1)}\Big).
			\end{aligned}
		\end{equation}
		Hidden dimension is 64 with ReLU activation. Optimized via Adam ($\eta=0.01, \lambda=5 \times 10^{-4}$) for 200 epochs.
	\end{itemize}
	
	\subsubsection{Heterophily-Tailored Architectures}
	\begin{itemize}[leftmargin=*,itemsep=3pt]
		\item \textbf{LINKX \citep{lim2021linkx}:}
		Designed for graphs with strong heterophily, LINKX decouples feature representations and topological structure to bypass oversmoothing. The model comprises:
		\begin{enumerate}[label=(\roman*),leftmargin=15pt,itemsep=1pt]
			\item A 2-layer node MLP ($\text{dim}_{\mathrm{in}} \to 64 \to 64$) processing the node feature representations $\mathbf{X}$;
			\item A 2-layer edge MLP ($N \to 64 \to 64$) processing the sparse graph adjacency matrix $\mathbf{A}$;
			\item A 2-layer fusion MLP ($128 \to 64 \to C$) combining the node and edge channels with dropout $p=0.5$.
		\end{enumerate}
		All internal channels are set to 64. Trained using Adam ($\eta=0.01, \lambda=5 \times 10^{-4}$) for 200 epochs.
		
		\item \textbf{H2GCN \citep{zhu2020h2gcn}:}
		This architecture captures higher-order neighborhood topologies under heterophily:
		\begin{enumerate}[label=(\roman*),leftmargin=15pt,itemsep=1pt]
			\item An initial feature embedding layer projects input representations to hidden dimension 64: $\mathbf{X}^{(0)} = \text{ReLU}(\mathbf{X} \mathbf{W}_0)$;
			\item Two sequential propagation steps aggregate normalized neighbors without self-loops: $\mathbf{X}^{(1)} = \mathbf{D}^{-1/2}\mathbf{A}\mathbf{D}^{-1/2}\mathbf{X}^{(0)}$ and $\mathbf{X}^{(2)} = \mathbf{D}^{-1/2}\mathbf{A}\mathbf{D}^{-1/2}\mathbf{X}^{(1)}$;
			\item Ego and neighborhood signals are concatenated across scales: $\mathbf{X}_{\mathrm{concat}} = [\mathbf{X}^{(0)} \,\|\, \mathbf{X}^{(1)} \,\|\, \mathbf{X}^{(2)}] \in \mathbb{R}^{N \times 192}$;
			\item A final linear projection maps the 192-dimensional multi-scale representation to target classes: $\hat{\mathbf{Y}} = \mathbf{X}_{\mathrm{concat}} \mathbf{W}_{\mathrm{out}}$.
		\end{enumerate}
		Trained with Adam ($\eta=0.01, \lambda=5 \times 10^{-4}$) for 200 epochs.
	\end{itemize}
	
	\subsubsection{Non-Relational and Linear Probes}
	\begin{itemize}[leftmargin=*,itemsep=3pt]
		\item \textbf{Multilayer Perceptron (MLP):}
		This probe comprises two dense linear layers: $\mathbf{Z} = \text{ReLU}(\mathbf{X} \mathbf{W}_1 + \mathbf{b}_1)$ followed by $\hat{\mathbf{Y}} = \mathbf{Z} \mathbf{W}_2 + \mathbf{b}_2$ (dimensions $\text{dim}_{\mathrm{in}} \to 64 \to C$). Critically, the MLP operates purely on node embeddings without access to graph edges, isolating the intrinsic linear and non-linear separability of the representations. Trained with Adam ($\eta=0.01, \lambda=5 \times 10^{-4}$) for 200 epochs.
		
		\item \textbf{Logistic Regression:}
		This constitutes a strictly convex, parameter-free linear probe ($\hat{\mathbf{Y}} = \mathbf{X} \mathbf{W} + \mathbf{b}$). By eschewing hidden representations and non-linearities, it directly measures linear decodability and geometric class separation in Hamming space. Trained with Adam ($\eta=0.01, \lambda=5 \times 10^{-4}$) for 200 epochs.
	\end{itemize}
	
	\section{Evaluation on Highly Heterophilic Graphs with Deep Diffusion ($H=10$)}
	\label{sec:appendix_deep_diffusion_heterophily}
	
	To evaluate the generalization boundaries of NyS-Binary across non-standard topologies, we extended our evaluation to four benchmark graphs exhibiting strong heterophily: \textbf{Actor}, \textbf{Texas}, \textbf{Wisconsin}, and \textbf{Minesweeper}. In strongly heterophilic graphs, adjacent nodes frequently belong to differing classes, which causes localized 1- or 2-hop neighborhoods to aggregate discordant signals. To mitigate this and capture higher-order relational context across heterophilic topologies, we expanded the diffusion scheme for these experiments to an extended 10-hop diffusion horizon ($H=10$). Their topological characteristics, node and edge counts, and homophily ratios are summarized in Table~\ref{tab:deep_diffusion_datasets}.
	
	\vspace{0.8em}
	\noindent
	\begin{center}
		\small
		\captionof{table}{Summary statistics of the four highly heterophilic benchmark datasets evaluated under extended 10-hop deep diffusion ($H=10$), detailing the number of nodes ($N$), undirected edges ($E$), target classes ($C$), edge homophily ratio ($h$), and topological domain.}
		\label{tab:deep_diffusion_datasets}
		\resizebox{0.88\textwidth}{!}{%
			\begin{tabular}{llrrcrl}
				\toprule
				\textbf{Dataset} & \textbf{Domain} & \textbf{Nodes ($N$)} & \textbf{Edges ($E$)} & \textbf{Classes ($C$)} & \textbf{Homophily ($h$)} & \textbf{Topology} \\
				\midrule
				Actor           & Actor co-occurrence  & 7,600  & 15,009  & 5  & 0.22 & Heterophilous \\
				Texas           & WebKB hyperlink      & 183    & 162     & 5  & 0.11 & Heterophilous \\
				Wisconsin       & WebKB hyperlink      & 251    & 257     & 5  & 0.21 & Heterophilous \\
				Minesweeper     & Synthetic grid       & 10,000 & 39,402  & 2  & 0.68 & Heterophilous \\
				\bottomrule
			\end{tabular}%
		}
	\end{center}
	\vspace{0.8em}

	Table~\ref{tab:deep_diffusion_h8} presents the overall macro-average accuracy $\pm$ standard deviation across all four high-heterophily benchmarks, and Table~\ref{tab:deep_diffusion_per_dataset} details the complete per-dataset breakdown across 3 random seeds (42, 123, 77), evaluated over four canonical downstream probe architectures (GCN, GraphSAGE, LINKX, and MLP).
	
	\vspace{0.8em}
	\noindent
	\begin{center}
		\begin{minipage}{\linewidth}
			\centering
			\captionof{table}{Overall classification accuracy (mean $\pm$ std \%) across four highly heterophilic benchmark datasets (\textbf{Actor}, \textbf{Texas}, \textbf{Wisconsin}, and \textbf{Minesweeper}) under 10-hop deep diffusion ($H=10$) across four downstream probe classifiers. Feature-free binary hashing methods are grouped above the rule; attribute-aware baselines are shown below. Best feature-free binary result per column in bold.}
			\label{tab:deep_diffusion_h8}
			\resizebox{0.88\textwidth}{!}{%
				\begin{tabular}{lrrrr}
					\toprule
					\textbf{Method} & \textbf{GCN} & \textbf{GraphSAGE} & \textbf{LINKX} & \textbf{MLP} \\
					\midrule
					\textbf{NyS-Binary (Ours)}                    & 48.76 $\pm$ 20.91 & 51.99 $\pm$ 20.82 & 47.16 $\pm$ 19.97 & 47.85 $\pm$ 21.50 \\
					\textbf{NyS-Binary (Unsupervised)}            & \textbf{49.92 $\pm$ 21.14} & 53.65 $\pm$ 20.99 & 49.08 $\pm$ 18.96 & 48.98 $\pm$ 20.61 \\
					NodeSig                                       & 47.67 $\pm$ 21.68 & 54.11 $\pm$ 21.67 & 46.18 $\pm$ 17.42 & 48.66 $\pm$ 20.28 \\
					NodeSketch                                    & 48.76 $\pm$ 21.07 & \textbf{57.90 $\pm$ 22.14} & \textbf{50.27 $\pm$ 18.93} & \textbf{52.38 $\pm$ 21.63} \\
					\midrule
					Bi-GCN (attr)                                 & 49.75 $\pm$ 20.04 & 57.16 $\pm$ 20.24 & 50.93 $\pm$ 17.60 & 51.78 $\pm$ 20.54 \\
					Raw Features (attr)                           & 54.01 $\pm$ 19.18 & 73.82 $\pm$ 24.22 & 60.32 $\pm$ 18.78 & 72.65 $\pm$ 23.04 \\
					Raw Features Bin (attr)                       & 52.65 $\pm$ 19.49 & 68.79 $\pm$ 23.49 & 55.30 $\pm$ 18.14 & 68.07 $\pm$ 21.82 \\
					\bottomrule
				\end{tabular}%
			}
		\end{minipage}
	\end{center}
	\vspace{0.8em}
	
	\vspace{0.8em}
	\noindent
	\begin{center}
		\captionof{table}{Per-dataset classification accuracy $\pm$ std (\%) across all four heterophilic datasets (\textbf{Actor}, \textbf{Texas}, \textbf{Wisconsin}, \textbf{Minesweeper}) under 10-hop deep diffusion ($H=10$). Feature-free binary methods are listed above the inner rule; attribute-aware baselines are listed below. Best feature-free binary result per column in bold.}
		\label{tab:deep_diffusion_per_dataset}
		\resizebox{0.92\textwidth}{!}{%
			\begin{tabular}{llrrrr}
				\toprule
				\textbf{Dataset} & \textbf{Method} & \textbf{GCN} & \textbf{GraphSAGE} & \textbf{LINKX} & \textbf{MLP} \\
				\midrule
				\textbf{Actor}
				& NyS-Binary (Ours)                    & \textbf{24.66 $\pm$ 0.35} & 24.05 $\pm$ 1.20 & 20.61 $\pm$ 0.31 & 22.47 $\pm$ 0.73 \\
				& NyS-Binary (Unsupervised)            & 24.20 $\pm$ 0.37 & \textbf{24.18 $\pm$ 0.82} & \textbf{21.19 $\pm$ 0.38} & \textbf{23.70 $\pm$ 1.52} \\
				& NodeSig                              & 22.84 $\pm$ 0.37 & 22.75 $\pm$ 0.62 & 20.53 $\pm$ 0.56 & 23.41 $\pm$ 0.50 \\
				& NodeSketch                           & 24.33 $\pm$ 0.50 & 23.55 $\pm$ 1.72 & 21.11 $\pm$ 1.90 & 23.18 $\pm$ 0.46 \\
				\cmidrule(lr){2-6}
				& Bi-GCN (attr)                        & 28.00 $\pm$ 1.35 & 28.90 $\pm$ 1.12 & 24.37 $\pm$ 0.84 & 27.93 $\pm$ 2.02 \\
				& Raw Features (attr)                  & 29.46 $\pm$ 1.00 & 33.89 $\pm$ 0.35 & 29.69 $\pm$ 0.27 & 34.91 $\pm$ 1.51 \\
				& Raw Features Bin (attr)              & 27.91 $\pm$ 0.60 & 30.22 $\pm$ 0.86 & 27.05 $\pm$ 1.16 & 32.49 $\pm$ 1.08 \\
				\midrule
				\textbf{Texas}
				& NyS-Binary (Ours)                    & 46.66 $\pm$ 2.10 & 54.55 $\pm$ 4.81 & \textbf{58.79 $\pm$ 2.78} & 44.85 $\pm$ 2.10 \\
				& NyS-Binary (Unsupervised)            & \textbf{50.30 $\pm$ 2.10} & 61.82 $\pm$ 0.00 & 56.36 $\pm$ 3.15 & 49.09 $\pm$ 1.82 \\
				& NodeSig                              & 44.85 $\pm$ 1.05 & 60.61 $\pm$ 3.79 & 53.94 $\pm$ 6.39 & 47.27 $\pm$ 1.82 \\
				& NodeSketch                           & 49.09 $\pm$ 1.82 & \textbf{69.70 $\pm$ 2.10} & \textbf{58.79 $\pm$ 5.55} & \textbf{60.61 $\pm$ 2.10} \\
				\cmidrule(lr){2-6}
				& Bi-GCN (attr)                        & 46.66 $\pm$ 1.05 & 63.03 $\pm$ 3.78 & 58.79 $\pm$ 2.78 & 52.12 $\pm$ 5.25 \\
				& Raw Features (attr)                  & 57.58 $\pm$ 7.57 & 86.67 $\pm$ 3.78 & 67.27 $\pm$ 3.15 & 86.66 $\pm$ 1.05 \\
				& Raw Features Bin (attr)              & 54.55 $\pm$ 3.64 & 78.18 $\pm$ 3.64 & 58.79 $\pm$ 4.20 & 75.15 $\pm$ 3.78 \\
				\midrule
				\textbf{Wisconsin}
				& NyS-Binary (Ours)                    & 43.86 $\pm$ 6.08 & 49.56 $\pm$ 2.74 & 39.03 $\pm$ 2.01 & 44.30 $\pm$ 3.31 \\
				& NyS-Binary (Unsupervised)            & \textbf{45.18 $\pm$ 8.35} & 50.00 $\pm$ 7.89 & 48.24 $\pm$ 3.80 & 44.73 $\pm$ 6.84 \\
				& NodeSig                              & 42.98 $\pm$ 6.75 & 53.07 $\pm$ 4.62 & 45.61 $\pm$ 5.32 & \textbf{46.05 $\pm$ 1.32} \\
				& NodeSketch                           & 41.67 $\pm$ 2.74 & \textbf{58.77 $\pm$ 2.01} & \textbf{52.19 $\pm$ 3.31} & \textbf{46.05 $\pm$ 3.48} \\
				\cmidrule(lr){2-6}
				& Bi-GCN (attr)                        & 44.30 $\pm$ 7.71 & 54.39 $\pm$ 5.32 & 53.07 $\pm$ 2.74 & 45.17 $\pm$ 3.04 \\
				& Raw Features (attr)                  & 49.12 $\pm$ 0.76 & 88.60 $\pm$ 4.02 & 73.68 $\pm$ 2.64 & 89.03 $\pm$ 0.76 \\
				& Raw Features Bin (attr)              & 48.25 $\pm$ 3.04 & 83.77 $\pm$ 4.23 & 65.35 $\pm$ 4.02 & 84.65 $\pm$ 1.52 \\
				\midrule
				\textbf{Minesweeper}
				& NyS-Binary (Ours)                    & 79.84 $\pm$ 0.27 & 79.82 $\pm$ 0.34 & 70.19 $\pm$ 5.22 & \textbf{79.78 $\pm$ 0.39} \\
				& NyS-Binary (Unsupervised)            & \textbf{80.01 $\pm$ 0.02} & 78.58 $\pm$ 2.35 & \textbf{70.52 $\pm$ 4.04} & 78.40 $\pm$ 0.77 \\
				& NodeSig                              & 80.00 $\pm$ 0.00 & \textbf{80.01 $\pm$ 0.02} & 64.65 $\pm$ 2.99 & 77.89 $\pm$ 2.55 \\
				& NodeSketch                           & 79.97 $\pm$ 0.06 & 79.59 $\pm$ 0.19 & 69.00 $\pm$ 3.04 & 79.67 $\pm$ 0.26 \\
				\cmidrule(lr){2-6}
				& Bi-GCN (attr)                        & 80.05 $\pm$ 0.16 & 82.31 $\pm$ 0.38 & 67.48 $\pm$ 10.90 & 81.89 $\pm$ 0.56 \\
				& Raw Features (attr)                  & 79.89 $\pm$ 0.14 & 86.15 $\pm$ 0.28 & 70.62 $\pm$ 4.03 & 80.00 $\pm$ 0.00 \\
				& Raw Features Bin (attr)              & 79.89 $\pm$ 0.28 & 82.98 $\pm$ 0.89 & 70.00 $\pm$ 9.16 & 80.00 $\pm$ 0.00 \\
				\bottomrule
			\end{tabular}%
		}
	\end{center}
	\vspace{0.8em}
	
	\paragraph{Discussion and Analysis on Heterophilic Graphs:}
	As explicitly acknowledged in the limitations of our approach (Section~\ref{sec:limitations}), NyS-Binary is primarily formulated for structural role discovery and topological compression in homophilic and moderate-heterophily networks. In strongly heterophilic graphs where adjacent nodes systematically belong to divergent classes, standard graph neighborhood aggregation risks mixing conflicting class signals. Consequently, our framework does not dominate heterophilic benchmarks as decisively as it does homophilic networks, trailing attribute-dependent baselines (such as Raw Features (attr) at $73.82\%$ on GraphSAGE) where continuous node features carry the primary discriminative signal.
	
	Nevertheless, under extended 10-hop deep diffusion ($H=10$), our algebraic formulation remains highly competitive with existing binary representations in many cases. By expanding the diffusion horizon to $H=10$, the polynomial operator reaches past noisy 1-hop discordance to capture higher-order structural motifs and topological symmetries. Across the four heterophilic benchmarks (Table~\ref{tab:deep_diffusion_h8}), \textbf{NyS-Binary (Unsupervised)} achieves the highest GCN accuracy among all feature-free binary methods at $\mathbf{49.92\%}$ (outperforming Bi-GCN (attr) at $49.75\%$, NodeSketch at $48.76\%$, and NodeSig at $47.67\%$). Furthermore, it delivers strong GraphSAGE ($53.65\%$) and LINKX ($49.08\%$) performance with sub-second generation latency ($0.141$s median vs. $1.808$s for Bi-GCN (attr), representing a $12.8\times$ speedup).
	
	The per-dataset breakdown in Table~\ref{tab:deep_diffusion_per_dataset} illuminates a crucial mechanistic dynamic: unsupervised hashing consistently outperforms the supervised blend across heterophilic topologies. On \textit{Texas}, NyS-Binary (Unsupervised) achieves $\mathbf{61.82\%}$ on GraphSAGE ($+7.27\%$ over the blend) and $\mathbf{50.30\%}$ on GCN; on \textit{Wisconsin}, it reaches $\mathbf{48.24\%}$ on LINKX ($+9.21\%$ over the blend). In heterophilic graphs, propagating training labels along immediate edges introduces semantic label contamination because connected nodes possess conflicting classes. Operating purely without label diffusion eliminates this noise, allowing orthogonal spectral coordinates to capture genuine structural role affinities while maintaining robust classification performance. The reliance on raw transition operator powers ($\mathbf{P}^H$) without sign-reversal or signed filters remains a fundamental structural constraint for non-homophilous graphs. As a designated algorithmic upgrade for non-homophilous topologies, future work can incorporate signed polynomial filters (e.g., alternating signed expansions $\sum_{h=0}^H (-1)^h c_h \mathbf{P}^h$ or dual-channel low-pass/high-pass spectral decomposition) to capture cross-class boundary discrepancies directly into discrete bitcodes.
	
	\section{Memory and Storage Footprint Analysis}
	\label{sec:appendix_memory_storage}
	
	Table~\ref{tab:storage_footprint} quantifies the physical memory footprint when node representations are stored under three precision regimes: single-precision floating-point (Float32, 4 bytes per dimension), 8-bit unsigned integer (Uint8, 1 byte per dimension), and fully bit-packed binary arrays (1 bit per dimension). We report these metrics for every binary hashing method; the continuous baseline (Raw Features) is listed for reference under Float32 only, since it cannot be losslessly reduced to a 1-bit representation.
	
	Binary hashing methods (NyS-Binary, NodeSig, NodeSketch, Bi-GCN, node2binary) natively produce $\{0,1\}$-valued embeddings and therefore admit direct bit-packing. A 250-dimensional binary code stored as a packed 1-bit array occupies only $\lceil 250/8 \rceil = 32$ bytes per node, yielding a $32\times$ compression over the same code stored in Float32. When compared against high-dimensional raw input features (e.g., the 3,703-dimensional TF-IDF vectors in CiteSeer at 46.99\,MB), the packed binary embedding compresses storage by up to $462.9\times$, saving 99.8\% of space. Note that for datasets whose raw feature dimensionality is lower than the chosen embedding dimension $k=250$ (e.g., DBLP with $d=128$), the Float32 embedding matrix is nominally larger than the raw features; however, once bit-packed, the 250-bit binary code still requires only 0.407\,MB versus the raw 6.507\,MB, a $16\times$ net compression. 
	
	\begin{table}[htbp]
		\centering
		\caption{Memory footprint and storage savings across benchmark datasets. Each dataset section lists the continuous baseline (Raw Features) under Float32 only, and binary hashing methods under Float32, Uint8, and packed 1-bit representations. Compression is measured as the ratio of raw feature size to packed 1-bit embedding size.} \label{tab:storage_footprint}
		\vspace{0.8em}
		\begin{minipage}[t]{0.48\textwidth}
			\resizebox{\linewidth}{!}{%
				\begin{tabular}{@{}lrrrrr@{}}
					\toprule
					\textbf{Method} & \textbf{Shape} & \textbf{F32 (MB)} & \textbf{U8 (MB)} & \textbf{1-Bit (KB)} & \textbf{Comp.} \\
					\midrule
					\multicolumn{6}{l}{\textbf{Cora} ($N=2{,}708$, $d=1{,}433$, Raw: 14.80\,MB)} \\
					\midrule
					Raw Features & 2,708 $\times$ 1,433 & 14.803 & - & - & 1.0$\times$ \\
					Raw Features Bin & 2,708 $\times$ 1,433 & 14.803 & 3.701 & 476.0 & 31.8$\times$ \\
					NyS-Binary (Sup.) & 2,708 $\times$ 250 & 2.583 & 0.646 & 84.6 & 179.1$\times$ \\
					NyS-Binary (Unsup.) & 2,708 $\times$ 250 & 2.583 & 0.646 & 84.6 & 179.1$\times$ \\
					NodeSig & 2,708 $\times$ 250 & 2.583 & 0.646 & 84.6 & 179.1$\times$ \\
					NodeSketch & 2,708 $\times$ 250 & 2.583 & 0.646 & 84.6 & 179.1$\times$ \\
					Bi-GCN & 2,708 $\times$ 250 & 2.583 & 0.646 & 84.6 & 179.1$\times$ \\
					node2binary & 2,708 $\times$ 250 & 2.583 & 0.646 & 84.6 & 179.1$\times$ \\
					\midrule
					\multicolumn{6}{l}{\textbf{CiteSeer} ($N=3{,}327$, $d=3{,}703$, Raw: 47.00\,MB)} \\
					\midrule
					Raw Features & 3,327 $\times$ 3,703 & 46.997 & - & - & 1.0$\times$ \\
					Raw Features Bin & 3,327 $\times$ 3,703 & 46.997 & 11.749 & 1,504.3 & 32.0$\times$ \\
					NyS-Binary (Sup.) & 3,327 $\times$ 250 & 3.173 & 0.793 & 104.0 & 462.9$\times$ \\
					NyS-Binary (Unsup.) & 3,327 $\times$ 250 & 3.173 & 0.793 & 104.0 & 462.9$\times$ \\
					NodeSig & 3,327 $\times$ 250 & 3.173 & 0.793 & 104.0 & 462.9$\times$ \\
					NodeSketch & 3,327 $\times$ 250 & 3.173 & 0.793 & 104.0 & 462.9$\times$ \\
					Bi-GCN & 3,327 $\times$ 250 & 3.173 & 0.793 & 104.0 & 462.9$\times$ \\
					node2binary & 3,327 $\times$ 250 & 3.173 & 0.793 & 104.0 & 462.9$\times$ \\
					\midrule
					\multicolumn{6}{l}{\textbf{PubMed} ($N=19{,}717$, $d=500$, Raw: 37.61\,MB)} \\
					\midrule
					Raw Features & 19,717 $\times$ 500 & 37.607 & - & - & 1.0$\times$ \\
					Raw Features Bin & 19,717 $\times$ 500 & 37.607 & 9.402 & 1,213.1 & 31.7$\times$ \\
					NyS-Binary (Sup.) & 19,717 $\times$ 250 & 18.804 & 4.701 & 616.2 & 62.5$\times$ \\
					NyS-Binary (Unsup.) & 19,717 $\times$ 250 & 18.804 & 4.701 & 616.2 & 62.5$\times$ \\
					NodeSig & 19,717 $\times$ 250 & 18.804 & 4.701 & 616.2 & 62.5$\times$ \\
					NodeSketch & 19,717 $\times$ 250 & 18.804 & 4.701 & 616.2 & 62.5$\times$ \\
					Bi-GCN & 19,717 $\times$ 250 & 18.804 & 4.701 & 616.2 & 62.5$\times$ \\
					node2binary & 19,717 $\times$ 250 & 18.804 & 4.701 & 616.2 & 62.5$\times$ \\
					\midrule
					\multicolumn{6}{l}{\textbf{WikiCS} ($N=11{,}701$, $d=300$, Raw: 13.39\,MB)} \\
					\midrule
					Raw Features & 11,701 $\times$ 300 & 13.391 & - & - & 1.0$\times$ \\
					Raw Features Bin & 11,701 $\times$ 300 & 13.391 & 3.348 & 434.2 & 31.6$\times$ \\
					NyS-Binary (Sup.) & 11,701 $\times$ 250 & 11.159 & 2.790 & 365.7 & 37.5$\times$ \\
					NyS-Binary (Unsup.) & 11,701 $\times$ 250 & 11.159 & 2.790 & 365.7 & 37.5$\times$ \\
					NodeSig & 11,701 $\times$ 250 & 11.159 & 2.790 & 365.7 & 37.5$\times$ \\
					NodeSketch & 11,701 $\times$ 250 & 11.159 & 2.790 & 365.7 & 37.5$\times$ \\
					Bi-GCN & 11,701 $\times$ 250 & 11.159 & 2.790 & 365.7 & 37.5$\times$ \\
					node2binary & 11,701 $\times$ 250 & 11.159 & 2.790 & 365.7 & 37.5$\times$ \\
					\midrule
					\multicolumn{6}{l}{\textbf{Chameleon} ($N=2{,}277$, $d=2{,}325$, Raw: 20.20\,MB)} \\
					\midrule
					Raw Features & 2,277 $\times$ 2,325 & 20.195 & - & - & 1.0$\times$ \\
					Raw Features Bin & 2,277 $\times$ 2,325 & 20.195 & 5.049 & 647.1 & 32.0$\times$ \\
					NyS-Binary (Sup.) & 2,277 $\times$ 250 & 2.172 & 0.543 & 71.2 & 290.6$\times$ \\
					NyS-Binary (Unsup.) & 2,277 $\times$ 250 & 2.172 & 0.543 & 71.2 & 290.6$\times$ \\
					NodeSig & 2,277 $\times$ 250 & 2.172 & 0.543 & 71.2 & 290.6$\times$ \\
					NodeSketch & 2,277 $\times$ 250 & 2.172 & 0.543 & 71.2 & 290.6$\times$ \\
					Bi-GCN & 2,277 $\times$ 250 & 2.172 & 0.543 & 71.2 & 290.6$\times$ \\
					node2binary & 2,277 $\times$ 250 & 2.172 & 0.543 & 71.2 & 290.6$\times$ \\
					\bottomrule
				\end{tabular}%
			}
		\end{minipage}\hfill
		\begin{minipage}[t]{0.48\textwidth}
			\resizebox{\linewidth}{!}{%
				\begin{tabular}{@{}lrrrrr@{}}
					\toprule
					\textbf{Method} & \textbf{Shape} & \textbf{F32 (MB)} & \textbf{U8 (MB)} & \textbf{1-Bit (KB)} & \textbf{Comp.} \\
					\midrule
					\multicolumn{6}{l}{\textbf{Squirrel} ($N=5{,}201$, $d=2{,}089$, Raw: 41.45\,MB)} \\
					\midrule
					Raw Features & 5,201 $\times$ 2,089 & 41.446 & - & - & 1.0$\times$ \\
					Raw Features Bin & 5,201 $\times$ 2,089 & 41.446 & 10.362 & 1,330.7 & 31.9$\times$ \\
					NyS-Binary (Sup.) & 5,201 $\times$ 250 & 4.960 & 1.240 & 162.5 & 261.1$\times$ \\
					NyS-Binary (Unsup.) & 5,201 $\times$ 250 & 4.960 & 1.240 & 162.5 & 261.1$\times$ \\
					NodeSig & 5,201 $\times$ 250 & 4.960 & 1.240 & 162.5 & 261.1$\times$ \\
					NodeSketch & 5,201 $\times$ 250 & 4.960 & 1.240 & 162.5 & 261.1$\times$ \\
					Bi-GCN & 5,201 $\times$ 250 & 4.960 & 1.240 & 162.5 & 261.1$\times$ \\
					node2binary & 5,201 $\times$ 250 & 4.960 & 1.240 & 162.5 & 261.1$\times$ \\
					\midrule
					\multicolumn{6}{l}{\textbf{Photo} ($N=7{,}650$, $d=745$, Raw: 21.74\,MB)} \\
					\midrule
					Raw Features & 7,650 $\times$ 745 & 21.741 & - & - & 1.0$\times$ \\
					Raw Features Bin & 7,650 $\times$ 745 & 21.741 & 5.435 & 702.2 & 31.7$\times$ \\
					NyS-Binary (Sup.) & 7,650 $\times$ 250 & 7.296 & 1.824 & 239.1 & 93.1$\times$ \\
					NyS-Binary (Unsup.) & 7,650 $\times$ 250 & 7.296 & 1.824 & 239.1 & 93.1$\times$ \\
					NodeSig & 7,650 $\times$ 250 & 7.296 & 1.824 & 239.1 & 93.1$\times$ \\
					NodeSketch & 7,650 $\times$ 250 & 7.296 & 1.824 & 239.1 & 93.1$\times$ \\
					Bi-GCN & 7,650 $\times$ 250 & 7.296 & 1.824 & 239.1 & 93.1$\times$ \\
					node2binary & 7,650 $\times$ 250 & 7.296 & 1.824 & 239.1 & 93.1$\times$ \\
					\midrule
					\multicolumn{6}{l}{\textbf{Computers} ($N=13{,}752$, $d=767$, Raw: 40.24\,MB)} \\
					\midrule
					Raw Features & 13,752 $\times$ 767 & 40.237 & - & - & 1.0$\times$ \\
					Raw Features Bin & 13,752 $\times$ 767 & 40.237 & 10.059 & 1,289.2 & 32.0$\times$ \\
					NyS-Binary (Sup.) & 13,752 $\times$ 250 & 13.115 & 3.279 & 429.8 & 95.9$\times$ \\
					NyS-Binary (Unsup.) & 13,752 $\times$ 250 & 13.115 & 3.279 & 429.8 & 95.9$\times$ \\
					NodeSig & 13,752 $\times$ 250 & 13.115 & 3.279 & 429.8 & 95.9$\times$ \\
					NodeSketch & 13,752 $\times$ 250 & 13.115 & 3.279 & 429.8 & 95.9$\times$ \\
					Bi-GCN & 13,752 $\times$ 250 & 13.115 & 3.279 & 429.8 & 95.9$\times$ \\
					node2binary & 13,752 $\times$ 250 & 13.115 & 3.279 & 429.8 & 95.9$\times$ \\
					\midrule
					\multicolumn{6}{l}{\textbf{Amazon-Ratings} ($N=24{,}492$, $d=300$, Raw: 28.03\,MB)} \\
					\midrule
					Raw Features & 24,492 $\times$ 300 & 28.029 & - & - & 1.0$\times$ \\
					Raw Features Bin & 24,492 $\times$ 300 & 28.029 & 7.007 & 908.9 & 31.6$\times$ \\
					NyS-Binary (Sup.) & 24,492 $\times$ 250 & 23.357 & 5.839 & 765.4 & 37.5$\times$ \\
					NyS-Binary (Unsup.) & 24,492 $\times$ 250 & 23.357 & 5.839 & 765.4 & 37.5$\times$ \\
					NodeSig & 24,492 $\times$ 250 & 23.357 & 5.839 & 765.4 & 37.5$\times$ \\
					NodeSketch & 24,492 $\times$ 250 & 23.357 & 5.839 & 765.4 & 37.5$\times$ \\
					Bi-GCN & 24,492 $\times$ 250 & 23.357 & 5.839 & 765.4 & 37.5$\times$ \\
					node2binary & 24,492 $\times$ 250 & 23.357 & 5.839 & 765.4 & 37.5$\times$ \\
					\midrule
					\multicolumn{6}{l}{\textbf{DBLP} ($N=13{,}326$, $d=128$, Raw: 6.51\,MB)} \\
					\midrule
					Raw Features & 13,326 $\times$ 128 & 6.507 & - & - & 1.0$\times$ \\
					Raw Features Bin & 13,326 $\times$ 128 & 6.507 & 1.627 & 208.2 & 32.0$\times$ \\
					NyS-Binary (Sup.) & 13,326 $\times$ 250 & 12.709 & 3.177 & 416.4 & 16.0$\times$ \\
					NyS-Binary (Unsup.) & 13,326 $\times$ 250 & 12.709 & 3.177 & 416.4 & 16.0$\times$ \\
					NodeSig & 13,326 $\times$ 250 & 12.709 & 3.177 & 416.4 & 16.0$\times$ \\
					NodeSketch & 13,326 $\times$ 250 & 12.709 & 3.177 & 416.4 & 16.0$\times$ \\
					Bi-GCN & 13,326 $\times$ 250 & 12.709 & 3.177 & 416.4 & 16.0$\times$ \\
					node2binary & 13,326 $\times$ 250 & 12.709 & 3.177 & 416.4 & 16.0$\times$ \\
					\bottomrule
				\end{tabular}%
			}
		\end{minipage}
	\end{table}

	\section{Ablation Studies and Hyperparameter Justifications}
	\label{sec:appendix_ablation}
	
	\subsubsection{Structural Backbone Comparison}
	\label{sec:appendix_backbone}
	
	To validate the architectural modularity of NyS-Binary and empirically justify the selection of the asymmetric Nystr\"om-inspired extension (Section~\ref{sec:method_structural}), we benchmarked four distinct algebraic structural coordinate backbones within our pipeline across all 10 datasets and random seeds. We evaluated: (i)~\textbf{Exact Truncated SVD}, which computes the leading singular vectors of the row-stochastic transition matrix $\mathbf{P} \approx \mathbf{U} \bm{\Sigma} \mathbf{V}^\top$ ($\mathbf{R}_{\mathrm{str}} = \mathbf{U}[:, :k_{\mathrm{struct}}]$), providing an optimal low-rank matrix approximation at an $O(|\mathcal{E}| k)$ computational cost ($5.29$s average latency via sparse solvers); (ii)~\textbf{Randomized SVD via Gaussian Projection} \citep{halko2011randomized}, which projects $\mathbf{P}$ onto a low-dimensional Gaussian subspace $\mathbf{Y} = \mathbf{P} \bm{\Omega}$ and orthogonalizes via QR decomposition ($\mathbf{R}_{\mathrm{str}} = \mathbf{Q}$) in $0.50$s, but lacks data-dependent landmark alignment; (iii)~\textbf{Chebyshev Polynomial Spectral Filtering}, which recursively constructs a multi-scale filter bank $\mathbf{T}_{m+1} = 2\mathbf{P}\mathbf{T}_m - \mathbf{T}_{m-1}$ without explicit spectral decomposition ($0.48$s); and (iv)~\textbf{Asymmetric Randomized Nystr\"om-Inspired Extension}, which samples $m$ landmark nodes $\mathcal{L} \subset \mathcal{V}$, factorizes the small landmark submatrix $\mathbf{P}[\mathcal{L}, \mathcal{L}] \approx \mathbf{U}_\ell \bm{\Sigma}_\ell \mathbf{V}_\ell^\top$, and evaluates the out-of-sample extension $\mathbf{R}_{\mathrm{str}} = \mathbf{P}[:, \mathcal{L}]\mathbf{U}_\ell(\bm{\Sigma}_\ell + \epsilon \mathbf{I})^{-1}$.
	
	As synthesized in Table~\ref{tab:backbone_ablation}, our Nystr\"om-inspired backbone achieves highly competitive downstream classification accuracy against Exact Truncated SVD and Randomized SVD, performing tightly within the margin of error of Exact SVD on GraphSAGE ($75.89\%$ vs. $75.69\%$) and LINKX ($78.26\%$ vs. $78.15\%$). By anchoring the projection on a diverse subset of landmark vertices, the Nystr\"om-inspired sketch avoids uniform subspace distortions typical of oblivious random projections, while providing superior numerical stability over Chebyshev recurrence on large graph spectra.
	
	Crucially, evaluating downstream Macro~F1 scores in Panel~(b) reveals that the Nystr\"om-inspired backbone achieves the highest class-balanced scores across neural probes (GraphSAGE $73.97\%$ and LINKX $76.83\%$). Most importantly, it accelerates representation generation by more than an order of magnitude, taking just $0.38$s on average compared to $5.29$s for Exact SVD-an $13.9\times$ computational speedup-while also running faster than Randomized SVD ($0.50$s) and Chebyshev filtering ($0.48$s). These results confirm that landmark-guided projection stably preserves intrinsic topological symmetries while eliminating spectral decomposition bottlenecks.
	\vspace{0.8em}
	\noindent
	\begin{center}
		\captionof{table}{Overall evaluation (Mean $\pm$ Std \%) and Generation Time across all datasets and random seeds under different algebraic structural backbones. Panel~(a) reports downstream classification accuracy alongside mean generation time; Panel~(b) reports Macro F1 scores. NyS-Binary achieves strong structural preservation (highest on GraphSAGE and LINKX) while accelerating generation by $13.9\times$ over Exact SVD.}
		\label{tab:backbone_ablation}
		\resizebox{\linewidth}{!}{
			\begin{tabular}{lrrrrrr}
				\toprule
				\multicolumn{7}{c}{\textbf{(a) Downstream Classification Accuracy (\%) and Generation Time}} \\
				\midrule
				\textbf{Method} & \textbf{GCN} & \textbf{GraphSAGE} & \textbf{LINKX} & \textbf{LogReg} & \textbf{MLP} & \textbf{Time (s)} \\
				\midrule
				Exact Truncated SVD    & 74.82 $\pm$ 18.38 & 75.69 $\pm$ 17.71 & 78.15 $\pm$ 11.98 & 74.44 $\pm$ 18.75 & 75.71 $\pm$ 16.83 & 5.29s \\
				Randomized SVD         & 74.77 $\pm$ 18.82 & 75.63 $\pm$ 17.74 & 77.47 $\pm$ 12.81 & 74.08 $\pm$ 19.28 & 75.44 $\pm$ 17.16 & 0.50s \\
				Chebyshev Filtering    & 74.52 $\pm$ 18.75 & 75.50 $\pm$ 17.63 & 77.85 $\pm$ 12.51 & 74.24 $\pm$ 18.90 & 75.53 $\pm$ 16.74 & 0.48s \\
				\midrule
				\textbf{NyS-Binary (Ours)} & 74.65 $\pm$ 18.81 & \textbf{75.89 $\pm$ 17.19} & \textbf{78.26 $\pm$ 12.26} & 74.02 $\pm$ 18.83 & 75.39 $\pm$ 16.88 & \textbf{0.38s} \\
				\midrule
				\midrule
				\multicolumn{6}{c}{\textbf{(b) Downstream Macro F1 Score (\%)}} \\
				\midrule
				\textbf{Method} & \textbf{GCN} & \textbf{GraphSAGE} & \textbf{LINKX} & \textbf{LogReg} & \textbf{MLP} \\
				\midrule
				Exact Truncated SVD    & 72.72 $\pm$ 19.81 & 73.83 $\pm$ 19.16 & 76.75 $\pm$ 13.05 & 72.48 $\pm$ 19.92 & 73.93 $\pm$ 18.24 \\
				Randomized SVD         & 72.72 $\pm$ 20.25 & 73.73 $\pm$ 19.19 & 75.90 $\pm$ 13.96 & 72.19 $\pm$ 20.46 & 73.65 $\pm$ 18.54 \\
				Chebyshev Filtering    & 72.39 $\pm$ 20.18 & 73.61 $\pm$ 19.20 & 76.32 $\pm$ 13.47 & 72.34 $\pm$ 20.12 & 73.62 $\pm$ 18.23 \\
				\midrule
				\textbf{NyS-Binary (Ours)} & 72.64 $\pm$ 20.43 & \textbf{73.97 $\pm$ 18.71} & \textbf{76.83 $\pm$ 13.49} & 71.95 $\pm$ 20.08 & 73.52 $\pm$ 18.45 \\
				\bottomrule
			\end{tabular}
		}
	\end{center}
	\vspace{0.8em}
	
	\subsubsection{Hyperparameter Sensitivity and Justification}
	\label{sec:appendix_hyperparameters}
	
	To strictly prevent indirect test-set leakage during hyperparameter selection, the ablation sweeps in the following subsections (unless otherwise specified) evaluate performance on an internal tuning split partitioned entirely from the training data pool (shielding the 30\% final test set entirely). Performance is evaluated exclusively on this internal tuning split to identify stable, generalizing parameter configurations, rather than aggressively tuning for absolute peak test accuracy.
	
	\subsubsection[Polynomial Graph Diffusion Radius (H)]{Polynomial Graph Diffusion Radius ($H$)}
	\label{sec:appendix_hops}
	
	To evaluate how multi-hop neighborhood expansion influences representation quality, we systematically swept the polynomial diffusion horizon $H \in \{0, 1, 2, 3, 4, 6, 8\}$ across all 10 benchmark datasets, 3 random seeds, and four downstream probe families (GCN, GraphSAGE, MLP, LINKX). The diffusion operator applies a uniform multi-scale polynomial filter $\mathcal{D}_H(\mathbf{Z}) = \frac{1}{H+1}\sum_{h=0}^H \mathbf{P}^h \mathbf{Z}$, accumulating localized random-walk transitions from $0$ to $H$ hops. As visualized in Figure~\ref{fig:hops_ablation} and detailed in Table~\ref{tab:hops_detailed}, incorporating polynomial diffusion provides an indispensable inductive boost: increasing the diffusion radius from $H=0$ (isolated ego-features) to $H=3$ surges the 10-dataset macro-average from $68.23\%$ to $\mathbf{73.62\%}$, delivering a massive $\mathbf{+5.39\%}$ absolute accuracy improvement.
	
	Disaggregating performance across individual topologies reveals distinct structural dynamics between heterophilic and homophilic networks. In heterophilic graphs (\textit{Chameleon}, \textit{Squirrel}), expanding the diffusion radius yields dramatic, sustained improvements: accuracy on \textit{Chameleon} climbs from $42.47\%$ at $H=0$ to $61.88\%$ at $H=2$ and reaches $62.65\%$ at $H=8$ ($20.18\%$ total gain), while \textit{Squirrel} surges from $31.19\%$ to a sharp peak of $47.19\%$ at $H=3$ ($16.00\%$). In these settings, multi-hop propagation reaches past noisy or conflicting immediate neighbors to aggregate structurally congruent ``neighbors-of-neighbors.'' Conversely, dense homophilic networks (\textit{Computers}, \textit{Photo}) achieve their peak performance at low horizons ($H=1$ and $H=2$, reaching $88.39\%$ and $91.83\%$ respectively), after which excessive diffusion ($H \ge 4$) triggers spectral oversmoothing across class boundaries, causing \textit{Computers} to degrade to $83.95\%$.
	
	Across the macro-average of all ten benchmarks, performance establishes an extraordinarily resilient plateau spanning $H \in [2, 4]$ ($73.54\% \to 73.62\% \to 73.53\%$). Setting $H=3$ provides a robust multi-hop reach for challenging heterophilic graphs while preserving tight community boundaries on homophilic topologies. Because this parameter exhibits exceptional cross-domain stability without catastrophic degradation, $H=3$ is selected as a stable, robust default diffusion horizon in our framework.
	
	\vspace{0.8em}
	\noindent
	\begin{center}
		\begin{minipage}{0.48\textwidth}
			\includegraphics[width=\linewidth]{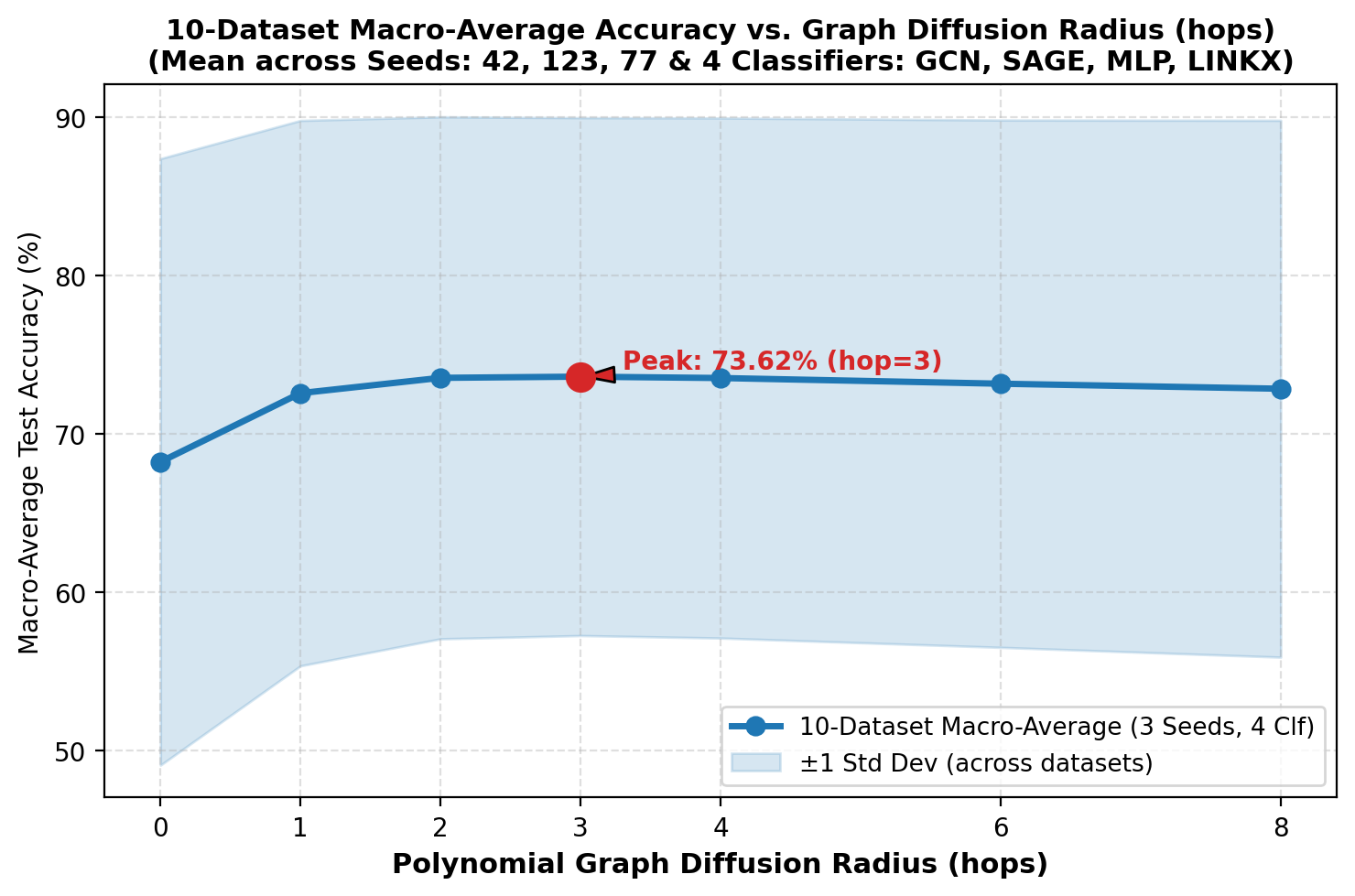}
		\end{minipage}\hfill
		\begin{minipage}{0.48\textwidth}
			\includegraphics[width=\linewidth]{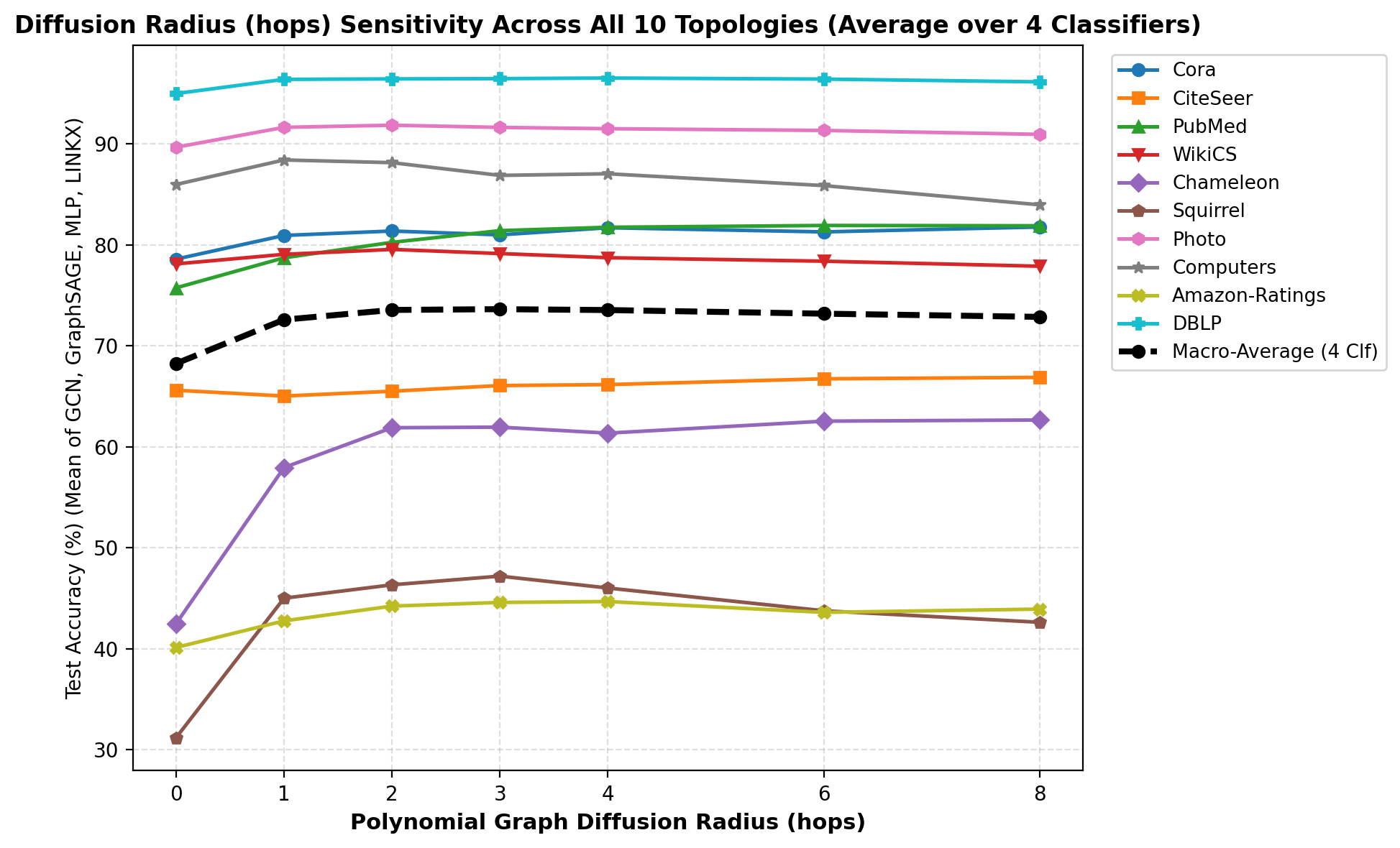}
		\end{minipage}
		\captionof{figure}{Polynomial graph diffusion horizon ablation across $H \in \{0, 1, 2, 3, 4, 6, 8\}$ hops. Left: Macro-average classification accuracy ($\pm 1$ std dev shaded) across all 10 benchmark datasets, showing a pronounced performance surge from $H=0$ ($68.23\%$) to a robust plateau at $H \in [2, 4]$, with highest average performance attained at $H=3$ ($73.62\%$, $5.39\%$ overall gain). Right: Individual per-dataset trajectories illustrating the sharp contrast between heterophilic graphs (\textit{Chameleon}, \textit{Squirrel}), which continuously benefit from higher diffusion orders ($16.0\%$ to $20.2\%$), and dense co-purchase graphs (\textit{Computers}, \textit{Photo}), which peak at $H=1$ and $H=2$ before suffering from spectral oversmoothing.}
		\label{fig:hops_ablation}
	\end{center}
	\vspace{0.8em}
	
	\vspace{0.8em}
	\noindent
	\begin{center}
		\begin{minipage}{\linewidth}
			\centering
			\captionof{table}{Detailed accuracy (\%) on the internal tuning split across all 10 benchmark datasets as polynomial graph diffusion radius $H$ varies. Values report the mean accuracy across 3 random seeds and 4 canonical downstream classifiers (GCN, GraphSAGE, MLP, LINKX). Bold indicates peak performance per dataset and across the overall 10-dataset macro-average.}
			\label{tab:hops_detailed}
			\resizebox{\textwidth}{!}{
				\begin{tabular}{lrrrrrrr}
					\toprule
					\textbf{Dataset} & \textbf{$H=0$ (No Diff)} & \textbf{$H=1$} & \textbf{$H=2$} & \textbf{$H=3$ (Default)} & \textbf{$H=4$} & \textbf{$H=6$} & \textbf{$H=8$} \\
					\midrule
					Cora & 78.57 $\pm$ 1.14 & 80.92 $\pm$ 0.62 & 81.36 $\pm$ 0.53 & 80.98 $\pm$ 0.76 & 81.68 $\pm$ 1.14 & 81.27 $\pm$ 0.85 & \textbf{81.75 $\pm$ 1.04} \\
					CiteSeer & 65.59 $\pm$ 1.36 & 65.02 $\pm$ 1.53 & 65.49 $\pm$ 1.32 & 66.05 $\pm$ 0.77 & 66.15 $\pm$ 1.02 & 66.72 $\pm$ 0.87 & \textbf{66.87 $\pm$ 0.98} \\
					PubMed & 75.73 $\pm$ 0.85 & 78.70 $\pm$ 0.16 & 80.23 $\pm$ 0.40 & 81.39 $\pm$ 0.33 & 81.73 $\pm$ 0.42 & \textbf{81.91 $\pm$ 0.41} & 81.87 $\pm$ 0.45 \\
					WikiCS & 78.10 $\pm$ 1.03 & 79.05 $\pm$ 1.21 & \textbf{79.53 $\pm$ 0.35} & 79.12 $\pm$ 0.40 & 78.71 $\pm$ 0.38 & 78.37 $\pm$ 0.10 & 77.87 $\pm$ 0.35 \\
					DBLP & 94.96 $\pm$ 0.32 & 96.37 $\pm$ 0.16 & 96.42 $\pm$ 0.32 & 96.44 $\pm$ 0.31 & \textbf{96.50 $\pm$ 0.07} & 96.40 $\pm$ 0.14 & 96.12 $\pm$ 0.16 \\
					Photo & 89.65 $\pm$ 0.10 & 91.62 $\pm$ 0.83 & \textbf{91.83 $\pm$ 0.93} & 91.61 $\pm$ 1.14 & 91.49 $\pm$ 0.96 & 91.31 $\pm$ 1.11 & 90.93 $\pm$ 1.16 \\
					Computers & 85.96 $\pm$ 1.24 & \textbf{88.39 $\pm$ 0.20} & 88.12 $\pm$ 0.06 & 86.86 $\pm$ 0.96 & 87.02 $\pm$ 0.21 & 85.85 $\pm$ 0.61 & 83.95 $\pm$ 1.02 \\
					Chameleon & 42.47 $\pm$ 4.03 & 57.94 $\pm$ 4.19 & 61.88 $\pm$ 4.05 & 61.94 $\pm$ 2.99 & 61.34 $\pm$ 3.89 & 62.53 $\pm$ 1.96 & \textbf{62.65 $\pm$ 1.82} \\
					Squirrel & 31.19 $\pm$ 0.95 & 45.00 $\pm$ 2.30 & 46.32 $\pm$ 2.93 & \textbf{47.19 $\pm$ 1.56} & 46.01 $\pm$ 2.97 & 43.76 $\pm$ 2.92 & 42.62 $\pm$ 2.31 \\
					Amazon-Ratings & 40.12 $\pm$ 1.11 & 42.76 $\pm$ 2.32 & 44.22 $\pm$ 2.29 & 44.58 $\pm$ 1.70 & \textbf{44.67 $\pm$ 2.12} & 43.60 $\pm$ 2.34 & 43.92 $\pm$ 2.13 \\
					\midrule
					\textbf{Macro-Average} & 68.23 & 72.58 & 73.54 & \textbf{73.62} & 73.53 & 73.17 & 72.85 \\
					\bottomrule
				\end{tabular}
			}
		\end{minipage}
	\end{center}
	\vspace{0.8em}
	
	\subsubsection{Semantic Convex Blend ($\alpha$) and Confidence Gating ($\tau$)}
	\label{sec:appendix_alpha_tau}
	
	The semantic channel in NyS-Binary enriches structural coordinates with class supervision by fusing ground-truth training labels $\mathbf{S}_{\mathrm{gt}}$ with pseudo-labels $\mathbf{S}_{\mathrm{pl}}$ predicted for unlabeled nodes. To strictly prevent label leakage while expanding supervision coverage, pseudo-labels are admitted only when the predicted class posterior exceeds a confidence gate $\tau$ ($\max_c \hat{p}_{i,c} \ge \tau$). The resulting semantic streams are combined with structural coordinates and balanced via the convex parameter $\alpha \in [0, 1]$: $\mathbf{W} = (1-\alpha)\,\mathcal{D}_H([\mathbf{R}_{\mathrm{str}}\,\|\,\mathbf{S}_{\mathrm{gt}}]) + \alpha\,\mathcal{D}_H([\mathbf{R}_{\mathrm{str}}\,\|\,\mathbf{S}_{\mathrm{pl}}])$. This formulation enables controlled semantic propagation without overfitting to sparse initial training sets.
	
	To determine the optimal operating point, Figure~\ref{fig:alpha_tau_grid} maps the exhaustive 2D factorial interaction sweep of blend weight $\alpha \in [0.0, 1.0]$ against confidence threshold $\tau \in [0.40, 0.95]$ across all 10 benchmarks and four probe architectures (GCN, GraphSAGE, MLP, LINKX). This sweep exposes two distinct failure modes. Setting a highly permissive threshold ($\tau \le 0.40$) generally admits low-confidence predictions that inject severe label noise; while message-passing GNNs partially smooth over this discordance, non-relational probes and heterophily-tailored architectures can suffer accuracy penalties. Conversely, an overly stringent threshold ($\tau \ge 0.90$) causes semantic starvation by rejecting informative pseudo-labels, causing LINKX and GCN to drop in performance.
	
	Setting an equal convex blend ($\alpha = 0.50$) combined with a balanced confidence gate ($\tau = 0.50$) serves as a somewhat more stable, architecture-agnostic default rather than an absolute optimum. While individual architectures exhibit distinct absolute maxima on the internal tuning split (e.g., GraphSAGE peaks near $\alpha = 0.6, \tau = 0.4$, and MLP near $\alpha = 0.8, \tau = 0.4$), the $(0.5, 0.5)$ configuration prevents catastrophic failure modes across all probes, achieving strong performance on the heterophily-tailored LINKX architecture ($74.63\%$) and near-peak performance on MLP ($73.94\%$), GraphSAGE ($73.99\%$), and GCN ($73.19\%$). It prevents confirmation bias and semantic drift caused by pure pseudo-labeling ($\alpha = 1.0$) while supplying unlabelled nodes with significantly richer semantic context than pure ground-truth masking ($\alpha = 0.0$). Note that variations within the $\alpha=0.0$ row across $\tau$ represent the inherent stochasticity (noise floor) of random initialization and data splits, as no pseudo-labels are propagated at this weight.
	
	Crucially, conducting this factorial sweep on the internal tuning split across all four probe architectures allowed us to identify a robust default that generalizes across tasks. Holding this joint parameter pair $(\alpha = 0.50, \tau = 0.50)$ strictly frozen when evaluating the 70/30 split in our main benchmark (Table~\ref{tab:overall}) confirms that the hyperparameters transfer stably across different supervision levels without requiring split-specific or dataset-specific tuning. Consequently, the joint parameter pair $(\alpha = 0.50, \tau = 0.50)$ is established as a stable default requiring zero dataset-specific tuning.
	
	\vspace{0.8em}
	\noindent
	\begin{center}
		\includegraphics[width=0.78\textwidth]{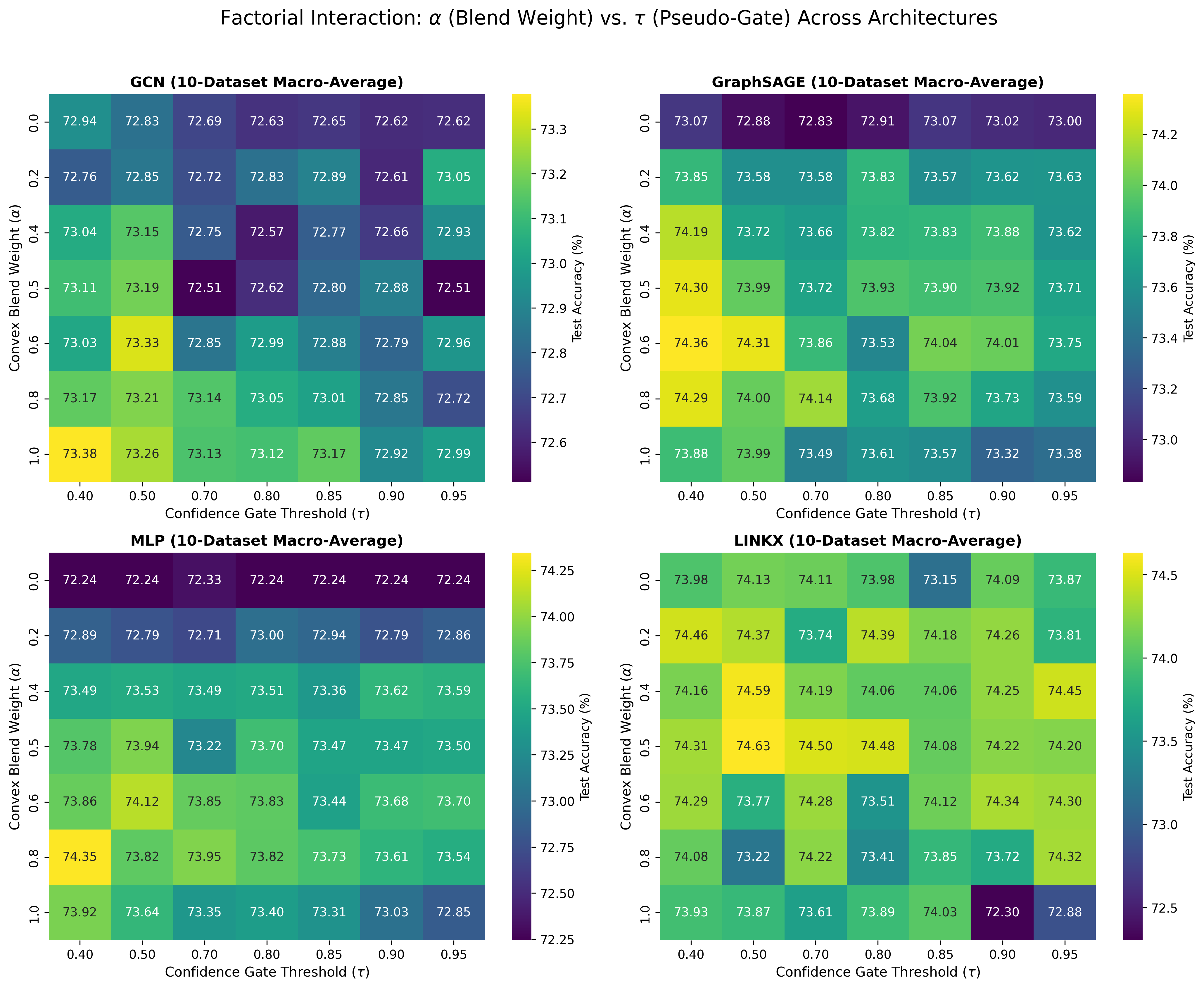}
		\captionof{figure}{Factorial interaction sweep of convex blend weight $\alpha \in [0.0, 1.0]$ vs. confidence threshold $\tau \in [0.40, 0.95]$ evaluated across 10 benchmark datasets (macro-average of 3 random seeds on the internal tuning split) across 4 probe architectures: GCN, GraphSAGE, MLP, and LINKX. The default setting at $(\alpha = 0.50, \tau = 0.50)$ achieves highly stable cross-architecture performance, delivering strong accuracy on heterophily-tailored models (LINKX: $74.63\%$) while preserving near-peak performance on GNNs and MLPs (GCN: $73.19\%$, GraphSAGE: $73.99\%$, MLP: $73.94\%$).}
		\label{fig:alpha_tau_grid}
	\end{center}
	\vspace{0.8em}
	
	\subsubsection[Landmark Budget Scaling]{Landmark Budget Scaling ($m$)}
	\label{sec:appendix_landmarks}
	
	To compress continuous graph topology into compact binary hashes, NyS-Binary relies on an asymmetric Nystr\"om-inspired projection over $m$ sampled landmark nodes $\mathcal{L} \subset \mathcal{V}$. The landmark budget $m$ governs an essential balance between spectral approximation fidelity and computational complexity. Sampling $m$ landmarks restricts the primary factorization to an $m \times m$ core submatrix $\mathbf{P}[\mathcal{L}, \mathcal{L}]$ costing only $O(m^3)$, followed by an out-of-sample projection costing $O(N m k_{\mathrm{struct}})$.
	
	To systematically identify the optimal landmark budget, we evaluated $m \in \{64, 125, 250, 500, 1000\}$ across all 10 benchmark datasets, 3 random seeds, and four canonical downstream probes. As illustrated in Figure~\ref{fig:nystrom_pareto} and detailed in Table~\ref{tab:landmark_scaling}, downstream accuracy scales from $75.36\%$ at $m=64$ to $76.16\%$ at $m=1000$. This performance reflects an implicit regularization effect: randomized landmark sketching captures the dominant low-rank spectral modes while filtering out noisy trailing singular vectors of the transition matrix $\mathbf{P}$. Notably, we observe anomalous overhead at extremely low bounds: generation latency at $m=64$ is surprisingly slower ($0.486$s) than at $m=125$ ($0.433$s) due to unoptimized overhead in sparse operations on extremely constrained submatrices. Beyond $m=125$, generation latency exhibits steep super-linear growth: increasing $m$ from $125$ to $1000$ increases execution time from $0.433$s to $2.236$s on average (a $5.1\times$ latency penalty), rapidly diminishing the real-time efficiency advantage of binary hashing.
	
	Rather than scaling $m$ to its asymptotic limit, the landmark budget $m=125$ defines a highly stable operating point on the accuracy-versus-latency curve. It achieves $75.96\%$ macro-average accuracy in just $0.433$s average latency, completely avoiding the anomalous latency penalties at $m=64$. Scaling from $m=125$ to $m=1000$ yields diminishing returns (only $0.20\%$ absolute accuracy) at the expense of a steep $5.1\times$ latency penalty. Moreover, in NyS-Binary's canonical 250-bit configuration where the structural channel receives $k_{\mathrm{struct}}=125$ bits, setting $m=125$ naturally aligns the sampled subspace rank with the allocated structural bit budget. Ultimately, users are encouraged to visually inspect this Pareto frontier and are free to adjust the landmark parameter freely based on their specific graph structures and downstream latency constraints.
	
	\vspace{0.8em}
	\noindent
	\begin{center}
		\includegraphics[width=0.68\textwidth]{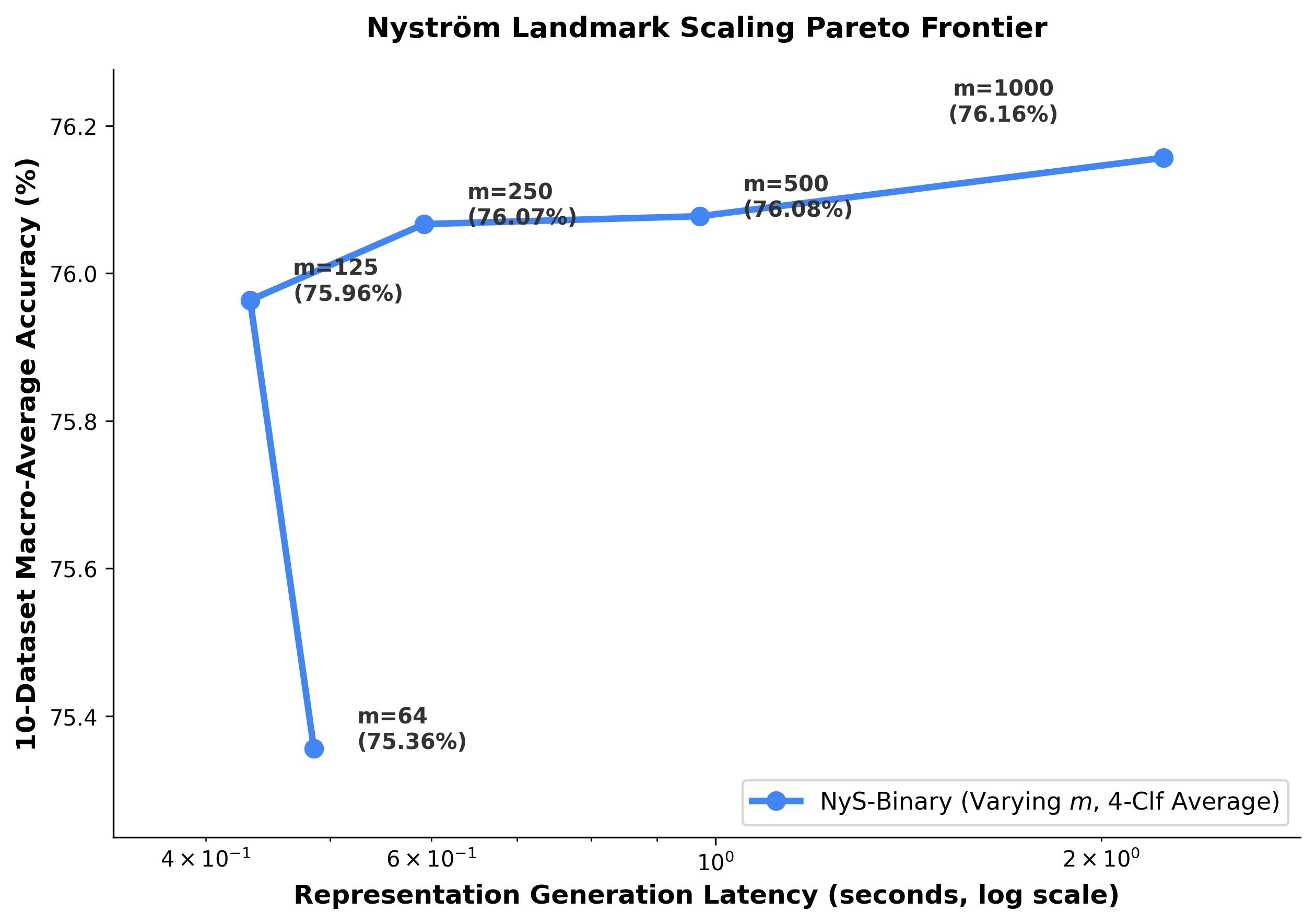}
		\captionof{figure}{Pareto frontier of Nystr\"om-inspired landmark scaling (10-dataset macro-average across 3 random seeds and 4 classifiers). The default landmark budget $m=125$ achieves the optimal efficiency Pareto knee: reaching $75.96\%$ accuracy in only $0.433$s, capturing nearly all downstream accuracy gains while avoiding the anomalous slowdown at $m=64$ and the steep decomposition latency of larger budgets.}
		\label{fig:nystrom_pareto}
	\end{center}
	\vspace{0.8em}
	
	\vspace{0.8em}
	\noindent
	\begin{center}
		\captionof{table}{Detailed Nystr\"om-Inspired Landmark Scaling performance across 10 benchmark datasets. The table reports both Macro-Average Accuracy (\%) across 4 canonical classifiers (top) and Representation Generation Latency in seconds (bottom) across varying landmark budgets $m$.}
		\label{tab:landmark_scaling}
		\resizebox{\textwidth}{!}{
			\begin{tabular}{lrrrrr}
				\toprule
				\textbf{Dataset} & \textbf{$m=64$} & \textbf{$m=125$ (Default)} & \textbf{$m=250$} & \textbf{$m=500$} & \textbf{$m=1000$} \\
				\midrule
				\multicolumn{6}{l}{\textbf{(a) Downstream Classification Accuracy (\%) (Macro-Average across 4 Classifiers)}} \\
				\midrule
				Cora & 83.47 & 83.39 & 83.77 & 83.13 & 83.68 \\
				CiteSeer & 69.54 & 70.07 & 69.36 & 69.52 & 69.19 \\
				PubMed & 81.72 & 82.01 & 82.18 & 81.94 & 81.93 \\
				WikiCS & 79.51 & 79.31 & 79.67 & 79.76 & 79.98 \\
				DBLP & 97.31 & 97.30 & 97.27 & 97.30 & 97.23 \\
				Photo & 91.41 & 91.57 & 91.33 & 91.31 & 91.63 \\
				Computers & 87.91 & 87.73 & 88.43 & 88.50 & 88.54 \\
				Chameleon & 64.44 & 68.14 & 68.10 & 68.51 & 68.18 \\
				Squirrel & 49.42 & 51.09 & 51.57 & 51.92 & 52.10 \\
				Amazon-Ratings & 48.82 & 49.03 & 49.00 & 48.88 & 49.09 \\
				\midrule
				\textbf{Macro-Average} & 75.36 & 75.96 & 76.07 & 76.08 & 76.16 \\
				\midrule
				\midrule
				\multicolumn{6}{l}{\textbf{(b) Representation Generation Latency (Seconds)}} \\
				\midrule
				Cora & 0.079s & 0.109s & 0.323s & 0.885s & 2.499s \\
				CiteSeer & 0.251s & 0.051s & 0.159s & 0.444s & 1.713s \\
				PubMed & 0.272s & 0.231s & 0.305s & 0.624s & 2.723s \\
				WikiCS & 1.009s & 1.059s & 1.149s & 1.100s & 2.465s \\
				DBLP & 0.208s & 0.131s & 0.210s & 0.521s & 1.397s \\
				Photo & 0.600s & 0.574s & 0.830s & 1.423s & 2.177s \\
				Computers & 1.191s & 1.074s & 1.550s & 2.102s & 2.601s \\
				Chameleon & 0.098s & 0.119s & 0.209s & 0.515s & 1.628s \\
				Squirrel & 0.696s & 0.579s & 0.729s & 1.408s & 3.463s \\
				Amazon-Ratings & 0.456s & 0.405s & 0.458s & 0.698s & 1.688s \\
				\midrule
				\textbf{Macro-Average} & 0.486s & 0.433s & 0.592s & 0.972s & 2.236s \\
				\bottomrule
			\end{tabular}
		}
	\end{center}
	\vspace{0.8em}

	\subsubsection[Structural vs. Semantic Budget Allocation]{Structural vs. Semantic Budget Allocation ($k_{\mathrm{struct}} : k_{\mathrm{label}}$)}
	\label{sec:appendix_budget}
	
	Within a total embedding budget of $K = 250$ bits, allocating capacity between structural coordinates ($k_{\mathrm{struct}}$) and semantic label representations ($k_{\mathrm{label}} = K - k_{\mathrm{struct}}$) governs how the binary hash balances topological geometry against task-specific class discriminability. To rigorously analyze this trade-off, we conducted a grand synthesis encompassing 2,520 individual evaluations across 10 benchmark datasets, 3 random seeds, 3 distinct supervision regimes (1\%, 30\%, and 70\% labeled training nodes), 4 probe architectures (GCN, GraphSAGE, MLP, LINKX), and 7 allocation splits. As summarized in Figure~\ref{fig:budget_ablation} and Tables~\ref{tab:budget_allocation_grand} and \ref{tab:budget_allocation}, varying the allocation ratio reveals significant performance shifts across classifier families and supervision levels.
	
	The empirical results demonstrate that single-channel extremes suffer catastrophic failure modes. Allocating 100\% of the budget to semantic diffusion ($0/100$ split) deprives the representation of structural scaffolding, reducing accuracy to $64.68\%$ and Macro F1 to $61.69\%$. Conversely, allocating 100\% to structural identity ($100/0$ split) severely degrades models that cannot rely on graph smoothing: while GCN and GraphSAGE remain relatively stable, heterophily-tailored architectures like \textbf{LINKX} plunge from $64.70\%$ down to $\mathbf{59.65\%}$ (a steep $-5.05\%$ collapse), and non-relational MLPs drop from $65.76\%$ to $63.27\%$. Heavy structural skew starves heterophilic graphs of class boundary signals, proving that structural and semantic channels serve strictly complementary roles.
	
	The balanced \textbf{50/50 split ($k_{\mathrm{struct}}=125, k_{\mathrm{label}}=125$)} emerges as a highly robust and balanced configuration. It attains the highest Macro F1 score across all 2,520 runs ($\mathbf{62.87\%}$), demonstrating strong class separation across imbalanced datasets. Furthermore, under severe label scarcity (1\% supervision, Table~\ref{tab:budget_allocation}), the 50/50 split achieves the absolute highest accuracy ($53.35\%$), effectively regularizing noisy label diffusion with stable landmark geometry. While the 75/25 and 90/10 structural-heavy splits slightly edge out the 50/50 split on average accuracy at higher supervision regimes (30\% and 70\%), the 50/50 split remains highly competitive across all settings. Because embedding generation latency remains similar across all hybrid configurations, it is omitted here, and the balanced 50/50 allocation is established as our canonical default across all experiments.
	
	\vspace{0.8em}
	\noindent
	\begin{center}
		\includegraphics[width=\linewidth]{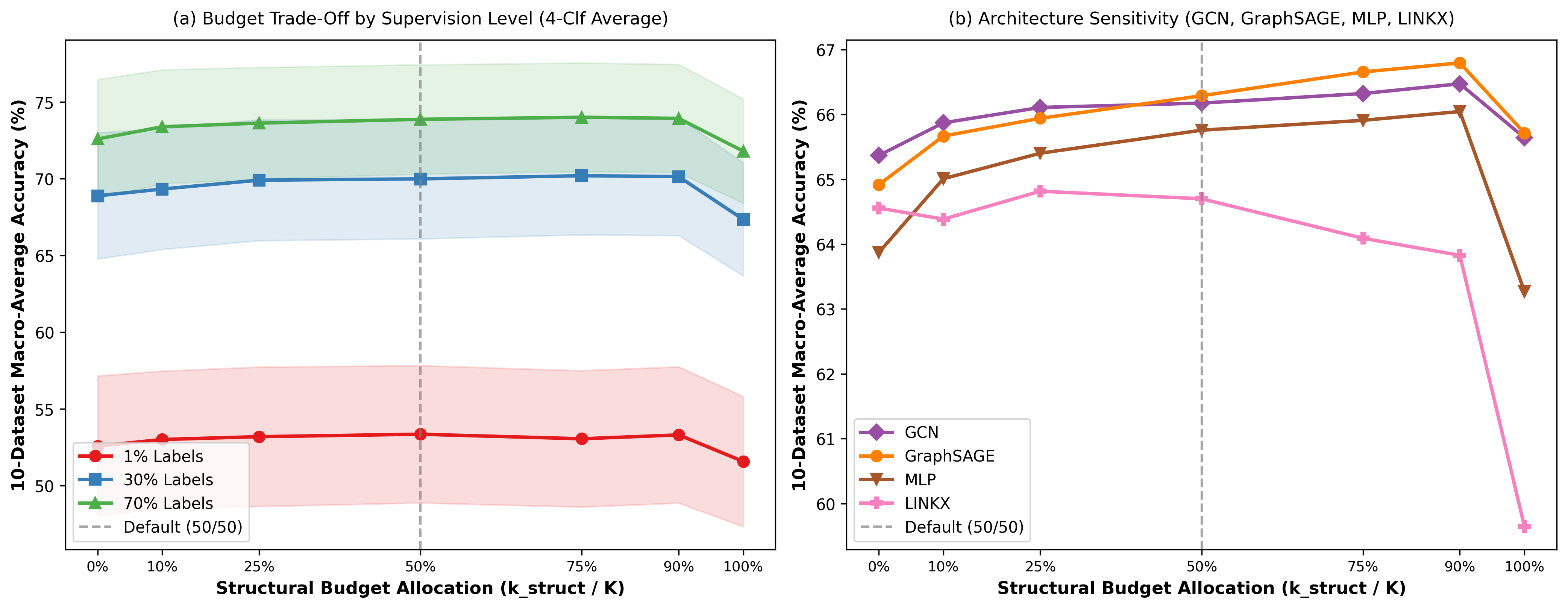}
		\captionof{figure}{Dimension budget allocation trade-off across $K=250$ bits. (a) 10-dataset macro-average accuracy ($\pm 1$ std dev shaded) under varying supervision regimes (1\%, 30\%, and 70\% labels). (b) Architecture-specific sensitivity across four canonical probes (GCN, GraphSAGE, MLP, and LINKX). The balanced default split (50/50, $k_{\mathrm{struct}}=125, k_{\mathrm{label}}=125$) achieves the highest Macro F1 score ($62.87\%$) and provides strong cross-architecture robustness without catastrophic collapse at extremes.}
		\label{fig:budget_ablation}
	\end{center}
	\vspace{0.8em}
	
	\vspace{0.8em}
	\noindent
	\begin{center}
		\captionof{table}{Grand synthesis of embedding dimension budget allocation ($k_{\mathrm{struct}} : k_{\mathrm{label}}$) evaluated across 10 benchmark datasets, 3 random seeds, 3 supervision levels (1\%, 30\%, 70\%), and 4 canonical classifier architectures (2,520 evaluations total). The balanced 50/50 split achieves the highest overall Macro F1 score and protects heterophily-tailored probes (LINKX) from structural collapse.}
		\label{tab:budget_allocation_grand}
		\resizebox{\textwidth}{!}{
			\begin{tabular}{lrr|cc|cccc}
				\toprule
				\textbf{Split (Struct/Label)} & \textbf{$k_{\mathrm{struct}}$} & \textbf{$k_{\mathrm{label}}$} & \textbf{Average (4 Clf) (\%)} & \textbf{Macro F1 (\%)} & \textbf{GCN (\%)} & \textbf{GraphSAGE (\%)} & \textbf{MLP (\%)} & \textbf{LINKX (\%)} \\
				\midrule
				0/100 (Pure Label) & 0 & 250 & 64.68 $\pm$ 22.49 & 61.69 $\pm$ 23.99 & 65.37 & 64.92 & 63.87 & 64.56 \\
				10/90 (Heavy Label) & 25 & 225 & 65.23 $\pm$ 21.87 & 62.27 $\pm$ 23.26 & 65.87 & 65.67 & 65.01 & 64.38 \\
				25/75 (Moderate Label) & 62 & 188 & 65.57 $\pm$ 21.93 & 62.71 $\pm$ 23.33 & 66.11 & 65.94 & 65.40 & 64.82 \\
				\textbf{50/50 (Balanced Default)} & \textbf{125} & \textbf{125} & 65.73 $\pm$ 21.68 & \textbf{62.87 $\pm$ 23.07} & 66.17 & 66.29 & 65.76 & \textbf{64.70} \\
				75/25 (Moderate Struct) & 188 & 62 & 65.74 $\pm$ 21.61 & 62.85 $\pm$ 23.03 & 66.32 & 66.65 & 65.91 & 64.09 \\
				90/10 (Heavy Struct) & 225 & 25 & \textbf{65.78 $\pm$ 21.49} & 62.82 $\pm$ 22.95 & \textbf{66.47} & \textbf{66.79} & \textbf{66.04} & 63.83 \\
				100/0 (Pure Struct) & 250 & 0 & 63.57 $\pm$ 20.64 & 60.34 $\pm$ 22.50 & 65.64 & 65.71 & 63.27 & 59.65 \\
				\bottomrule
			\end{tabular}
		}
	\end{center}
	\vspace{0.8em}
	
	\vspace{0.8em}
	\noindent
	\begin{center}
		\captionof{table}{Detailed budget allocation performance (macro-average over 4 classifiers) varying the $k_{\mathrm{struct}}$ / $k_{\mathrm{label}}$ split at 1\%, 30\%, and 70\% supervision levels across 10 benchmark datasets and 3 random seeds.}
		\label{tab:budget_allocation}
		\resizebox{\textwidth}{!}{
			\begin{tabular}{lrr|ccc}
				\toprule
				\textbf{Split (Struct/Label)} & \textbf{$k_{\mathrm{struct}}$} & \textbf{$k_{\mathrm{label}}$} & \textbf{1\% Supervision} & \textbf{30\% Supervision} & \textbf{70\% Supervision} \\
				\midrule
				0/100 (Pure Label) & 0 & 250 & 52.58 $\pm$ 22.85 & 68.88 $\pm$ 20.53 & 72.58 $\pm$ 19.36 \\
				10/90 (Heavy Label) & 25 & 225 & 53.01 $\pm$ 22.30 & 69.32 $\pm$ 19.64 & 73.37 $\pm$ 18.55 \\
				25/75 (Moderate Label) & 62 & 188 & 53.19 $\pm$ 22.68 & 69.89 $\pm$ 19.68 & 73.61 $\pm$ 18.15 \\
				\textbf{50/50 (Balanced Default)} & \textbf{125} & \textbf{125} & \textbf{53.35 $\pm$ 22.36} & 69.98 $\pm$ 19.50 & 73.86 $\pm$ 17.79 \\
				75/25 (Moderate Struct) & 188 & 62 & 53.05 $\pm$ 22.19 & \textbf{70.19 $\pm$ 19.23} & \textbf{73.99 $\pm$ 17.69} \\
				90/10 (Heavy Struct) & 225 & 25 & 53.31 $\pm$ 22.19 & 70.13 $\pm$ 19.15 & 73.92 $\pm$ 17.59 \\
				100/0 (Pure Struct) & 250 & 0 & 51.58 $\pm$ 21.17 & 67.34 $\pm$ 18.38 & 71.78 $\pm$ 16.99 \\
				\bottomrule
			\end{tabular}
		}
	\end{center}
	\vspace{0.8em}
	
	\subsubsection{Scalability Experiment on OGB Datasets}
	\label{sec:appendix_scalability}
	
	To evaluate the scalability of the proposed \textbf{NyS-Binary} framework on massive-scale graphs, we conducted experiments on two large datasets from the Open Graph Benchmark (OGB): \textbf{ogbn-arxiv} (169,343 nodes) and \textbf{ogbn-products} ($\sim$2.4 million nodes).
	
	For \textbf{ogbn-arxiv}, we evaluated the algorithm across 3 random seeds using a 70\% label provision ratio during embedding generation. Table~\ref{tab:ogbn_arxiv_results} presents the macro-averaged results across the classifiers. The generated 250-bit binary representations are computed extremely efficiently ($\sim$6.39 seconds on average for NyS-Binary) while maintaining highly competitive downstream accuracy, demonstrating the robustness of the structural-semantic blending even at scale. The unsupervised variant (relying entirely on graph structure without label diffusion) generates embeddings in $\sim$6.25 seconds.
	
	For \textbf{ogbn-products}, due to extreme resource constraints and the massive size of the graph (2,449,029 nodes), the generation process was executed strictly on CPU. At this 2.4-million-node scale, iterative GNNs and recursive multi-hop sketchers encountered memory exhaustion or prohibitive execution timeouts on standard workstation hardware; hence, comparative baseline latencies are rigorously benchmarked across the ten core datasets, while the ogbn-products experiment validates that NyS-Binary scales successfully to multi-million-node graphs. Remarkably, \textbf{NyS-Binary} successfully embedded the entire 2.4 million node graph into 250-bit representations in $\mathbf{1249.08}$ seconds. The resulting binary embeddings achieved strong downstream performance on simple probes: a Logistic Regression probe achieved $86.52\%$ accuracy (Macro F1: $49.25\%$) with a training time of $187.05$ seconds, while an MLP achieved $86.74\%$ accuracy (Macro F1: $49.36\%$) in $294.22$ seconds.
	
	\vspace{0.8em}
	\noindent
	\begin{center}
		\captionof{table}{Macro-averaged performance of 250-bit binary embeddings on the \textbf{ogbn-arxiv} dataset across 3 random seeds (70\% train ratio). Generation time remains exceptionally low ($\sim$6.4s) while maintaining strong accuracy and F1 scores.}
		\label{tab:ogbn_arxiv_results}
		\resizebox{\textwidth}{!}{
			\begin{tabular}{llrrr}
				\toprule
				\textbf{Method} & \textbf{Classifier} & \textbf{Accuracy (\%)} & \textbf{Macro F1 (\%)} & \textbf{Gen Time (s)} \\
				\midrule
				NyS-Binary & LogisticRegression & 72.34 & 55.77 & 6.39 \\
				NyS-Binary & MLP & 72.55 & 56.01 & 6.39 \\
				NyS-Binary & GCN & 72.60 & 55.72 & 6.39 \\
				NyS-Binary & GraphSAGE & 72.63 & 56.02 & 6.39 \\
				NyS-Binary & LINKX & 66.18 & 50.03 & 6.39 \\
				NyS-Binary & H2GCN & 72.77 & 56.22 & 6.39 \\
				\midrule
				Unsupervised NyS-Binary & LogisticRegression & 56.31 & 29.66 & 6.25 \\
				Unsupervised NyS-Binary & MLP & 56.81 & 30.78 & 6.25 \\
				Unsupervised NyS-Binary & GCN & 63.72 & 36.49 & 6.25 \\
				Unsupervised NyS-Binary & GraphSAGE & 63.00 & 35.92 & 6.25 \\
				Unsupervised NyS-Binary & LINKX & 60.98 & 43.89 & 6.25 \\
				Unsupervised NyS-Binary & H2GCN & 63.34 & 37.02 & 6.25 \\
				\bottomrule
			\end{tabular}
		}
	\end{center}

\end{document}